\documentclass{egpubl}
\usepackage{eurovis2022}
\usepackage{graphicx}
\graphicspath{ {./images/} }
\usepackage{enumitem}
\usepackage{pifont}% http://ctan.org/pkg/pifont
\newcommand{\cmark}{\ding{51}}%
\newcommand{\xmark}{\ding{55}}%
\usepackage[table,xcdraw]{xcolor}
\definecolor{orange}{RGB}{237,125,49}
\definecolor{green}{RGB}{112,173,71}
\definecolor{blue}{RGB}{68,114,196}
\definecolor{red}{RGB}{255,0,0}
\definecolor{purple}{RGB}{112,48,160}
\definecolor{brown}{RGB}{165,42,42}
\definecolor{gold}{rgb}{0.83, 0.69, 0.22}
\definecolor{fluorescentpink}{rgb}{1.0, 0.08, 0.58}
\definecolor{lightseagreen}{rgb}{0.13, 0.7, 0.67}
\setlist[itemize]{leftmargin=*,label=\largebullet}
\newcommand{\largebullet}{\scalebox{1.05}{$\bullet$}} 

\usepackage{tikz}
\usetikzlibrary{decorations.pathmorphing}

\usepackage{amssymb}
 \STAR                   % uncomment for STAR contribution
 \STAREurovis   % for EuroVis additional material 
\usepackage[T1]{fontenc}
\usepackage{dfadobe}  

\usepackage{cite}  % comment out for biblatex with backend=biber
\BibtexOrBiblatex
\electronicVersion
\PrintedOrElectronic
\usepackage{egweblnk}
\usepackage{color}
\usepackage{multirow}

\newcommand{\change}[1]{{\leavevmode\color{black}#1}}%Revision change
\title[EG \LaTeX\ Author Guidelines]%
{Natural Language Generation for Visualizations: State of the Art, Challenges and Future Directions}

\author[E. Hoque \& Mohammed Saidul Islam]
{\parbox{\textwidth}{\centering E. Hoque$^{1}$
		and M. Saidul Islam$^{1}$
	}
	\\
	{\parbox{\textwidth}{\centering $^1$Intelligent Visualization Lab, York University, Toronto, Canada
		}
	}
}
\begin{document}

\maketitle
%-------------------------------------------------------------------------

%-------------------------------------------------------------------------

%-------------------------------------------------------------------------
\begin{abstract}
Natural language and visualization are two complementary modalities of human communication that play a crucial role in conveying information effectively. While visualizations help people discover trends, patterns, and anomalies in data, natural language descriptions help explain these insights. Thus, combining text with visualizations is a prevalent technique for effectively delivering the core message of the data. Given the rise of natural language generation (NLG), there is a growing interest in automatically creating natural language descriptions for visualizations, which can be used as chart captions, answering questions about charts, or telling data-driven stories. In this survey, we systematically review the state of the art on NLG for visualizations and introduce a taxonomy of the problem. The NLG tasks fall within the domain of Natural Language Interfaces (NLI) for visualization, an area that has garnered significant attention from both the research community and industry. To narrow down the scope of the survey, we primarily concentrate on the research works that focus on text generation for visualizations. To characterize the NLG problem and the design space of proposed solutions, we pose \change{five Wh-questions}, why and how NLG tasks are performed for visualizations, what the task inputs and outputs are, as well as where and when the generated texts are integrated with visualizations. We categorize the solutions used in the surveyed papers based on these "\change{five Wh-questions}." Finally, we discuss the key challenges and potential avenues for future research in this domain.



\end{abstract}

\vspace{-4mm}
\section{Introduction}
\label{sec-intro}

Natural language and visualization are two powerful modalities of human communication. Visualizations are effective for identifying patterns, trends, and outliers in data, while natural language can explain key insights in visualizations~\cite{hoque2022chart}. People often combine text with charts to effectively convey the primary data message, direct readers' focus to specific chart elements, and provide explanations that might otherwise go unnoticed~\cite{kosara2013storytelling}. Recent research suggests that users prefer charts with more textual annotations explaining key points over charts with fewer annotations or text alone~\cite{stokes2022striking}. In this context, the question arises: Can we automatically generate natural language text to explain charts, create data-driven stories, and answer questions about charts?

Natural Language Generation (NLG) is a sub-field of AI and computational linguistics dedicated to creating systems that generate human-like text, enabling machines to communicate insights and explanations~\cite{reiter1997building}. Recent years have witnessed remarkable progress in NLG technology, particularly with the development and widespread adoption of large language models 
like ChatGPT~\cite{Chatgpt}. With the ability to generate human-like text, such models can be leveraged to solve various natural language interaction tasks with visualizations\cite{hoque2018applying-pragmatics, srinivasan2018orko}. 
NLG techniques have been applied to Visualizations in generating natural language captions to summarize trends, patterns, and outliers in charts~\cite{hoque2022chart, tang2023vistext}, as well as generating explanatory answers to questions about charts~\cite{kantharaj-etal-2022-opencqa}.
Researchers have also explored how to automatically incorporate generated texts with visualizations to tell data-driven stories~\cite{shi2021calliope}.

\begin{figure*}
    \centering
    \includegraphics[width=\textwidth]{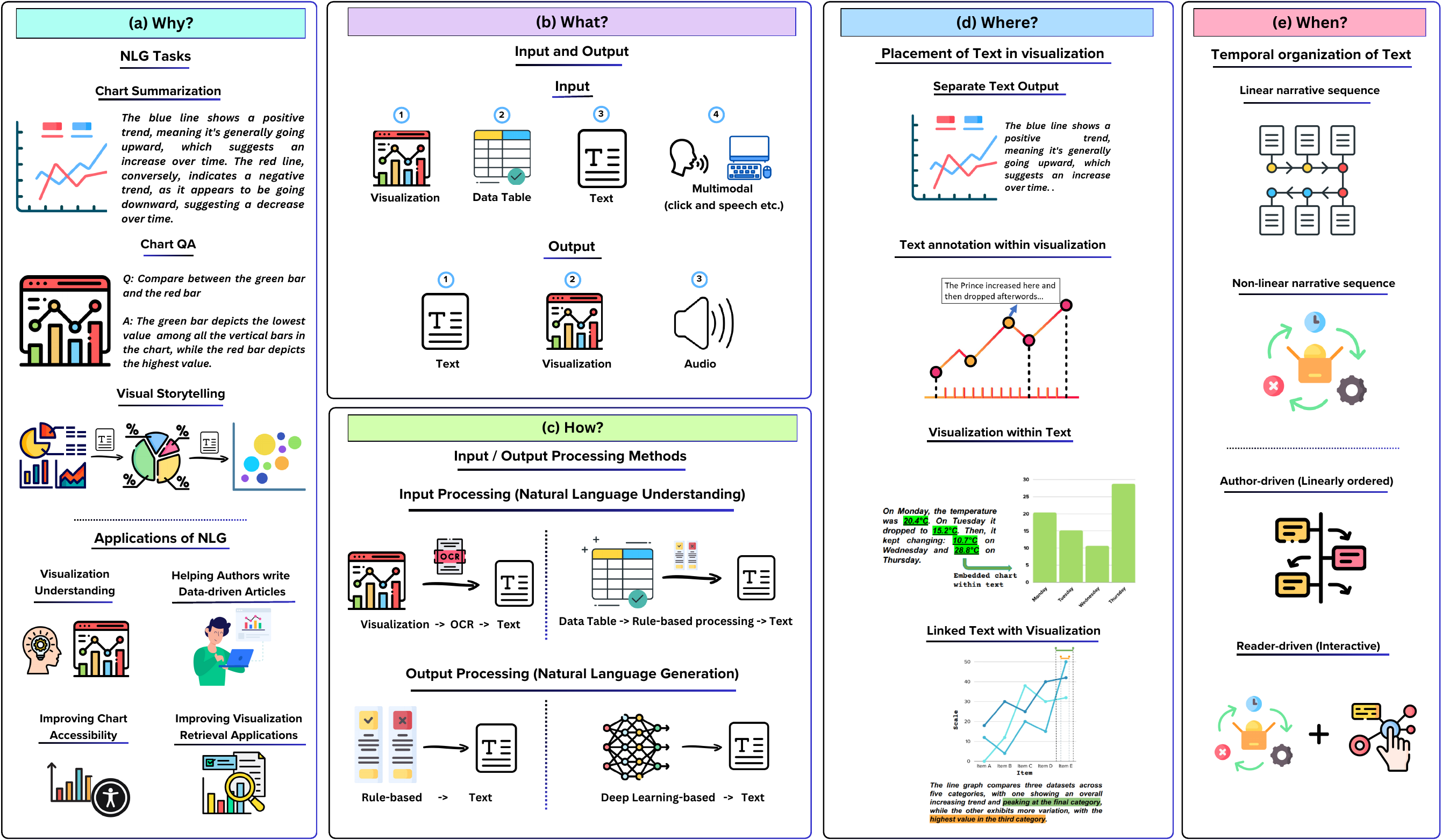}
    \caption{An overview of the problem space of NLG with visualization, covering each of the \textbf{Wh-question} dimensions. Here, from left, part (a) represents the \textbf{Why} dimension, (b) represents the \textbf{What} dimension, (c) represents the \textbf{How} dimension, (d) represents the \textbf{Where} dimension, and (e) represents the \textbf{When} dimension. \change{In the `\textit{What}' dimension, the numbers (e.g., `1', `2', `3', etc.) in both the input and output refer to individual input and output types. However, these individual input/output types can also be combined to create a combined input/output type.}}
    \label{fig:nlg-vis-overview}
\end{figure*}

In this survey, we systematically review the state of the art on natural language generation for visualizations and introduce a taxonomy of the problem as outlined in Section~\ref{sec: pso}. The NLG tasks fall within the domain of Natural Language Interfaces (NLI) for visualization, an area that has garnered significant attention from both the research community \cite{hoque2018applying-pragmatics, srinivasan2018orko} and industry \cite{askdata}. 
To narrow down the scope of the survey, we primarily concentrate on research works that focus on text generation for visualizations. 
In order to characterize the NLG problem and the design space of proposed solutions, we pose \change{five Wh-questions}, \textit{why} and \textit{how} NLG tasks are performed for visualizations, \textit{what} the task inputs and outputs are, as well as \textit{where} and \textit{when} the generated texts are integrated with visualizations (see Figure \ref{fig:nlg-vis-overview}). We categorize the solutions 
used in the surveyed papers based on these ``\change{five Wh-questions}.'' Finally, we discuss the challenges and potential avenues for future research in this domain.

Despite the surging interest in NLG for visualizations, a comprehensive survey on this topic is notably absent.
Some recent surveys focus on natural language interfaces for visualizations \cite{srinivasan2020ask, voigt-etal-2022-survey, shen2023towards}, however, they focus more broadly on how to explore data and perform visual analysis using natural language without providing specific focus on NLG. Others have focused on chart question answering \cite{hoque2022chart} and visual data storytelling~\cite{riche2018data, tong2018storytelling} but they cover NLG very briefly. 
Others address NLG tasks in domains that are not related to visualizations, i.e., automatic text summarization \cite{kassas2021automatic} and open domain question answering \cite{zhu2021retrieving} as well as hallucinations  \cite{ji2023surveyhallucinations} and faithfulness in the generated text \cite{li2022faithfulness}.
However, a dedicated survey exclusively focusing on NLG tasks for visualizations is lacking. This survey aims to bridge that gap by proposing a formal taxonomy 
and categorizing existing solutions.

\section{Methodology and Outline of the Survey}\label{sec: pso}
\change{We first discuss the methodology for conducting the survey including the selection criteria followed by an outline of the survey.}

\change{\subsection{Survey Methodology}}
\change{To thoroughly examine the integration of natural language generation (NLG) in visualization, we conducted a wide-ranging review of relevant academic publications spanning the last twenty years (2004-2024). Our methodology involved searching for key terms such as ``natural language generation with visualization'', ``natural language interface for visualizations'', ``chart question answering, ``chart summarization'', ``chart captioning'', ``visual storytelling'', ``narrative visualization'', ``Figure captioning'', ``automated data-storytelling'' etc. in Google Scholar and Arxiv, resulting in over 500 papers. These papers originated from diverse fields ranging from visualization (VIS), natural language processing (NLP), human-computer interaction (HCI), and Computer Vision (CV), drawn from top-tier conferences and journals like IEEE Vis, EuroVis, TVCG, ACL, and EMNLP (further venues are listed in Table \ref{table:venues}). 

During the review process, we first examined the titles of papers from these publications to identify candidate papers for inclusion in our survey. Next, we scrutinized the abstracts of the candidate papers to determine whether they were related to the Natural Language Generation in the Visualization domain.  Finally, we thoroughly examined the papers, applying specific screening criteria (SCs) to find the most relevant papers:

\begin{enumerate}[label=\textbf{SC\arabic*:},leftmargin=*]
    \item Articles related to visualization research that do not involve machine-generated natural language text (or speech) were excluded. We specifically targeted outputs produced through computational means, \change{rather than those authored by humans \cite{mörth2022scrollyvis, kwon2014visjockey}}. 
   \item Papers that focus on generating text but do not involve visualizations were excluded. \change{For example, natural language text generation for texts \cite{fan-etal-2018-hierarchical, fang2021outline, wang-etal-2023-improving-pacing} and tabular data \cite{puduppully2019datatotext, ribeiro-etal-2021-investigating, song-etal-2020-structural} were not included in the survey.} Interested readers are referred to existing survey papers for further exploration of these areas (e.g., \cite{zhang2023survey, ji2023survey, celikyilmaz2020evaluation}).
    \item We did not focus on Chart question answering or NLI systems that primarily involved natural language understanding (e.g., question, user intents) \change{but did not generate explanatory texts \cite{masry-etal-2022-chartqa, kahou2018figureqa, kushal2018dvqa}}. For further exploration of these areas, readers are directed to existing survey papers~\cite{hoque2022chart, srinivasan2020ask, voigt-etal-2022-survey, shen2023towards}.
\end{enumerate}

Additionally, we considered the works that take data tables as input and produce both texts and visualizations, as the generated texts are related to visualizations to select the final candidate papers for our review. Furthermore, guided by our established screening-out criteria (SCs), we performed a thorough evaluation of the articles selected, so that each article met our stringent standards for both research quality and direct relevance to our study. Simultaneously, we assessed the impact and significance of the research findings, favoring articles that provided fresh perspectives or marked a step forward within the field. This thorough evaluation yielded 122 papers that are broadly related to the survey topic, out of which 30 papers form the core group that matches the strict screening criteria (listed in Table \ref{tab:literature-alongwh-dim}).

After refining the survey scope, we analyzed the relevant papers to delineate categories through an iterative coding approach, aiding us in characterizing the problem space. Specifically, we identified five dimensions of the problem posed as \change{five Wh-questions}: \textit{why} and \textit{how} NLG tasks are performed for visualizations, \textit{what} the task inputs and outputs are, as well as \textit{where} and \textit{when} the generated texts are integrated with visualizations (see Figure \ref{fig:nlg-vis-overview}). We categorized the solutions utilized in the surveyed papers based on these "\change{five Wh-questions}." Our rationale for posing "Wh" questions is that they facilitate systematic analysis and comprehension of the survey topic and they have been adopted in visualization literature~\cite{brehmer2013multi, voigt2022and}. For example, Brehmer and Munzner expressed the typology of abstract visualization tasks using why, how, and what questions \cite{brehmer2013multi}. However, the questions of our survey dimensions are posed in somewhat different contexts (e.g., `how' refers to how the NLG method processes the input and generates the output as opposed to methods for visual encoding and interaction techniques in ~\cite{brehmer2013multi}) and involve additional questions (`when' and `where') absent in ~\cite{brehmer2013multi}.}

\begin{table}[t!]
\centering
\caption{Relevant venues for the survey.}
\resizebox{\columnwidth}{!}{%
\begin{tabular}{ll}
\hline
\textbf{VIS} & IEEE VIS, PacificVIS, TVCG, \\
             & EuroVis (Computer Graphics Forum) \\ \hline
\textbf{NLP} & ACL, EMNLP, EACL, iNLG, CL, TACL \\             %AACL-IJCNLP
\textbf{HCI} & CHI, TOCHI, UIST, IUI \\ \hline
\textbf{ML \& CV} & ICML, ICLR, CVPR \\ \hline %SIGIR
\hline
\end{tabular} %
}
\label{table:venues}
\end{table}

\change{\subsection{Survey Outline}}
Figure \ref{fig:nlg-vis-overview} visually summarizes the design space of the NLG with visualization problem along the \change{five Wh-questions} dimensions that we summarize below: 
% % Please add the following required packages to your document preamble:
% % \usepackage{multirow}
% % \usepackage{graphicx}
% \begin{table*}[]
% \centering
% \caption{The table presents a summary of the relation between our proposed taxonomy and existing literature and how existing works cover each of the \textbf{Wh-Question} aspects. Here, ``\textcolor{green}\cmark'' denotes a research work is relevant to any one of the dimensions, while ``\textcolor{red}\xmark'' denotes otherwise.}

\begin{table*}[]
\centering
\caption{The table presents a summary of the relation between our proposed taxonomy and existing literature and how existing works cover each of the \textbf{Wh-Question} dimensions. Here, ``\textcolor{green}\cmark'' denotes a research work relevant to any one of the dimensions, while ``\textcolor{red}\xmark'' denotes otherwise. Additionally, \textbf{Chart Summ.} denotes \textbf{Chart Summarization}, \textbf{Chart Cap.} denotes \textbf{Chart Captioning}, \textbf{Vis.} denotes 
%\textbf{Visual} (in case of Visual Storytelling) / 
\textbf{Visualization}. %(otherwise).
}
\label{tab:literature-alongwh-dim}
\resizebox{\textwidth}{!}{%
\begin{tabular}{|c|cccccccccccccccccccc|} \hline
&
  \multicolumn{20}{c|}{Dimensions} \\ \cline{2-21} 
 &
  \multicolumn{3}{c|}{Why?} &
  \multicolumn{7}{c|}{What?} &
  \multicolumn{2}{c|}{How?} &
  \multicolumn{4}{c|}{Where?} &
  \multicolumn{4}{c|}{When?} \\ \cline{2-21} 
 &
  \multicolumn{1}{c|}{} &
  \multicolumn{1}{c|}{} &
  \multicolumn{1}{c|}{} &
  \multicolumn{4}{c|}{Input} &
  \multicolumn{3}{c|}{Output} &
  \multicolumn{1}{c|}{} &
  \multicolumn{1}{c|}{} &
  \multicolumn{1}{c|}{} &
  \multicolumn{1}{c|}{} &
  \multicolumn{1}{c|}{} &
  \multicolumn{1}{c|}{} &
  \multicolumn{1}{c|}{} &
  \multicolumn{1}{c|}{} &
  \multicolumn{1}{c|}{} &
   \\ \cline{5-11}
 &
  \multicolumn{1}{c|}{} &
  \multicolumn{1}{c|}{} &
  \multicolumn{1}{c|}{} &
  \multicolumn{1}{c|}{} &
  \multicolumn{1}{c|}{} &
  \multicolumn{1}{c|}{} &
  \multicolumn{1}{c|}{} &
  \multicolumn{1}{c|}{} &
  \multicolumn{1}{c|}{} &
  \multicolumn{1}{c|}{} &
  \multicolumn{1}{c|}{} &
  \multicolumn{1}{c|}{} &
  \multicolumn{1}{c|}{} &
  \multicolumn{1}{c|}{} &
  \multicolumn{1}{c|}{} &
  \multicolumn{1}{c|}{} &
  \multicolumn{1}{c|}{} &
  \multicolumn{1}{c|}{} &
  \multicolumn{1}{c|}{} &
   \\
\multirow{-5}{*}{Literature} &
  \multicolumn{1}{c|}{\multirow{-3}{*}{\begin{tabular}[c]{@{}c@{}}Chart \\ Summ. / Cap. \end{tabular}}} &
  \multicolumn{1}{c|}{\multirow{-3}{*}{ChartQA}} &
  \multicolumn{1}{c|}{\multirow{-3}{*}{\begin{tabular}[c]{@{}c@{}}Visual\\ Storytelling\end{tabular}}} &
  \multicolumn{1}{c|}{\multirow{-2}{*}{Text}} &
  \multicolumn{1}{c|}{\multirow{-2}{*}{Vis.}} &
  \multicolumn{1}{c|}{\multirow{-2}{*}{\begin{tabular}[c]{@{}c@{}}Data \\ Table\end{tabular}}} &
  \multicolumn{1}{c|}{\multirow{-2}{*}{\begin{tabular}[c]{@{}c@{}}Multi\\ modal\end{tabular}}} &
  \multicolumn{1}{c|}{\multirow{-2}{*}{Text}} &
  \multicolumn{1}{c|}{\multirow{-2}{*}{Vis.}} &
  \multicolumn{1}{c|}{\multirow{-2}{*}{\begin{tabular}[c]{@{}c@{}}Multi\\ modal\end{tabular}}} &
  \multicolumn{1}{c|}{\multirow{-3}{*}{\begin{tabular}[c]{@{}c@{}}Rule \\ based\end{tabular}}} &
  \multicolumn{1}{c|}{\multirow{-3}{*}{\begin{tabular}[c]{@{}c@{}}Deep \\ Learning \\ based\end{tabular}}} &
  \multicolumn{1}{c|}{\multirow{-3}{*}{\begin{tabular}[c]{@{}c@{}}Separate \\ Text \\ Output\end{tabular}}} &
  \multicolumn{1}{c|}{\multirow{-3}{*}{\begin{tabular}[c]{@{}c@{}}Text \\ Annotation \\ within Vis.\end{tabular}}} &
  \multicolumn{1}{c|}{\multirow{-3}{*}{\begin{tabular}[c]{@{}c@{}}Vis. \\ within \\ Text \end{tabular}}} &
  \multicolumn{1}{c|}{\multirow{-3}{*}{\begin{tabular}[c]{@{}c@{}}Linked \\ Text \\ with Vis.\end{tabular}}} &
  \multicolumn{1}{c|}{\multirow{-3}{*}{Linear}} &
  \multicolumn{1}{c|}{\multirow{-3}{*}{\begin{tabular}[c]{@{}c@{}}Non\\ linear\end{tabular}}} &
  \multicolumn{1}{c|}{\multirow{-3}{*}{\begin{tabular}[c]{@{}c@{}}Author \\ Driven\end{tabular}}} &
  \multirow{-3}{*}{\begin{tabular}[c]{@{}c@{}}Reader \\ Driven\end{tabular}} \\ \hline
\rowcolor[HTML]{EFEFEF} 
Mittal et al. \cite{mittal-etal-1998-describing} &
  \multicolumn{1}{c|}{\cellcolor[HTML]{EFEFEF}\textcolor{green}\cmark} &
  \multicolumn{1}{c|}{\cellcolor[HTML]{EFEFEF}\textcolor{red}\xmark} &
  \multicolumn{1}{c|}{\cellcolor[HTML]{EFEFEF}\textcolor{red}\xmark} &
  \multicolumn{1}{c|}{\cellcolor[HTML]{EFEFEF}\textcolor{green}\cmark} &
  \multicolumn{1}{c|}{\cellcolor[HTML]{EFEFEF}\textcolor{red}\xmark} &
  \multicolumn{1}{c|}{\cellcolor[HTML]{EFEFEF}\textcolor{green}\cmark} &
  \multicolumn{1}{c|}{\cellcolor[HTML]{EFEFEF}\textcolor{red}\xmark} &
  \multicolumn{1}{c|}{\cellcolor[HTML]{EFEFEF}\textcolor{green}\cmark} &
  \multicolumn{1}{c|}{\cellcolor[HTML]{EFEFEF}\textcolor{red}\xmark} &
  \multicolumn{1}{c|}{\cellcolor[HTML]{EFEFEF}\textcolor{red}\xmark} &
  \multicolumn{1}{c|}{\cellcolor[HTML]{EFEFEF}\textcolor{green}\cmark} &
  \multicolumn{1}{c|}{\cellcolor[HTML]{EFEFEF}\textcolor{red}\xmark} &
  \multicolumn{1}{c|}{\cellcolor[HTML]{EFEFEF}\textcolor{green}\cmark} &
  \multicolumn{1}{c|}{\cellcolor[HTML]{EFEFEF}\textcolor{red}\xmark} &
  \multicolumn{1}{c|}{\cellcolor[HTML]{EFEFEF}\textcolor{red}\xmark} &
  \multicolumn{1}{c|}{\cellcolor[HTML]{EFEFEF}\textcolor{red}\xmark} &
  \multicolumn{1}{c|}{\cellcolor[HTML]{EFEFEF}\textcolor{red}\xmark} &
  \multicolumn{1}{c|}{\cellcolor[HTML]{EFEFEF}\textcolor{red}\xmark} &
  \multicolumn{1}{c|}{\cellcolor[HTML]{EFEFEF}\textcolor{red}\xmark} &
  \textcolor{red}\xmark \\
\rowcolor[HTML]{FFFFFF} 
Reiter et al. \cite{reiter2007architecture} &
  \multicolumn{1}{c|}{\cellcolor[HTML]{FFFFFF}\textcolor{green}\cmark} &
  \multicolumn{1}{c|}{\cellcolor[HTML]{FFFFFF}\textcolor{red}\xmark} &
  \multicolumn{1}{c|}{\cellcolor[HTML]{FFFFFF}\textcolor{red}\xmark} &
  \multicolumn{1}{c|}{\cellcolor[HTML]{FFFFFF}\textcolor{green}\cmark} &
  \multicolumn{1}{c|}{\cellcolor[HTML]{FFFFFF}\textcolor{red}\xmark} &
  \multicolumn{1}{c|}{\cellcolor[HTML]{FFFFFF}\textcolor{green}\cmark} &
  \multicolumn{1}{c|}{\cellcolor[HTML]{FFFFFF}\textcolor{red}\xmark} &
  \multicolumn{1}{c|}{\cellcolor[HTML]{FFFFFF}\textcolor{green}\cmark} &
  \multicolumn{1}{c|}{\cellcolor[HTML]{FFFFFF}\textcolor{red}\xmark} &
  \multicolumn{1}{c|}{\cellcolor[HTML]{FFFFFF}\textcolor{red}\xmark} &
  \multicolumn{1}{c|}{\cellcolor[HTML]{FFFFFF}\textcolor{green}\cmark} &
  \multicolumn{1}{c|}{\cellcolor[HTML]{FFFFFF}\textcolor{red}\xmark} &
  \multicolumn{1}{c|}{\cellcolor[HTML]{FFFFFF}\textcolor{green}\cmark} &
  \multicolumn{1}{c|}{\cellcolor[HTML]{FFFFFF}\textcolor{red}\xmark} &
  \multicolumn{1}{c|}{\cellcolor[HTML]{FFFFFF}\textcolor{red}\xmark} &
  \multicolumn{1}{c|}{\cellcolor[HTML]{FFFFFF}\textcolor{red}\xmark} &
  \multicolumn{1}{c|}{\cellcolor[HTML]{FFFFFF}\textcolor{red}\xmark} &
  \multicolumn{1}{c|}{\cellcolor[HTML]{FFFFFF}\textcolor{red}\xmark} &
  \multicolumn{1}{c|}{\cellcolor[HTML]{FFFFFF}\textcolor{red}\xmark} &
  \cellcolor[HTML]{FFFFFF} \textcolor{red}\xmark \\
\rowcolor[HTML]{EFEFEF} 
Demir et al. \cite{demir2012summarizing} &
  \multicolumn{1}{c|}{\cellcolor[HTML]{EFEFEF}\textcolor{green}\cmark} &
  \multicolumn{1}{c|}{\cellcolor[HTML]{EFEFEF}\textcolor{red}\xmark} &
  \multicolumn{1}{c|}{\cellcolor[HTML]{EFEFEF}\textcolor{red}\xmark} &
  \multicolumn{1}{c|}{\cellcolor[HTML]{EFEFEF}\textcolor{green}\cmark} &
  \multicolumn{1}{c|}{\cellcolor[HTML]{EFEFEF}\textcolor{red}\xmark} &
  \multicolumn{1}{c|}{\cellcolor[HTML]{EFEFEF}\textcolor{red}\xmark} &
  \multicolumn{1}{c|}{\cellcolor[HTML]{EFEFEF}\textcolor{red}\xmark} &
  \multicolumn{1}{c|}{\cellcolor[HTML]{EFEFEF}\textcolor{green}\cmark} &
  \multicolumn{1}{c|}{\cellcolor[HTML]{EFEFEF}\textcolor{red}\xmark} &
  \multicolumn{1}{c|}{\cellcolor[HTML]{EFEFEF}\textcolor{red}\xmark} &
  \multicolumn{1}{c|}{\cellcolor[HTML]{EFEFEF}\textcolor{green}\cmark} &
  \multicolumn{1}{c|}{\cellcolor[HTML]{EFEFEF}\textcolor{red}\xmark} &
  \multicolumn{1}{c|}{\cellcolor[HTML]{EFEFEF}\textcolor{green}\cmark} &
  \multicolumn{1}{c|}{\cellcolor[HTML]{EFEFEF}\textcolor{red}\xmark} &
  \multicolumn{1}{c|}{\cellcolor[HTML]{EFEFEF}\textcolor{red}\xmark} &
  \multicolumn{1}{c|}{\cellcolor[HTML]{EFEFEF}\textcolor{red}\xmark} &
  \multicolumn{1}{c|}{\cellcolor[HTML]{EFEFEF}\textcolor{red}\xmark} &
  \multicolumn{1}{c|}{\cellcolor[HTML]{EFEFEF}\textcolor{red}\xmark} &
  \multicolumn{1}{c|}{\cellcolor[HTML]{EFEFEF}\textcolor{red}\xmark} &
  \textcolor{red}\xmark \\
\rowcolor[HTML]{FFFFFF} 
Ferres et al. \cite{ferres2013evaluating} &
  \multicolumn{1}{c|}{\cellcolor[HTML]{FFFFFF}\textcolor{green}\cmark} &
  \multicolumn{1}{c|}{\cellcolor[HTML]{FFFFFF}\textcolor{red}\xmark} &
  \multicolumn{1}{c|}{\cellcolor[HTML]{FFFFFF}\textcolor{red}\xmark} &
  \multicolumn{1}{c|}{\cellcolor[HTML]{FFFFFF}\textcolor{red}\xmark} &
  \multicolumn{1}{c|}{\cellcolor[HTML]{FFFFFF}\textcolor{red}\xmark} &
  \multicolumn{1}{c|}{\cellcolor[HTML]{FFFFFF}\textcolor{green}\cmark} &
  \multicolumn{1}{c|}{\cellcolor[HTML]{FFFFFF}\textcolor{red}\xmark} &
  \multicolumn{1}{c|}{\cellcolor[HTML]{FFFFFF}\textcolor{green}\cmark} &
  \multicolumn{1}{c|}{\cellcolor[HTML]{FFFFFF}\textcolor{red}\xmark} &
  \multicolumn{1}{c|}{\cellcolor[HTML]{FFFFFF}\textcolor{red}\xmark} &
  \multicolumn{1}{c|}{\cellcolor[HTML]{FFFFFF}\textcolor{green}\cmark} &
  \multicolumn{1}{c|}{\cellcolor[HTML]{FFFFFF}\textcolor{red}\xmark} &
  \multicolumn{1}{c|}{\cellcolor[HTML]{FFFFFF}\textcolor{green}\cmark} &
  \multicolumn{1}{c|}{\cellcolor[HTML]{FFFFFF}\textcolor{red}\xmark} &
  \multicolumn{1}{c|}{\cellcolor[HTML]{FFFFFF}\textcolor{red}\xmark} &
  \multicolumn{1}{c|}{\cellcolor[HTML]{FFFFFF}\textcolor{red}\xmark} &
  \multicolumn{1}{c|}{\cellcolor[HTML]{FFFFFF}\textcolor{red}\xmark} &
  \multicolumn{1}{c|}{\cellcolor[HTML]{FFFFFF}\textcolor{red}\xmark} &
  \multicolumn{1}{c|}{\cellcolor[HTML]{FFFFFF}\textcolor{red}\xmark} &
  \textcolor{red}\xmark \\
\rowcolor[HTML]{EFEFEF} 
Chen et al. \cite{chen2019figure} &
  \multicolumn{1}{c|}{\cellcolor[HTML]{EFEFEF}\textcolor{green}\cmark} &
  \multicolumn{1}{c|}{\cellcolor[HTML]{EFEFEF}\textcolor{red}\xmark} &
  \multicolumn{1}{c|}{\cellcolor[HTML]{EFEFEF}\textcolor{red}\xmark} &
  \multicolumn{1}{c|}{\cellcolor[HTML]{EFEFEF}\textcolor{red}\xmark} &
  \multicolumn{1}{c|}{\cellcolor[HTML]{EFEFEF}\textcolor{green}\cmark} &
  \multicolumn{1}{c|}{\cellcolor[HTML]{EFEFEF}\textcolor{red}\xmark} &
  \multicolumn{1}{c|}{\cellcolor[HTML]{EFEFEF}\textcolor{red}\xmark} &
  \multicolumn{1}{c|}{\cellcolor[HTML]{EFEFEF}\textcolor{green}\cmark} &
  \multicolumn{1}{c|}{\cellcolor[HTML]{EFEFEF}\textcolor{red}\xmark} &
  \multicolumn{1}{c|}{\cellcolor[HTML]{EFEFEF}\textcolor{red}\xmark} &
  \multicolumn{1}{c|}{\cellcolor[HTML]{EFEFEF}\textcolor{red}\xmark} &
  \multicolumn{1}{c|}{\cellcolor[HTML]{EFEFEF}\textcolor{green}\cmark} &
  \multicolumn{1}{c|}{\cellcolor[HTML]{EFEFEF}\textcolor{green}\cmark} &
  \multicolumn{1}{c|}{\cellcolor[HTML]{EFEFEF}\textcolor{red}\xmark} &
  \multicolumn{1}{c|}{\cellcolor[HTML]{EFEFEF}\textcolor{red}\xmark} &
  \multicolumn{1}{c|}{\cellcolor[HTML]{EFEFEF}\textcolor{red}\xmark} &
  \multicolumn{1}{c|}{\cellcolor[HTML]{EFEFEF}\textcolor{red}\xmark} &
  \multicolumn{1}{c|}{\cellcolor[HTML]{EFEFEF}\textcolor{red}\xmark} &
  \multicolumn{1}{c|}{\cellcolor[HTML]{EFEFEF}\textcolor{red}\xmark} &
  \textcolor{red}\xmark \\
\rowcolor[HTML]{FFFFFF} 
Voder \cite{srinivasan2019augmenting} &
  \multicolumn{1}{c|}{\cellcolor[HTML]{FFFFFF}\textcolor{green}\cmark} &
  \multicolumn{1}{c|}{\cellcolor[HTML]{FFFFFF}\textcolor{red}\xmark} &
  \multicolumn{1}{c|}{\cellcolor[HTML]{FFFFFF}\textcolor{red}\xmark} &
  \multicolumn{1}{c|}{\cellcolor[HTML]{FFFFFF}\textcolor{red}\xmark} &
  \multicolumn{1}{c|}{\cellcolor[HTML]{FFFFFF}\textcolor{green}\cmark} &
  \multicolumn{1}{c|}{\cellcolor[HTML]{FFFFFF}\textcolor{red}\xmark} &
  \multicolumn{1}{c|}{\cellcolor[HTML]{FFFFFF}\textcolor{red}\xmark} &
  \multicolumn{1}{c|}{\cellcolor[HTML]{FFFFFF}\textcolor{green}\cmark} &
  \multicolumn{1}{c|}{\cellcolor[HTML]{FFFFFF}\textcolor{red}\xmark} &
  \multicolumn{1}{c|}{\cellcolor[HTML]{FFFFFF}\textcolor{red}\xmark} &
  \multicolumn{1}{c|}{\cellcolor[HTML]{FFFFFF}\textcolor{green}\cmark} &
  \multicolumn{1}{c|}{\cellcolor[HTML]{FFFFFF}\textcolor{red}\xmark} &
  \multicolumn{1}{c|}{\cellcolor[HTML]{FFFFFF}\textcolor{green}\cmark} &
  \multicolumn{1}{c|}{\cellcolor[HTML]{FFFFFF}\textcolor{red}\xmark} &
  \multicolumn{1}{c|}{\cellcolor[HTML]{FFFFFF}\textcolor{red}\xmark} &
  \multicolumn{1}{c|}{\cellcolor[HTML]{FFFFFF}\textcolor{green}\cmark} &
  \multicolumn{1}{c|}{\cellcolor[HTML]{FFFFFF}\textcolor{red}\xmark} &
  \multicolumn{1}{c|}{\cellcolor[HTML]{FFFFFF}\textcolor{red}\xmark} &
  \multicolumn{1}{c|}{\cellcolor[HTML]{FFFFFF}\textcolor{red}\xmark} &
  \textcolor{red}\xmark \\
\rowcolor[HTML]{EFEFEF} 
Obeid et al. \cite{obeid2020charttotext} &
  \multicolumn{1}{c|}{\cellcolor[HTML]{EFEFEF}\textcolor{green}\cmark} &
  \multicolumn{1}{c|}{\cellcolor[HTML]{EFEFEF}\textcolor{red}\xmark} &
  \multicolumn{1}{c|}{\cellcolor[HTML]{EFEFEF}\textcolor{red}\xmark} &
  \multicolumn{1}{c|}{\cellcolor[HTML]{EFEFEF}\textcolor{green}\cmark} &
  \multicolumn{1}{c|}{\cellcolor[HTML]{EFEFEF}\textcolor{red}\xmark} &
  \multicolumn{1}{c|}{\cellcolor[HTML]{EFEFEF}\textcolor{green}\cmark} &
  \multicolumn{1}{c|}{\cellcolor[HTML]{EFEFEF}\textcolor{red}\xmark} &
  \multicolumn{1}{c|}{\cellcolor[HTML]{EFEFEF}\textcolor{green}\cmark} &
  \multicolumn{1}{c|}{\cellcolor[HTML]{EFEFEF}\textcolor{red}\xmark} &
  \multicolumn{1}{c|}{\cellcolor[HTML]{EFEFEF}\textcolor{red}\xmark} &
  \multicolumn{1}{c|}{\cellcolor[HTML]{EFEFEF}\textcolor{red}\xmark} &
  \multicolumn{1}{c|}{\cellcolor[HTML]{EFEFEF}\textcolor{green}\cmark} &
  \multicolumn{1}{c|}{\cellcolor[HTML]{EFEFEF}\textcolor{green}\cmark} &
  \multicolumn{1}{c|}{\cellcolor[HTML]{EFEFEF}\textcolor{red}\xmark} &
  \multicolumn{1}{c|}{\cellcolor[HTML]{EFEFEF}\textcolor{red}\xmark} &
  \multicolumn{1}{c|}{\cellcolor[HTML]{EFEFEF}\textcolor{red}\xmark} &
  \multicolumn{1}{c|}{\cellcolor[HTML]{EFEFEF}\textcolor{red}\xmark} &
  \multicolumn{1}{c|}{\cellcolor[HTML]{EFEFEF}\textcolor{red}\xmark} &
  \multicolumn{1}{c|}{\cellcolor[HTML]{EFEFEF}\textcolor{red}\xmark} &
  \textcolor{red}\xmark \\
\rowcolor[HTML]{FFFFFF} 
Spreafico et al. \cite{speafico2020neural} &
  \multicolumn{1}{c|}{\cellcolor[HTML]{FFFFFF}\textcolor{green}\cmark} &
  \multicolumn{1}{c|}{\cellcolor[HTML]{FFFFFF}\textcolor{red}\xmark} &
  \multicolumn{1}{c|}{\cellcolor[HTML]{FFFFFF}\textcolor{red}\xmark} &
  \multicolumn{1}{c|}{\cellcolor[HTML]{FFFFFF}\textcolor{green}\cmark} &
  \multicolumn{1}{c|}{\cellcolor[HTML]{FFFFFF}\textcolor{red}\xmark} &
  \multicolumn{1}{c|}{\cellcolor[HTML]{FFFFFF}\textcolor{green}\cmark} &
  \multicolumn{1}{c|}{\cellcolor[HTML]{FFFFFF}\textcolor{red}\xmark} &
  \multicolumn{1}{c|}{\cellcolor[HTML]{FFFFFF}\textcolor{green}\cmark} &
  \multicolumn{1}{c|}{\cellcolor[HTML]{FFFFFF}\textcolor{red}\xmark} &
  \multicolumn{1}{c|}{\cellcolor[HTML]{FFFFFF}\textcolor{red}\xmark} &
  \multicolumn{1}{c|}{\cellcolor[HTML]{FFFFFF}\textcolor{red}\xmark} &
  \multicolumn{1}{c|}{\cellcolor[HTML]{FFFFFF}\textcolor{green}\cmark} &
  \multicolumn{1}{c|}{\cellcolor[HTML]{FFFFFF}\textcolor{green}\cmark} &
  \multicolumn{1}{c|}{\cellcolor[HTML]{FFFFFF}\textcolor{red}\xmark} &
  \multicolumn{1}{c|}{\cellcolor[HTML]{FFFFFF}\textcolor{red}\xmark} &
  \multicolumn{1}{c|}{\cellcolor[HTML]{FFFFFF}\textcolor{red}\xmark} &
  \multicolumn{1}{c|}{\cellcolor[HTML]{FFFFFF}\textcolor{red}\xmark} &
  \multicolumn{1}{c|}{\cellcolor[HTML]{FFFFFF}\textcolor{red}\xmark} &
  \multicolumn{1}{c|}{\cellcolor[HTML]{FFFFFF}\textcolor{red}\xmark} &
  \textcolor{red}\xmark \\
\rowcolor[HTML]{EFEFEF} 
SciCAP \cite{hsu-etal-2021-scicap-generating} &
  \multicolumn{1}{c|}{\cellcolor[HTML]{EFEFEF}\textcolor{green}\cmark} &
  \multicolumn{1}{c|}{\cellcolor[HTML]{EFEFEF}\textcolor{red}\xmark} &
  \multicolumn{1}{c|}{\cellcolor[HTML]{EFEFEF}\textcolor{red}\xmark} &
  \multicolumn{1}{c|}{\cellcolor[HTML]{EFEFEF}\textcolor{red}\xmark} &
  \multicolumn{1}{c|}{\cellcolor[HTML]{EFEFEF}\textcolor{green}\cmark} &
  \multicolumn{1}{c|}{\cellcolor[HTML]{EFEFEF}\textcolor{red}\xmark} &
  \multicolumn{1}{c|}{\cellcolor[HTML]{EFEFEF}\textcolor{red}\xmark} &
  \multicolumn{1}{c|}{\cellcolor[HTML]{EFEFEF}\textcolor{green}\cmark} &
  \multicolumn{1}{c|}{\cellcolor[HTML]{EFEFEF}\textcolor{red}\xmark} &
  \multicolumn{1}{c|}{\cellcolor[HTML]{EFEFEF}\textcolor{red}\xmark} &
  \multicolumn{1}{c|}{\cellcolor[HTML]{EFEFEF}\textcolor{red}\xmark} &
  \multicolumn{1}{c|}{\cellcolor[HTML]{EFEFEF}\textcolor{green}\cmark} &
  \multicolumn{1}{c|}{\cellcolor[HTML]{EFEFEF}\textcolor{green}\cmark} &
  \multicolumn{1}{c|}{\cellcolor[HTML]{EFEFEF}\textcolor{red}\xmark} &
  \multicolumn{1}{c|}{\cellcolor[HTML]{EFEFEF}\textcolor{red}\xmark} &
  \multicolumn{1}{c|}{\cellcolor[HTML]{EFEFEF}\textcolor{red}\xmark} &
  \multicolumn{1}{c|}{\cellcolor[HTML]{EFEFEF}\textcolor{red}\xmark} &
  \multicolumn{1}{c|}{\cellcolor[HTML]{EFEFEF}\textcolor{red}\xmark} &
  \multicolumn{1}{c|}{\cellcolor[HTML]{EFEFEF}\textcolor{red}\xmark} &
  \textcolor{red}\xmark \\
\rowcolor[HTML]{FFFFFF} 
Chart-to-Text \cite{kantharaj-etal-2022-chart} &
  \multicolumn{1}{c|}{\cellcolor[HTML]{FFFFFF}\textcolor{green}\cmark} &
  \multicolumn{1}{c|}{\cellcolor[HTML]{FFFFFF}\textcolor{red}\xmark} &
  \multicolumn{1}{c|}{\cellcolor[HTML]{FFFFFF}\textcolor{red}\xmark} &
  \multicolumn{1}{c|}{\cellcolor[HTML]{FFFFFF}\textcolor{green}\cmark} &
  \multicolumn{1}{c|}{\cellcolor[HTML]{FFFFFF}\textcolor{green}\cmark} &
  \multicolumn{1}{c|}{\cellcolor[HTML]{FFFFFF}\textcolor{green}\cmark} &
  \multicolumn{1}{c|}{\cellcolor[HTML]{FFFFFF}\textcolor{red}\xmark} &
  \multicolumn{1}{c|}{\cellcolor[HTML]{FFFFFF}\textcolor{green}\cmark} &
  \multicolumn{1}{c|}{\cellcolor[HTML]{FFFFFF}\textcolor{red}\xmark} &
  \multicolumn{1}{c|}{\cellcolor[HTML]{FFFFFF}\textcolor{red}\xmark} &
  \multicolumn{1}{c|}{\cellcolor[HTML]{FFFFFF}\textcolor{red}\xmark} &
  \multicolumn{1}{c|}{\cellcolor[HTML]{FFFFFF}\textcolor{green}\cmark} &
  \multicolumn{1}{c|}{\cellcolor[HTML]{FFFFFF}\textcolor{green}\cmark} &
  \multicolumn{1}{c|}{\cellcolor[HTML]{FFFFFF}\textcolor{red}\xmark} &
  \multicolumn{1}{c|}{\cellcolor[HTML]{FFFFFF}\textcolor{red}\xmark} &
  \multicolumn{1}{c|}{\cellcolor[HTML]{FFFFFF}\textcolor{red}\xmark} &
  \multicolumn{1}{c|}{\cellcolor[HTML]{FFFFFF}\textcolor{red}\xmark} &
  \multicolumn{1}{c|}{\cellcolor[HTML]{FFFFFF}\textcolor{red}\xmark} &
  \multicolumn{1}{c|}{\cellcolor[HTML]{FFFFFF}\textcolor{red}\xmark} &
  \textcolor{red}\xmark \\
\rowcolor[HTML]{EFEFEF} 
Tan et al. \cite{tan2022scientific} &
  \multicolumn{1}{c|}{\cellcolor[HTML]{EFEFEF}\textcolor{green}\cmark} &
  \multicolumn{1}{c|}{\cellcolor[HTML]{EFEFEF}\textcolor{red}\xmark} &
  \multicolumn{1}{c|}{\cellcolor[HTML]{EFEFEF}\textcolor{red}\xmark} &
  \multicolumn{1}{c|}{\cellcolor[HTML]{EFEFEF}\textcolor{green}\cmark} &
  \multicolumn{1}{c|}{\cellcolor[HTML]{EFEFEF}\textcolor{green}\cmark} &
  \multicolumn{1}{c|}{\cellcolor[HTML]{EFEFEF}\textcolor{red}\xmark} &
  \multicolumn{1}{c|}{\cellcolor[HTML]{EFEFEF}\textcolor{red}\xmark} &
  \multicolumn{1}{c|}{\cellcolor[HTML]{EFEFEF}\textcolor{green}\cmark} &
  \multicolumn{1}{c|}{\cellcolor[HTML]{EFEFEF}\textcolor{red}\xmark} &
  \multicolumn{1}{c|}{\cellcolor[HTML]{EFEFEF}\textcolor{red}\xmark} &
  \multicolumn{1}{c|}{\cellcolor[HTML]{EFEFEF}\textcolor{red}\xmark} &
  \multicolumn{1}{c|}{\cellcolor[HTML]{EFEFEF}\textcolor{green}\cmark} &
  \multicolumn{1}{c|}{\cellcolor[HTML]{EFEFEF}\textcolor{green}\cmark} &
  \multicolumn{1}{c|}{\cellcolor[HTML]{EFEFEF}\textcolor{red}\xmark} &
  \multicolumn{1}{c|}{\cellcolor[HTML]{EFEFEF}\textcolor{red}\xmark} &
  \multicolumn{1}{c|}{\cellcolor[HTML]{EFEFEF}\textcolor{red}\xmark} &
  \multicolumn{1}{c|}{\cellcolor[HTML]{EFEFEF}\textcolor{red}\xmark} &
  \multicolumn{1}{c|}{\cellcolor[HTML]{EFEFEF}\textcolor{red}\xmark} &
  \multicolumn{1}{c|}{\cellcolor[HTML]{EFEFEF}\textcolor{red}\xmark} &
  \textcolor{red}\xmark \\
\rowcolor[HTML]{FFFFFF} 
LineCap \cite{mahinpei2022linecap} &
  \multicolumn{1}{c|}{\cellcolor[HTML]{FFFFFF}\textcolor{green}\cmark} &
  \multicolumn{1}{c|}{\cellcolor[HTML]{FFFFFF}\textcolor{red}\xmark} &
  \multicolumn{1}{c|}{\cellcolor[HTML]{FFFFFF}\textcolor{red}\xmark} &
  \multicolumn{1}{c|}{\cellcolor[HTML]{FFFFFF}\textcolor{red}\xmark} &
  \multicolumn{1}{c|}{\cellcolor[HTML]{FFFFFF}\textcolor{green}\cmark} &
  \multicolumn{1}{c|}{\cellcolor[HTML]{FFFFFF}\textcolor{red}\xmark} &
  \multicolumn{1}{c|}{\cellcolor[HTML]{FFFFFF}\textcolor{red}\xmark} &
  \multicolumn{1}{c|}{\cellcolor[HTML]{FFFFFF}\textcolor{green}\cmark} &
  \multicolumn{1}{c|}{\cellcolor[HTML]{FFFFFF}\textcolor{red}\xmark} &
  \multicolumn{1}{c|}{\cellcolor[HTML]{FFFFFF}\textcolor{red}\xmark} &
  \multicolumn{1}{c|}{\cellcolor[HTML]{FFFFFF}\textcolor{red}\xmark} &
  \multicolumn{1}{c|}{\cellcolor[HTML]{FFFFFF}\textcolor{green}\cmark} &
  \multicolumn{1}{c|}{\cellcolor[HTML]{FFFFFF}\textcolor{green}\cmark} &
  \multicolumn{1}{c|}{\cellcolor[HTML]{FFFFFF}\textcolor{red}\xmark} &
  \multicolumn{1}{c|}{\cellcolor[HTML]{FFFFFF}\textcolor{red}\xmark} &
  \multicolumn{1}{c|}{\cellcolor[HTML]{FFFFFF}\textcolor{red}\xmark} &
  \multicolumn{1}{c|}{\cellcolor[HTML]{FFFFFF}\textcolor{red}\xmark} &
  \multicolumn{1}{c|}{\cellcolor[HTML]{FFFFFF}\textcolor{red}\xmark} &
  \multicolumn{1}{c|}{\cellcolor[HTML]{FFFFFF}\textcolor{red}\xmark} &
  \textcolor{red}\xmark \\
  \rowcolor[HTML]{EFEFEF} 
Intentable \cite{choi2022intentable} &
  \multicolumn{1}{c|}{\cellcolor[HTML]{EFEFEF}\textcolor{green}\cmark} &
  \multicolumn{1}{c|}{\cellcolor[HTML]{EFEFEF}\textcolor{red}\xmark} &
  \multicolumn{1}{c|}{\cellcolor[HTML]{EFEFEF}\textcolor{red}\xmark} &
  \multicolumn{1}{c|}{\cellcolor[HTML]{EFEFEF}\textcolor{green}\cmark} &
  \multicolumn{1}{c|}{\cellcolor[HTML]{EFEFEF}\textcolor{green}\cmark} &
  \multicolumn{1}{c|}{\cellcolor[HTML]{EFEFEF}\textcolor{green}\cmark} &
  \multicolumn{1}{c|}{\cellcolor[HTML]{EFEFEF}\textcolor{red}\xmark} &
  \multicolumn{1}{c|}{\cellcolor[HTML]{EFEFEF}\textcolor{green}\cmark} &
  \multicolumn{1}{c|}{\cellcolor[HTML]{EFEFEF}\textcolor{red}\xmark} &
  \multicolumn{1}{c|}{\cellcolor[HTML]{EFEFEF}\textcolor{red}\xmark} &
  \multicolumn{1}{c|}{\cellcolor[HTML]{EFEFEF}\textcolor{red}\xmark} &
  \multicolumn{1}{c|}{\cellcolor[HTML]{EFEFEF}\textcolor{green}\cmark} &
  \multicolumn{1}{c|}{\cellcolor[HTML]{EFEFEF}\textcolor{green}\cmark} &
  \multicolumn{1}{c|}{\cellcolor[HTML]{EFEFEF}\textcolor{red}\xmark} &
  \multicolumn{1}{c|}{\cellcolor[HTML]{EFEFEF}\textcolor{red}\xmark} &
  \multicolumn{1}{c|}{\cellcolor[HTML]{EFEFEF}\textcolor{red}\xmark} &
  \multicolumn{1}{c|}{\cellcolor[HTML]{EFEFEF}\textcolor{red}\xmark} &
  \multicolumn{1}{c|}{\cellcolor[HTML]{EFEFEF}\textcolor{red}\xmark} &
  \multicolumn{1}{c|}{\cellcolor[HTML]{EFEFEF}\textcolor{red}\xmark} &
  \textcolor{red}\xmark \\
\rowcolor[HTML]{FFFFFF} 
VoxLens \cite{sharif2022voxlens} &
  \multicolumn{1}{c|}{\cellcolor[HTML]{FFFFFF}\textcolor{green}\cmark} &
  \multicolumn{1}{c|}{\cellcolor[HTML]{FFFFFF}\textcolor{green}\cmark} &
  \multicolumn{1}{c|}{\cellcolor[HTML]{FFFFFF}\textcolor{red}\xmark} &
  \multicolumn{1}{c|}{\cellcolor[HTML]{FFFFFF}\textcolor{red}\xmark} &
  \multicolumn{1}{c|}{\cellcolor[HTML]{FFFFFF}\textcolor{green}\cmark} &
  \multicolumn{1}{c|}{\cellcolor[HTML]{FFFFFF}\textcolor{red}\xmark} &
  \multicolumn{1}{c|}{\cellcolor[HTML]{FFFFFF}\textcolor{green}\cmark} &
  \multicolumn{1}{c|}{\cellcolor[HTML]{FFFFFF}\textcolor{green}\cmark} &
  \multicolumn{1}{c|}{\cellcolor[HTML]{FFFFFF}\textcolor{red}\xmark} &
  \multicolumn{1}{c|}{\cellcolor[HTML]{FFFFFF}\textcolor{green}\cmark} &
  \multicolumn{1}{c|}{\cellcolor[HTML]{FFFFFF}\textcolor{green}\cmark} &
  \multicolumn{1}{c|}{\cellcolor[HTML]{FFFFFF}\textcolor{red}\xmark} &
  \multicolumn{1}{c|}{\cellcolor[HTML]{FFFFFF}\textcolor{green}\cmark} &
  \multicolumn{1}{c|}{\cellcolor[HTML]{FFFFFF}\textcolor{red}\xmark} &
  \multicolumn{1}{c|}{\cellcolor[HTML]{FFFFFF}\textcolor{red}\xmark} &
  \multicolumn{1}{c|}{\cellcolor[HTML]{FFFFFF}\textcolor{red}\xmark} &
  \multicolumn{1}{c|}{\cellcolor[HTML]{FFFFFF}\textcolor{red}\xmark} &
  \multicolumn{1}{c|}{\cellcolor[HTML]{FFFFFF}\textcolor{red}\xmark} &
  \multicolumn{1}{c|}{\cellcolor[HTML]{FFFFFF}\textcolor{red}\xmark} &
  \textcolor{red}\xmark \\
\rowcolor[HTML]{EFEFEF} 
Azimuth \cite{srinivasan2023azimuth} &
  \multicolumn{1}{c|}{\cellcolor[HTML]{EFEFEF}\textcolor{green}\cmark} &
  \multicolumn{1}{c|}{\cellcolor[HTML]{EFEFEF}\textcolor{red}\xmark} &
  \multicolumn{1}{c|}{\cellcolor[HTML]{EFEFEF}\textcolor{red}\xmark} &
  \multicolumn{1}{c|}{\cellcolor[HTML]{EFEFEF}\textcolor{green}\cmark} &
  \multicolumn{1}{c|}{\cellcolor[HTML]{EFEFEF}\textcolor{red}\xmark} &
  \multicolumn{1}{c|}{\cellcolor[HTML]{EFEFEF}\textcolor{red}\xmark} &
  \multicolumn{1}{c|}{\cellcolor[HTML]{EFEFEF}\textcolor{red}\xmark} &
  \multicolumn{1}{c|}{\cellcolor[HTML]{EFEFEF}\textcolor{green}\cmark} &
  \multicolumn{1}{c|}{\cellcolor[HTML]{EFEFEF}\textcolor{green}\cmark} &
  \multicolumn{1}{c|}{\cellcolor[HTML]{EFEFEF}\textcolor{red}\xmark} &
  \multicolumn{1}{c|}{\cellcolor[HTML]{EFEFEF}\textcolor{green}\cmark} &
  \multicolumn{1}{c|}{\cellcolor[HTML]{EFEFEF}\textcolor{red}\xmark} &
  \multicolumn{1}{c|}{\cellcolor[HTML]{EFEFEF}\textcolor{green}\cmark} &
  \multicolumn{1}{c|}{\cellcolor[HTML]{EFEFEF}\textcolor{red}\xmark} &
  \multicolumn{1}{c|}{\cellcolor[HTML]{EFEFEF}\textcolor{red}\xmark} &
  \multicolumn{1}{c|}{\cellcolor[HTML]{EFEFEF}\textcolor{red}\xmark} &
  \multicolumn{1}{c|}{\cellcolor[HTML]{EFEFEF}\textcolor{red}\xmark} &
  \multicolumn{1}{c|}{\cellcolor[HTML]{EFEFEF}\textcolor{red}\xmark} &
  \multicolumn{1}{c|}{\cellcolor[HTML]{EFEFEF}\textcolor{red}\xmark} &
  \textcolor{red}\xmark \\
\rowcolor[HTML]{FFFFFF} 
ChartSumm \cite{rahman2023chartsumm} &
  \multicolumn{1}{c|}{\cellcolor[HTML]{FFFFFF}\textcolor{green}\cmark} &
  \multicolumn{1}{c|}{\cellcolor[HTML]{FFFFFF}\textcolor{red}\xmark} &
  \multicolumn{1}{c|}{\cellcolor[HTML]{FFFFFF}\textcolor{red}\xmark} &
  \multicolumn{1}{c|}{\cellcolor[HTML]{FFFFFF}\textcolor{green}\cmark} &
  \multicolumn{1}{c|}{\cellcolor[HTML]{FFFFFF}\textcolor{red}\xmark} &
  \multicolumn{1}{c|}{\cellcolor[HTML]{FFFFFF}\textcolor{green}\cmark} &
  \multicolumn{1}{c|}{\cellcolor[HTML]{FFFFFF}\textcolor{red}\xmark} &
  \multicolumn{1}{c|}{\cellcolor[HTML]{FFFFFF}\textcolor{green}\cmark} &
  \multicolumn{1}{c|}{\cellcolor[HTML]{FFFFFF}\textcolor{red}\xmark} &
  \multicolumn{1}{c|}{\cellcolor[HTML]{FFFFFF}\textcolor{red}\xmark} &
  \multicolumn{1}{c|}{\cellcolor[HTML]{FFFFFF}\textcolor{red}\xmark} &
  \multicolumn{1}{c|}{\cellcolor[HTML]{FFFFFF}\textcolor{green}\cmark} &
  \multicolumn{1}{c|}{\cellcolor[HTML]{FFFFFF}\textcolor{green}\cmark} &
  \multicolumn{1}{c|}{\cellcolor[HTML]{FFFFFF}\textcolor{red}\xmark} &
  \multicolumn{1}{c|}{\cellcolor[HTML]{FFFFFF}\textcolor{red}\xmark} &
  \multicolumn{1}{c|}{\cellcolor[HTML]{FFFFFF}\textcolor{red}\xmark} &
  \multicolumn{1}{c|}{\cellcolor[HTML]{FFFFFF}\textcolor{red}\xmark} &
  \multicolumn{1}{c|}{\cellcolor[HTML]{FFFFFF}\textcolor{red}\xmark} &
  \multicolumn{1}{c|}{\cellcolor[HTML]{FFFFFF}\textcolor{red}\xmark} &
  \textcolor{red}\xmark \\
\rowcolor[HTML]{EFEFEF} 
VisText \cite{tang2023vistext} &
  \multicolumn{1}{c|}{\cellcolor[HTML]{EFEFEF}\textcolor{green}\cmark} &
  \multicolumn{1}{c|}{\cellcolor[HTML]{EFEFEF}\textcolor{red}\xmark} &
  \multicolumn{1}{c|}{\cellcolor[HTML]{EFEFEF}\textcolor{red}\xmark} &
  \multicolumn{1}{c|}{\cellcolor[HTML]{EFEFEF}\textcolor{green}\cmark} &
  \multicolumn{1}{c|}{\cellcolor[HTML]{EFEFEF}\textcolor{green}\cmark} &
  \multicolumn{1}{c|}{\cellcolor[HTML]{EFEFEF}\textcolor{green}\cmark} &
  \multicolumn{1}{c|}{\cellcolor[HTML]{EFEFEF}\textcolor{red}\xmark} &
  \multicolumn{1}{c|}{\cellcolor[HTML]{EFEFEF}\textcolor{green}\cmark} &
  \multicolumn{1}{c|}{\cellcolor[HTML]{EFEFEF}\textcolor{red}\xmark} &
  \multicolumn{1}{c|}{\cellcolor[HTML]{EFEFEF}\textcolor{red}\xmark} &
  \multicolumn{1}{c|}{\cellcolor[HTML]{EFEFEF}\textcolor{red}\xmark} &
  \multicolumn{1}{c|}{\cellcolor[HTML]{EFEFEF}\textcolor{green}\cmark} &
  \multicolumn{1}{c|}{\cellcolor[HTML]{EFEFEF}\textcolor{green}\cmark} &
  \multicolumn{1}{c|}{\cellcolor[HTML]{EFEFEF}\textcolor{red}\xmark} &
  \multicolumn{1}{c|}{\cellcolor[HTML]{EFEFEF}\textcolor{red}\xmark} &
  \multicolumn{1}{c|}{\cellcolor[HTML]{EFEFEF}\textcolor{red}\xmark} &
  \multicolumn{1}{c|}{\cellcolor[HTML]{EFEFEF}\textcolor{red}\xmark} &
  \multicolumn{1}{c|}{\cellcolor[HTML]{EFEFEF}\textcolor{red}\xmark} &
  \multicolumn{1}{c|}{\cellcolor[HTML]{EFEFEF}\textcolor{red}\xmark} &
  \textcolor{red}\xmark \\
\rowcolor[HTML]{FFFFFF} 
UniChart \cite{masry2023unichart} &
  \multicolumn{1}{c|}{\cellcolor[HTML]{FFFFFF}\textcolor{green}\cmark} &
  \multicolumn{1}{c|}{\cellcolor[HTML]{FFFFFF}\textcolor{green}\cmark} &
  \multicolumn{1}{c|}{\cellcolor[HTML]{FFFFFF}\textcolor{red}\xmark} &
  \multicolumn{1}{c|}{\cellcolor[HTML]{FFFFFF}\textcolor{green}\cmark} &
  \multicolumn{1}{c|}{\cellcolor[HTML]{FFFFFF}\textcolor{green}\cmark} &
  \multicolumn{1}{c|}{\cellcolor[HTML]{FFFFFF}\textcolor{green}\cmark} &
  \multicolumn{1}{c|}{\cellcolor[HTML]{FFFFFF}\textcolor{red}\xmark} &
  \multicolumn{1}{c|}{\cellcolor[HTML]{FFFFFF}\textcolor{green}\cmark} &
  \multicolumn{1}{c|}{\cellcolor[HTML]{FFFFFF}\textcolor{red}\xmark} &
  \multicolumn{1}{c|}{\cellcolor[HTML]{FFFFFF}\textcolor{red}\xmark} &
  \multicolumn{1}{c|}{\cellcolor[HTML]{FFFFFF}\textcolor{red}\xmark} &
  \multicolumn{1}{c|}{\cellcolor[HTML]{FFFFFF}\textcolor{green}\cmark} &
  \multicolumn{1}{c|}{\cellcolor[HTML]{FFFFFF}\textcolor{green}\cmark} &
  \multicolumn{1}{c|}{\cellcolor[HTML]{FFFFFF}\textcolor{red}\xmark} &
  \multicolumn{1}{c|}{\cellcolor[HTML]{FFFFFF}\textcolor{red}\xmark} &
  \multicolumn{1}{c|}{\cellcolor[HTML]{FFFFFF}\textcolor{red}\xmark} &
  \multicolumn{1}{c|}{\cellcolor[HTML]{FFFFFF}\textcolor{red}\xmark} &
  \multicolumn{1}{c|}{\cellcolor[HTML]{FFFFFF}\textcolor{red}\xmark} &
  \multicolumn{1}{c|}{\cellcolor[HTML]{FFFFFF}\textcolor{red}\xmark} &
  \textcolor{red}\xmark \\
  \rowcolor[HTML]{EFEFEF} 
AutoTitle \cite{liu2023autotitle} &
  \multicolumn{1}{c|}{\cellcolor[HTML]{EFEFEF}\textcolor{green}\cmark} &
  \multicolumn{1}{c|}{\cellcolor[HTML]{EFEFEF}\textcolor{red}\xmark} &
  \multicolumn{1}{c|}{\cellcolor[HTML]{EFEFEF}\textcolor{red}\xmark} &
  \multicolumn{1}{c|}{\cellcolor[HTML]{EFEFEF}\textcolor{red}\xmark} &
  \multicolumn{1}{c|}{\cellcolor[HTML]{EFEFEF}\textcolor{green}\cmark} &
  \multicolumn{1}{c|}{\cellcolor[HTML]{EFEFEF}\textcolor{green}\cmark} &
  \multicolumn{1}{c|}{\cellcolor[HTML]{EFEFEF}\textcolor{red}\xmark} &
  \multicolumn{1}{c|}{\cellcolor[HTML]{EFEFEF}\textcolor{green}\cmark} &
  \multicolumn{1}{c|}{\cellcolor[HTML]{EFEFEF}\textcolor{red}\xmark} &
  \multicolumn{1}{c|}{\cellcolor[HTML]{EFEFEF}\textcolor{red}\xmark} &
  \multicolumn{1}{c|}{\cellcolor[HTML]{EFEFEF}\textcolor{red}\xmark} &
  \multicolumn{1}{c|}{\cellcolor[HTML]{EFEFEF}\textcolor{green}\cmark} &
  \multicolumn{1}{c|}{\cellcolor[HTML]{EFEFEF}\textcolor{green}\cmark} &
  \multicolumn{1}{c|}{\cellcolor[HTML]{EFEFEF}\textcolor{red}\xmark} &
  \multicolumn{1}{c|}{\cellcolor[HTML]{EFEFEF}\textcolor{red}\xmark} &
  \multicolumn{1}{c|}{\cellcolor[HTML]{EFEFEF}\textcolor{red}\xmark} &
  \multicolumn{1}{c|}{\cellcolor[HTML]{EFEFEF}\textcolor{red}\xmark} &
  \multicolumn{1}{c|}{\cellcolor[HTML]{EFEFEF}\textcolor{red}\xmark} &
  \multicolumn{1}{c|}{\cellcolor[HTML]{EFEFEF}\textcolor{red}\xmark} &
  \textcolor{red}\xmark
  \\ \hline 
\rowcolor[HTML]{FFFFFF} 
Hoque et al. \cite{hoque2018applying-pragmatics} &
  \multicolumn{1}{c|}{\cellcolor[HTML]{FFFFFF}\textcolor{red}\xmark} &
  \multicolumn{1}{c|}{\cellcolor[HTML]{FFFFFF}\textcolor{green}\cmark} &
  \multicolumn{1}{c|}{\cellcolor[HTML]{FFFFFF}\textcolor{red}\xmark} &
  \multicolumn{1}{c|}{\cellcolor[HTML]{FFFFFF}\textcolor{green}\cmark} &
  \multicolumn{1}{c|}{\cellcolor[HTML]{FFFFFF}\textcolor{red}\xmark} &
  \multicolumn{1}{c|}{\cellcolor[HTML]{FFFFFF}\textcolor{red}\xmark} &
  \multicolumn{1}{c|}{\cellcolor[HTML]{FFFFFF}\textcolor{green}\cmark} &
  \multicolumn{1}{c|}{\cellcolor[HTML]{FFFFFF}\textcolor{red}\xmark} &
  \multicolumn{1}{c|}{\cellcolor[HTML]{FFFFFF}\textcolor{red}\xmark} &
  \multicolumn{1}{c|}{\cellcolor[HTML]{FFFFFF}\textcolor{green}\cmark} &
  \multicolumn{1}{c|}{\cellcolor[HTML]{FFFFFF}\textcolor{green}\cmark} &
  \multicolumn{1}{c|}{\cellcolor[HTML]{FFFFFF}\textcolor{red}\xmark} &
  \multicolumn{1}{c|}{\cellcolor[HTML]{FFFFFF}\textcolor{green}\cmark} &
  \multicolumn{1}{c|}{\cellcolor[HTML]{FFFFFF}\textcolor{red}\xmark} &
  \multicolumn{1}{c|}{\cellcolor[HTML]{FFFFFF}\textcolor{red}\xmark} &
  \multicolumn{1}{c|}{\cellcolor[HTML]{FFFFFF}\textcolor{red}\xmark} &
  \multicolumn{1}{c|}{\cellcolor[HTML]{FFFFFF}\textcolor{red}\xmark} &
  \multicolumn{1}{c|}{\cellcolor[HTML]{FFFFFF}\textcolor{red}\xmark} &
  \multicolumn{1}{c|}{\cellcolor[HTML]{FFFFFF}\textcolor{red}\xmark} &
  \textcolor{red}\xmark \\
\rowcolor[HTML]{EFEFEF} 
Kim et al. \cite{kim2020answering} &
  \multicolumn{1}{c|}{\cellcolor[HTML]{EFEFEF}\textcolor{red}\xmark} &
  \multicolumn{1}{c|}{\cellcolor[HTML]{EFEFEF}\textcolor{green}\cmark} &
  \multicolumn{1}{c|}{\cellcolor[HTML]{EFEFEF}\textcolor{red}\xmark} &
  \multicolumn{1}{c|}{\cellcolor[HTML]{EFEFEF}\textcolor{green}\cmark} &
  \multicolumn{1}{c|}{\cellcolor[HTML]{EFEFEF}\textcolor{green}\cmark} &
  \multicolumn{1}{c|}{\cellcolor[HTML]{EFEFEF}\textcolor{red}\xmark} &
  \multicolumn{1}{c|}{\cellcolor[HTML]{EFEFEF}\textcolor{red}\xmark} &
  \multicolumn{1}{c|}{\cellcolor[HTML]{EFEFEF}\textcolor{green}\cmark} &
  \multicolumn{1}{c|}{\cellcolor[HTML]{EFEFEF}\textcolor{red}\xmark} &
  \multicolumn{1}{c|}{\cellcolor[HTML]{EFEFEF}\textcolor{red}\xmark} &
  \multicolumn{1}{c|}{\cellcolor[HTML]{EFEFEF}\textcolor{red}\xmark} &
  \multicolumn{1}{c|}{\cellcolor[HTML]{EFEFEF}\textcolor{green}\cmark} &
  \multicolumn{1}{c|}{\cellcolor[HTML]{EFEFEF}\textcolor{green}\cmark} &
  \multicolumn{1}{c|}{\cellcolor[HTML]{EFEFEF}\textcolor{red}\xmark} &
  \multicolumn{1}{c|}{\cellcolor[HTML]{EFEFEF}\textcolor{red}\xmark} &
  \multicolumn{1}{c|}{\cellcolor[HTML]{EFEFEF}\textcolor{red}\xmark} &
  \multicolumn{1}{c|}{\cellcolor[HTML]{EFEFEF}\textcolor{red}\xmark} &
  \multicolumn{1}{c|}{\cellcolor[HTML]{EFEFEF}\textcolor{red}\xmark} &
  \multicolumn{1}{c|}{\cellcolor[HTML]{EFEFEF}\textcolor{red}\xmark} &
  \textcolor{red}\xmark \\
\rowcolor[HTML]{FFFFFF} 
OpenCQA \cite{kantharaj-etal-2022-opencqa} &
  \multicolumn{1}{c|}{\cellcolor[HTML]{FFFFFF}\textcolor{red}\xmark} &
  \multicolumn{1}{c|}{\cellcolor[HTML]{FFFFFF}\textcolor{green}\cmark} &
  \multicolumn{1}{c|}{\cellcolor[HTML]{FFFFFF}\textcolor{red}\xmark} &
  \multicolumn{1}{c|}{\cellcolor[HTML]{FFFFFF}\textcolor{green}\cmark} &
  \multicolumn{1}{c|}{\cellcolor[HTML]{FFFFFF}\textcolor{green}\cmark} &
  \multicolumn{1}{c|}{\cellcolor[HTML]{FFFFFF}\textcolor{green}\cmark} &
  \multicolumn{1}{c|}{\cellcolor[HTML]{FFFFFF}\textcolor{red}\xmark} &
  \multicolumn{1}{c|}{\cellcolor[HTML]{FFFFFF}\textcolor{green}\cmark} &
  \multicolumn{1}{c|}{\cellcolor[HTML]{FFFFFF}\textcolor{red}\xmark} &
  \multicolumn{1}{c|}{\cellcolor[HTML]{FFFFFF}\textcolor{red}\xmark} &
  \multicolumn{1}{c|}{\cellcolor[HTML]{FFFFFF}\textcolor{red}\xmark} &
  \multicolumn{1}{c|}{\cellcolor[HTML]{FFFFFF}\textcolor{green}\cmark} &
  \multicolumn{1}{c|}{\cellcolor[HTML]{FFFFFF}\textcolor{green}\cmark} &
  \multicolumn{1}{c|}{\cellcolor[HTML]{FFFFFF}\textcolor{red}\xmark} &
  \multicolumn{1}{c|}{\cellcolor[HTML]{FFFFFF}\textcolor{red}\xmark} &
  \multicolumn{1}{c|}{\cellcolor[HTML]{FFFFFF}\textcolor{red}\xmark} &
  \multicolumn{1}{c|}{\cellcolor[HTML]{FFFFFF}\textcolor{red}\xmark} &
  \multicolumn{1}{c|}{\cellcolor[HTML]{FFFFFF}\textcolor{red}\xmark} &
  \multicolumn{1}{c|}{\cellcolor[HTML]{FFFFFF}\textcolor{red}\xmark} &
  \textcolor{red}\xmark \\
% \rowcolor[HTML]{EFEFEF} 
% ChartQA \cite{masry-etal-2022-chartqa} &
%   \multicolumn{1}{c|}{\cellcolor[HTML]{EFEFEF}\textcolor{red}\xmark} &
%   \multicolumn{1}{c|}{\cellcolor[HTML]{EFEFEF}\textcolor{green}\cmark} &
%   \multicolumn{1}{c|}{\cellcolor[HTML]{EFEFEF}\textcolor{red}\xmark} &
%   \multicolumn{1}{c|}{\cellcolor[HTML]{EFEFEF}\textcolor{green}\cmark} &
%   \multicolumn{1}{c|}{\cellcolor[HTML]{EFEFEF}\textcolor{red}\xmark} &
%   \multicolumn{1}{c|}{\cellcolor[HTML]{EFEFEF}\textcolor{green}\cmark} &
%   \multicolumn{1}{c|}{\cellcolor[HTML]{EFEFEF}\textcolor{red}\xmark} &
%   \multicolumn{1}{c|}{\cellcolor[HTML]{EFEFEF}\textcolor{green}\cmark} &
%   \multicolumn{1}{c|}{\cellcolor[HTML]{EFEFEF}\textcolor{red}\xmark} &
%   \multicolumn{1}{c|}{\cellcolor[HTML]{EFEFEF}\textcolor{red}\xmark} &
%   \multicolumn{1}{c|}{\cellcolor[HTML]{EFEFEF}\textcolor{red}\xmark} &
%   \multicolumn{1}{c|}{\cellcolor[HTML]{EFEFEF}\textcolor{green}\cmark} &
%   \multicolumn{1}{c|}{\cellcolor[HTML]{EFEFEF}\textcolor{green}\cmark} &
%   \multicolumn{1}{c|}{\cellcolor[HTML]{EFEFEF}\textcolor{red}\xmark} &
%   \multicolumn{1}{c|}{\cellcolor[HTML]{EFEFEF}\textcolor{red}\xmark} &
%   \multicolumn{1}{c|}{\cellcolor[HTML]{EFEFEF}\textcolor{red}\xmark} &
%   \multicolumn{1}{c|}{\cellcolor[HTML]{EFEFEF}\textcolor{red}\xmark} &
%   \multicolumn{1}{c|}{\cellcolor[HTML]{EFEFEF}\textcolor{red}\xmark} &
%   \multicolumn{1}{c|}{\cellcolor[HTML]{EFEFEF}\textcolor{red}\xmark} &
%   \textcolor{red}\xmark \\ 
\rowcolor[HTML]{EFEFEF} 
SciGraphQA \cite{li2023scigraphqa} &
  \multicolumn{1}{c|}{\cellcolor[HTML]{EFEFEF}\textcolor{red}\xmark} &
  \multicolumn{1}{c|}{\cellcolor[HTML]{EFEFEF}\textcolor{green}\cmark} &
  \multicolumn{1}{c|}{\cellcolor[HTML]{EFEFEF}\textcolor{red}\xmark} &
  \multicolumn{1}{c|}{\cellcolor[HTML]{EFEFEF}\textcolor{green}\cmark} &
  \multicolumn{1}{c|}{\cellcolor[HTML]{EFEFEF}\textcolor{red}\xmark} &
  \multicolumn{1}{c|}{\cellcolor[HTML]{EFEFEF}\textcolor{red}\xmark} &
  \multicolumn{1}{c|}{\cellcolor[HTML]{EFEFEF}\textcolor{red}\xmark} &
  \multicolumn{1}{c|}{\cellcolor[HTML]{EFEFEF}\textcolor{green}\cmark} &
  \multicolumn{1}{c|}{\cellcolor[HTML]{EFEFEF}\textcolor{red}\xmark} &
  \multicolumn{1}{c|}{\cellcolor[HTML]{EFEFEF}\textcolor{red}\xmark} &
  \multicolumn{1}{c|}{\cellcolor[HTML]{EFEFEF}\textcolor{red}\xmark} &
  \multicolumn{1}{c|}{\cellcolor[HTML]{EFEFEF}\textcolor{green}\cmark} &
  \multicolumn{1}{c|}{\cellcolor[HTML]{EFEFEF}\textcolor{green}\cmark} &
  \multicolumn{1}{c|}{\cellcolor[HTML]{EFEFEF}\textcolor{red}\xmark} &
  \multicolumn{1}{c|}{\cellcolor[HTML]{EFEFEF}\textcolor{red}\xmark} &
  \multicolumn{1}{c|}{\cellcolor[HTML]{EFEFEF}\textcolor{red}\xmark} &
  \multicolumn{1}{c|}{\cellcolor[HTML]{EFEFEF}\textcolor{red}\xmark} &
  \multicolumn{1}{c|}{\cellcolor[HTML]{EFEFEF}\textcolor{red}\xmark} &
  \multicolumn{1}{c|}{\cellcolor[HTML]{EFEFEF}\textcolor{red}\xmark} &
  \textcolor{red}\xmark \\
\rowcolor[HTML]{FFFFFF} 
Song et al. \cite{song2023marrying} &
  \multicolumn{1}{c|}{\cellcolor[HTML]{FFFFFF}\textcolor{red}\xmark} &
  \multicolumn{1}{c|}{\cellcolor[HTML]{FFFFFF}\textcolor{green}\cmark} &
  \multicolumn{1}{c|}{\cellcolor[HTML]{FFFFFF}\textcolor{red}\xmark} &
  \multicolumn{1}{c|}{\cellcolor[HTML]{FFFFFF}\textcolor{green}\cmark} &
  \multicolumn{1}{c|}{\cellcolor[HTML]{FFFFFF}\textcolor{red}\xmark} &
  \multicolumn{1}{c|}{\cellcolor[HTML]{FFFFFF}\textcolor{green}\cmark} &
  \multicolumn{1}{c|}{\cellcolor[HTML]{FFFFFF}\textcolor{red}\xmark} &
  \multicolumn{1}{c|}{\cellcolor[HTML]{FFFFFF}\textcolor{green}\cmark} &
  \multicolumn{1}{c|}{\cellcolor[HTML]{FFFFFF}\textcolor{red}\xmark} &
  \multicolumn{1}{c|}{\cellcolor[HTML]{FFFFFF}\textcolor{red}\xmark} &
  \multicolumn{1}{c|}{\cellcolor[HTML]{FFFFFF}\textcolor{red}\xmark} &
  \multicolumn{1}{c|}{\cellcolor[HTML]{FFFFFF}\textcolor{green}\cmark} &
  \multicolumn{1}{c|}{\cellcolor[HTML]{FFFFFF}\textcolor{green}\cmark} &
  \multicolumn{1}{c|}{\cellcolor[HTML]{FFFFFF}\textcolor{red}\xmark} &
  \multicolumn{1}{c|}{\cellcolor[HTML]{FFFFFF}\textcolor{red}\xmark} &
  \multicolumn{1}{c|}{\cellcolor[HTML]{FFFFFF}\textcolor{red}\xmark} &
  \multicolumn{1}{c|}{\cellcolor[HTML]{FFFFFF}\textcolor{red}\xmark} &
  \multicolumn{1}{c|}{\cellcolor[HTML]{FFFFFF}\textcolor{red}\xmark} &
  \multicolumn{1}{c|}{\cellcolor[HTML]{FFFFFF}\textcolor{red}\xmark} &
  \textcolor{red}\xmark \\ \hline
\rowcolor[HTML]{EFEFEF} 
Contextifier \cite{hullman2013contextifier} &
  \multicolumn{1}{c|}{\cellcolor[HTML]{EFEFEF}\textcolor{red}\xmark} &
  \multicolumn{1}{c|}{\cellcolor[HTML]{EFEFEF}\textcolor{red}\xmark} &
  \multicolumn{1}{c|}{\cellcolor[HTML]{EFEFEF}\textcolor{green}\cmark} &
  \multicolumn{1}{c|}{\cellcolor[HTML]{EFEFEF}\textcolor{green}\cmark} &
  \multicolumn{1}{c|}{\cellcolor[HTML]{EFEFEF}\textcolor{red}\xmark} &
  \multicolumn{1}{c|}{\cellcolor[HTML]{EFEFEF}\textcolor{red}\xmark} &
  \multicolumn{1}{c|}{\cellcolor[HTML]{EFEFEF}\textcolor{red}\xmark} &
  \multicolumn{1}{c|}{\cellcolor[HTML]{EFEFEF}\textcolor{red}\xmark} &
  \multicolumn{1}{c|}{\cellcolor[HTML]{EFEFEF}\textcolor{red}\xmark} &
  \multicolumn{1}{c|}{\cellcolor[HTML]{EFEFEF}\textcolor{green}\cmark} &
  \multicolumn{1}{c|}{\cellcolor[HTML]{EFEFEF}\textcolor{green}\cmark} &
  \multicolumn{1}{c|}{\cellcolor[HTML]{EFEFEF}\textcolor{red}\xmark} &
  \multicolumn{1}{c|}{\cellcolor[HTML]{EFEFEF}\textcolor{red}\xmark} &
  \multicolumn{1}{c|}{\cellcolor[HTML]{EFEFEF}\textcolor{green}\cmark} &
  \multicolumn{1}{c|}{\cellcolor[HTML]{EFEFEF}\textcolor{red}\xmark} &
  \multicolumn{1}{c|}{\cellcolor[HTML]{EFEFEF}\textcolor{red}\xmark} &
  \multicolumn{1}{c|}{\cellcolor[HTML]{EFEFEF}\textcolor{red}\xmark} &
  \multicolumn{1}{c|}{\cellcolor[HTML]{EFEFEF}\textcolor{red}\xmark} &
  \multicolumn{1}{c|}{\cellcolor[HTML]{EFEFEF}\textcolor{green}\cmark} &
  \textcolor{red}\xmark \\
\rowcolor[HTML]{FFFFFF} 
Metoyer et al. \cite{metoyer2018coupling} &
  \multicolumn{1}{c|}{\cellcolor[HTML]{FFFFFF}\textcolor{red}\xmark} &
  \multicolumn{1}{c|}{\cellcolor[HTML]{FFFFFF}\textcolor{red}\xmark} &
  \multicolumn{1}{c|}{\cellcolor[HTML]{FFFFFF}\textcolor{green}\cmark} &
  \multicolumn{1}{c|}{\cellcolor[HTML]{FFFFFF}\textcolor{green}\cmark} &
  \multicolumn{1}{c|}{\cellcolor[HTML]{FFFFFF}\textcolor{green}\cmark} &
  \multicolumn{1}{c|}{\cellcolor[HTML]{FFFFFF}\textcolor{red}\xmark} &
  \multicolumn{1}{c|}{\cellcolor[HTML]{FFFFFF}\textcolor{red}\xmark} &
  \multicolumn{1}{c|}{\cellcolor[HTML]{FFFFFF}\textcolor{red}\xmark} &
  \multicolumn{1}{c|}{\cellcolor[HTML]{FFFFFF}\textcolor{red}\xmark} &
  \multicolumn{1}{c|}{\cellcolor[HTML]{FFFFFF}\textcolor{green}\cmark} &
  \multicolumn{1}{c|}{\cellcolor[HTML]{FFFFFF}\textcolor{red}\xmark} &
  \multicolumn{1}{c|}{\cellcolor[HTML]{FFFFFF}\textcolor{green}\cmark} &
  \multicolumn{1}{c|}{\cellcolor[HTML]{FFFFFF}\textcolor{red}\xmark} &
  \multicolumn{1}{c|}{\cellcolor[HTML]{FFFFFF}\textcolor{red}\xmark} &
  \multicolumn{1}{c|}{\cellcolor[HTML]{FFFFFF}\textcolor{red}\xmark} &
  \multicolumn{1}{c|}{\cellcolor[HTML]{FFFFFF}\textcolor{green}\cmark} &
  \multicolumn{1}{c|}{\cellcolor[HTML]{FFFFFF}\textcolor{red}\xmark} &
  \multicolumn{1}{c|}{\cellcolor[HTML]{FFFFFF}\textcolor{red}\xmark} &
  \multicolumn{1}{c|}{\cellcolor[HTML]{FFFFFF}\textcolor{green}\cmark} &
  \textcolor{red}\xmark \\
% \rowcolor[HTML]{EFEFEF} 
% Brehmer et al. \cite{brehmer2019timeline} &
%   \multicolumn{1}{c|}{\cellcolor[HTML]{EFEFEF}\textcolor{red}\xmark} &
%   \multicolumn{1}{c|}{\cellcolor[HTML]{EFEFEF}\textcolor{red}\xmark} &
%   \multicolumn{1}{c|}{\cellcolor[HTML]{EFEFEF}\textcolor{green}\cmark} &
%   \multicolumn{1}{c|}{\cellcolor[HTML]{EFEFEF}\textcolor{red}\xmark} &
%   \multicolumn{1}{c|}{\cellcolor[HTML]{EFEFEF}\textcolor{red}\xmark} &
%   \multicolumn{1}{c|}{\cellcolor[HTML]{EFEFEF}\textcolor{green}\cmark} &
%   \multicolumn{1}{c|}{\cellcolor[HTML]{EFEFEF}\textcolor{red}\xmark} &
%   \multicolumn{1}{c|}{\cellcolor[HTML]{EFEFEF}\textcolor{red}\xmark} &
%   \multicolumn{1}{c|}{\cellcolor[HTML]{EFEFEF}\textcolor{red}\xmark} &
%   \multicolumn{1}{c|}{\cellcolor[HTML]{EFEFEF}\textcolor{green}\cmark} &
%   \multicolumn{1}{c|}{\cellcolor[HTML]{EFEFEF}\textcolor{green}\cmark} &
%   \multicolumn{1}{c|}{\cellcolor[HTML]{EFEFEF}\textcolor{red}\xmark} &
%   \multicolumn{1}{c|}{\cellcolor[HTML]{EFEFEF}\textcolor{red}\xmark} &
%   \multicolumn{1}{c|}{\cellcolor[HTML]{EFEFEF}\textcolor{red}\xmark} &
%   \multicolumn{1}{c|}{\cellcolor[HTML]{EFEFEF}\textcolor{red}\xmark} &
%   \multicolumn{1}{c|}{\cellcolor[HTML]{EFEFEF}\textcolor{green}\cmark} &
%   \multicolumn{1}{c|}{\cellcolor[HTML]{EFEFEF}\textcolor{green}\cmark} &
%   \multicolumn{1}{c|}{\cellcolor[HTML]{EFEFEF}\textcolor{red}\xmark} &
%   \multicolumn{1}{c|}{\cellcolor[HTML]{EFEFEF}\textcolor{green}\cmark} &
%   \textcolor{red}\xmark \\
\rowcolor[HTML]{EFEFEF} 
Text-to-Viz \cite{cui2020text-to-viz} &
  \multicolumn{1}{c|}{\cellcolor[HTML]{EFEFEF}\textcolor{red}\xmark} &
  \multicolumn{1}{c|}{\cellcolor[HTML]{EFEFEF}\textcolor{red}\xmark} &
  \multicolumn{1}{c|}{\cellcolor[HTML]{EFEFEF}\textcolor{green}\cmark} &
  \multicolumn{1}{c|}{\cellcolor[HTML]{EFEFEF}\textcolor{green}\cmark} &
  \multicolumn{1}{c|}{\cellcolor[HTML]{EFEFEF}\textcolor{red}\xmark} &
  \multicolumn{1}{c|}{\cellcolor[HTML]{EFEFEF}\textcolor{red}\xmark} &
  \multicolumn{1}{c|}{\cellcolor[HTML]{EFEFEF}\textcolor{red}\xmark} &
  \multicolumn{1}{c|}{\cellcolor[HTML]{EFEFEF}\textcolor{red}\xmark} &
  \multicolumn{1}{c|}{\cellcolor[HTML]{EFEFEF}\textcolor{red}\xmark} &
  \multicolumn{1}{c|}{\cellcolor[HTML]{EFEFEF}\textcolor{green}\cmark} &
  \multicolumn{1}{c|}{\cellcolor[HTML]{EFEFEF}\textcolor{green}\cmark} &
  \multicolumn{1}{c|}{\cellcolor[HTML]{EFEFEF}\textcolor{red}\xmark} &
  \multicolumn{1}{c|}{\cellcolor[HTML]{EFEFEF}\textcolor{red}\xmark} &
  \multicolumn{1}{c|}{\cellcolor[HTML]{EFEFEF}\textcolor{green}\cmark} &
  \multicolumn{1}{c|}{\cellcolor[HTML]{EFEFEF}\textcolor{red}\xmark} &
  \multicolumn{1}{c|}{\cellcolor[HTML]{EFEFEF}\textcolor{red}\xmark} &
  \multicolumn{1}{c|}{\cellcolor[HTML]{EFEFEF}\textcolor{green}\cmark} &
  \multicolumn{1}{c|}{\cellcolor[HTML]{EFEFEF}\textcolor{red}\xmark} &
  \multicolumn{1}{c|}{\cellcolor[HTML]{EFEFEF}\textcolor{green}\cmark} &
  \textcolor{red}\xmark \\
\rowcolor[HTML]{FFFFFF} 
Chen et al. \cite{chen2020towards} &
  \multicolumn{1}{c|}{\cellcolor[HTML]{FFFFFF}\textcolor{red}\xmark} &
  \multicolumn{1}{c|}{\cellcolor[HTML]{FFFFFF}\textcolor{red}\xmark} &
  \multicolumn{1}{c|}{\cellcolor[HTML]{FFFFFF}\textcolor{green}\cmark} &
  \multicolumn{1}{c|}{\cellcolor[HTML]{FFFFFF}\textcolor{red}\xmark} &
  \multicolumn{1}{c|}{\cellcolor[HTML]{FFFFFF}\textcolor{green}\cmark} &
  \multicolumn{1}{c|}{\cellcolor[HTML]{FFFFFF}\textcolor{red}\xmark} &
  \multicolumn{1}{c|}{\cellcolor[HTML]{FFFFFF}\textcolor{red}\xmark} &
  \multicolumn{1}{c|}{\cellcolor[HTML]{FFFFFF}\textcolor{red}\xmark} &
  \multicolumn{1}{c|}{\cellcolor[HTML]{FFFFFF}\textcolor{red}\xmark} &
  \multicolumn{1}{c|}{\cellcolor[HTML]{FFFFFF}\textcolor{green}\cmark} &
  \multicolumn{1}{c|}{\cellcolor[HTML]{FFFFFF}\textcolor{red}\xmark} &
  \multicolumn{1}{c|}{\cellcolor[HTML]{FFFFFF}\textcolor{green}\cmark} &
  \multicolumn{1}{c|}{\cellcolor[HTML]{FFFFFF}\textcolor{red}\xmark} &
  \multicolumn{1}{c|}{\cellcolor[HTML]{FFFFFF}\textcolor{green}\cmark} &
  \multicolumn{1}{c|}{\cellcolor[HTML]{FFFFFF}\textcolor{red}\xmark} &
  \multicolumn{1}{c|}{\cellcolor[HTML]{FFFFFF}\textcolor{red}\xmark} &
  \multicolumn{1}{c|}{\cellcolor[HTML]{FFFFFF}\textcolor{green}\cmark} &
  \multicolumn{1}{c|}{\cellcolor[HTML]{FFFFFF}\textcolor{red}\xmark} &
  \multicolumn{1}{c|}{\cellcolor[HTML]{FFFFFF}\textcolor{green}\cmark} &
  \textcolor{red}\xmark \\
\rowcolor[HTML]{EFEFEF} 
DataShot \cite{wang2020datashot} &
  \multicolumn{1}{c|}{\cellcolor[HTML]{EFEFEF}\textcolor{red}\xmark} &
  \multicolumn{1}{c|}{\cellcolor[HTML]{EFEFEF}\textcolor{red}\xmark} &
  \multicolumn{1}{c|}{\cellcolor[HTML]{EFEFEF}\textcolor{green}\cmark} &
  \multicolumn{1}{c|}{\cellcolor[HTML]{EFEFEF}\textcolor{red}\xmark} &
  \multicolumn{1}{c|}{\cellcolor[HTML]{EFEFEF}\textcolor{red}\xmark} &
  \multicolumn{1}{c|}{\cellcolor[HTML]{EFEFEF}\textcolor{green}\cmark} &
  \multicolumn{1}{c|}{\cellcolor[HTML]{EFEFEF}\textcolor{red}\xmark} &
  \multicolumn{1}{c|}{\cellcolor[HTML]{EFEFEF}\textcolor{red}\xmark} &
  \multicolumn{1}{c|}{\cellcolor[HTML]{EFEFEF}\textcolor{red}\xmark} &
  \multicolumn{1}{c|}{\cellcolor[HTML]{EFEFEF}\textcolor{green}\cmark} &
  \multicolumn{1}{c|}{\cellcolor[HTML]{EFEFEF}\textcolor{green}\cmark} &
  \multicolumn{1}{c|}{\cellcolor[HTML]{EFEFEF}\textcolor{red}\xmark} &
  \multicolumn{1}{c|}{\cellcolor[HTML]{EFEFEF}\textcolor{red}\xmark} &
  \multicolumn{1}{c|}{\cellcolor[HTML]{EFEFEF}\textcolor{green}\cmark} &
  \multicolumn{1}{c|}{\cellcolor[HTML]{EFEFEF}\textcolor{red}\xmark} &
  \multicolumn{1}{c|}{\cellcolor[HTML]{EFEFEF}\textcolor{red}\xmark} &
  \multicolumn{1}{c|}{\cellcolor[HTML]{EFEFEF}\textcolor{green}\cmark} &
  \multicolumn{1}{c|}{\cellcolor[HTML]{EFEFEF}\textcolor{red}\xmark} &
  \multicolumn{1}{c|}{\cellcolor[HTML]{EFEFEF}\textcolor{green}\cmark} &
  \textcolor{red}\xmark \\
% \rowcolor[HTML]{EFEFEF} 
% ADVISor \cite{liu2021advisor} &
%   \multicolumn{1}{c|}{\cellcolor[HTML]{EFEFEF}\textcolor{red}\xmark} &
%   \multicolumn{1}{c|}{\cellcolor[HTML]{EFEFEF}\textcolor{red}\xmark} &
%   \multicolumn{1}{c|}{\cellcolor[HTML]{EFEFEF}\textcolor{green}\cmark} &
%   \multicolumn{1}{c|}{\cellcolor[HTML]{EFEFEF}\textcolor{green}\cmark} &
%   \multicolumn{1}{c|}{\cellcolor[HTML]{EFEFEF}\textcolor{red}\xmark} &
%   \multicolumn{1}{c|}{\cellcolor[HTML]{EFEFEF}\textcolor{green}\cmark} &
%   \multicolumn{1}{c|}{\cellcolor[HTML]{EFEFEF}\textcolor{red}\xmark} &
%   \multicolumn{1}{c|}{\cellcolor[HTML]{EFEFEF}\textcolor{red}\xmark} &
%   \multicolumn{1}{c|}{\cellcolor[HTML]{EFEFEF}\textcolor{red}\xmark} &
%   \multicolumn{1}{c|}{\cellcolor[HTML]{EFEFEF}\textcolor{green}\cmark} &
%   \multicolumn{1}{c|}{\cellcolor[HTML]{EFEFEF}\textcolor{red}\xmark} &
%   \multicolumn{1}{c|}{\cellcolor[HTML]{EFEFEF}\textcolor{green}\cmark} &
%   \multicolumn{1}{c|}{\cellcolor[HTML]{EFEFEF}\textcolor{red}\xmark} &
%   \multicolumn{1}{c|}{\cellcolor[HTML]{EFEFEF}\textcolor{green}\cmark} &
%   \multicolumn{1}{c|}{\cellcolor[HTML]{EFEFEF}\textcolor{red}\xmark} &
%   \multicolumn{1}{c|}{\cellcolor[HTML]{EFEFEF}\textcolor{red}\xmark} &
%   \multicolumn{1}{c|}{\cellcolor[HTML]{EFEFEF}\textcolor{red}\xmark} &
%   \multicolumn{1}{c|}{\cellcolor[HTML]{EFEFEF}\textcolor{red}\xmark} &
%   \multicolumn{1}{c|}{\cellcolor[HTML]{EFEFEF}\textcolor{green}\cmark} &
%   \textcolor{red}\xmark \\
\rowcolor[HTML]{FFFFFF} 
Calliope \cite{shi2021calliope} &
  \multicolumn{1}{c|}{\cellcolor[HTML]{FFFFFF}\textcolor{red}\xmark} &
  \multicolumn{1}{c|}{\cellcolor[HTML]{FFFFFF}\textcolor{red}\xmark} &
  \multicolumn{1}{c|}{\cellcolor[HTML]{FFFFFF}\textcolor{green}\cmark} &
  \multicolumn{1}{c|}{\cellcolor[HTML]{FFFFFF}\textcolor{red}\xmark} &
  \multicolumn{1}{c|}{\cellcolor[HTML]{FFFFFF}\textcolor{red}\xmark} &
  \multicolumn{1}{c|}{\cellcolor[HTML]{FFFFFF}\textcolor{green}\cmark} &
  \multicolumn{1}{c|}{\cellcolor[HTML]{FFFFFF}\textcolor{red}\xmark} &
  \multicolumn{1}{c|}{\cellcolor[HTML]{FFFFFF}\textcolor{red}\xmark} &
  \multicolumn{1}{c|}{\cellcolor[HTML]{FFFFFF}\textcolor{red}\xmark} &
  \multicolumn{1}{c|}{\cellcolor[HTML]{FFFFFF}\textcolor{green}\cmark} &
  \multicolumn{1}{c|}{\cellcolor[HTML]{FFFFFF}\textcolor{green}\cmark} &
  \multicolumn{1}{c|}{\cellcolor[HTML]{FFFFFF}\textcolor{red}\xmark} &
  \multicolumn{1}{c|}{\cellcolor[HTML]{FFFFFF}\textcolor{red}\xmark} &
  \multicolumn{1}{c|}{\cellcolor[HTML]{FFFFFF}\textcolor{green}\cmark} &
  \multicolumn{1}{c|}{\cellcolor[HTML]{FFFFFF}\textcolor{red}\xmark} &
  \multicolumn{1}{c|}{\cellcolor[HTML]{FFFFFF}\textcolor{green}\cmark} &
  \multicolumn{1}{c|}{\cellcolor[HTML]{FFFFFF}\textcolor{green}\cmark} &
  \multicolumn{1}{c|}{\cellcolor[HTML]{FFFFFF}\textcolor{red}\xmark} &
  \multicolumn{1}{c|}{\cellcolor[HTML]{FFFFFF}\textcolor{green}\cmark} &
  \textcolor{red}\xmark \\
\rowcolor[HTML]{EFEFEF} 
Socrates \cite{wu2023socrates} &
  \multicolumn{1}{c|}{\cellcolor[HTML]{EFEFEF}\textcolor{red}\xmark} &
  \multicolumn{1}{c|}{\cellcolor[HTML]{EFEFEF}\textcolor{red}\xmark} &
  \multicolumn{1}{c|}{\cellcolor[HTML]{EFEFEF}\textcolor{green}\cmark} &
  \multicolumn{1}{c|}{\cellcolor[HTML]{EFEFEF}\textcolor{red}\xmark} &
  \multicolumn{1}{c|}{\cellcolor[HTML]{EFEFEF}\textcolor{red}\xmark} &
  \multicolumn{1}{c|}{\cellcolor[HTML]{EFEFEF}\textcolor{green}\cmark} &
  \multicolumn{1}{c|}{\cellcolor[HTML]{EFEFEF}\textcolor{red}\xmark} &
  \multicolumn{1}{c|}{\cellcolor[HTML]{EFEFEF}\textcolor{red}\xmark} &
  \multicolumn{1}{c|}{\cellcolor[HTML]{EFEFEF}\textcolor{red}\xmark} &
  \multicolumn{1}{c|}{\cellcolor[HTML]{EFEFEF}\textcolor{green}\cmark} &
  \multicolumn{1}{c|}{\cellcolor[HTML]{EFEFEF}\textcolor{green}\cmark} &
  \multicolumn{1}{c|}{\cellcolor[HTML]{EFEFEF}\textcolor{red}\xmark} &
  \multicolumn{1}{c|}{\cellcolor[HTML]{EFEFEF}\textcolor{red}\xmark} &
  \multicolumn{1}{c|}{\cellcolor[HTML]{EFEFEF}\textcolor{green}\cmark} &
  \multicolumn{1}{c|}{\cellcolor[HTML]{EFEFEF}\textcolor{red}\xmark} &
  \multicolumn{1}{c|}{\cellcolor[HTML]{EFEFEF}\textcolor{green}\cmark} &
  \multicolumn{1}{c|}{\cellcolor[HTML]{EFEFEF}\textcolor{red}\xmark} &
  \multicolumn{1}{c|}{\cellcolor[HTML]{EFEFEF}\textcolor{green}\cmark} &
  \multicolumn{1}{c|}{\cellcolor[HTML]{EFEFEF}\textcolor{green}\cmark} &
  \textcolor{red}\xmark \\
\rowcolor[HTML]{FFFFFF} 
Charagraph \cite{masson2023Charagraph} &
  \multicolumn{1}{c|}{\cellcolor[HTML]{FFFFFF}\textcolor{red}\xmark} &
  \multicolumn{1}{c|}{\cellcolor[HTML]{FFFFFF}\textcolor{red}\xmark} &
  \multicolumn{1}{c|}{\cellcolor[HTML]{FFFFFF}\textcolor{green}\cmark} &
  \multicolumn{1}{c|}{\cellcolor[HTML]{FFFFFF}\textcolor{red}\xmark} &
  \multicolumn{1}{c|}{\cellcolor[HTML]{FFFFFF}\textcolor{red}\xmark} &
  \multicolumn{1}{c|}{\cellcolor[HTML]{FFFFFF}\textcolor{red}\xmark} &
  \multicolumn{1}{c|}{\cellcolor[HTML]{FFFFFF}\textcolor{green}\cmark} &
  \multicolumn{1}{c|}{\cellcolor[HTML]{FFFFFF}\textcolor{red}\xmark} &
  \multicolumn{1}{c|}{\cellcolor[HTML]{FFFFFF}\textcolor{red}\xmark} &
  \multicolumn{1}{c|}{\cellcolor[HTML]{FFFFFF}\textcolor{green}\cmark} &
  \multicolumn{1}{c|}{\cellcolor[HTML]{FFFFFF}\textcolor{green}\cmark} &
  \multicolumn{1}{c|}{\cellcolor[HTML]{FFFFFF}\textcolor{red}\xmark} &
  \multicolumn{1}{c|}{\cellcolor[HTML]{FFFFFF}\textcolor{red}\xmark} &
  \multicolumn{1}{c|}{\cellcolor[HTML]{FFFFFF}\textcolor{red}\xmark} &
  \multicolumn{1}{c|}{\cellcolor[HTML]{FFFFFF}\textcolor{green}\cmark} &
  \multicolumn{1}{c|}{\cellcolor[HTML]{FFFFFF}\textcolor{red}\xmark} &
  \multicolumn{1}{c|}{\cellcolor[HTML]{FFFFFF}\textcolor{red}\xmark} &
  \multicolumn{1}{c|}{\cellcolor[HTML]{FFFFFF}\textcolor{green}\cmark} &
  \multicolumn{1}{c|}{\cellcolor[HTML]{FFFFFF}\textcolor{red}\xmark} &
  \textcolor{green}\cmark \\
\rowcolor[HTML]{EFEFEF} 
ChartStory \cite{zhao2023chartstory} &
  \multicolumn{1}{c|}{\cellcolor[HTML]{EFEFEF}\textcolor{red}\xmark} &
  \multicolumn{1}{c|}{\cellcolor[HTML]{EFEFEF}\textcolor{red}\xmark} &
  \multicolumn{1}{c|}{\cellcolor[HTML]{EFEFEF}\textcolor{green}\cmark} &
  \multicolumn{1}{c|}{\cellcolor[HTML]{EFEFEF}\textcolor{red}\xmark} &
  \multicolumn{1}{c|}{\cellcolor[HTML]{EFEFEF}\textcolor{green}\cmark} &
  \multicolumn{1}{c|}{\cellcolor[HTML]{EFEFEF}\textcolor{red}\xmark} &
  \multicolumn{1}{c|}{\cellcolor[HTML]{EFEFEF}\textcolor{red}\xmark} &
  \multicolumn{1}{c|}{\cellcolor[HTML]{EFEFEF}\textcolor{green}\cmark} &
  \multicolumn{1}{c|}{\cellcolor[HTML]{EFEFEF}\textcolor{green}\cmark} &
  \multicolumn{1}{c|}{\cellcolor[HTML]{EFEFEF}\textcolor{red}\xmark} &
  \multicolumn{1}{c|}{\cellcolor[HTML]{EFEFEF}\textcolor{green}\cmark} &
  \multicolumn{1}{c|}{\cellcolor[HTML]{EFEFEF}\textcolor{red}\xmark} &
  \multicolumn{1}{c|}{\cellcolor[HTML]{EFEFEF}\textcolor{green}\cmark} &
  \multicolumn{1}{c|}{\cellcolor[HTML]{EFEFEF}\textcolor{red}\xmark} &
  \multicolumn{1}{c|}{\cellcolor[HTML]{EFEFEF}\textcolor{red}\xmark} &
  \multicolumn{1}{c|}{\cellcolor[HTML]{EFEFEF}\textcolor{red}\xmark} &
  \multicolumn{1}{c|}{\cellcolor[HTML]{EFEFEF}\textcolor{green}\cmark} &
  \multicolumn{1}{c|}{\cellcolor[HTML]{EFEFEF}\textcolor{red}\xmark} &
  \multicolumn{1}{c|}{\cellcolor[HTML]{EFEFEF}\textcolor{green}\cmark} &
  \multicolumn{1}{c|}{\cellcolor[HTML]{EFEFEF}\textcolor{red}\xmark} \\ 
\rowcolor[HTML]{FFFFFF} 
Erato \cite{sun2023erato} &
  \multicolumn{1}{c|}{\cellcolor[HTML]{FFFFFF}\textcolor{red}\xmark} &
  \multicolumn{1}{c|}{\cellcolor[HTML]{FFFFFF}\textcolor{red}\xmark} &
  \multicolumn{1}{c|}{\cellcolor[HTML]{FFFFFF}\textcolor{green}\cmark} &
  \multicolumn{1}{c|}{\cellcolor[HTML]{FFFFFF}\textcolor{red}\xmark} &
  \multicolumn{1}{c|}{\cellcolor[HTML]{FFFFFF}\textcolor{red}\xmark} &
  \multicolumn{1}{c|}{\cellcolor[HTML]{FFFFFF}\textcolor{green}\cmark} &
  \multicolumn{1}{c|}{\cellcolor[HTML]{FFFFFF}\textcolor{red}\xmark} &
  \multicolumn{1}{c|}{\cellcolor[HTML]{FFFFFF}\textcolor{red}\xmark} &
  \multicolumn{1}{c|}{\cellcolor[HTML]{FFFFFF}\textcolor{red}\xmark} &
  \multicolumn{1}{c|}{\cellcolor[HTML]{FFFFFF}\textcolor{green}\cmark} &
  \multicolumn{1}{c|}{\cellcolor[HTML]{FFFFFF}\textcolor{green}\cmark} &
  \multicolumn{1}{c|}{\cellcolor[HTML]{FFFFFF}\textcolor{red}\xmark} &
  \multicolumn{1}{c|}{\cellcolor[HTML]{FFFFFF}\textcolor{red}\xmark} &
  \multicolumn{1}{c|}{\cellcolor[HTML]{FFFFFF}\textcolor{green}\cmark} &
  \multicolumn{1}{c|}{\cellcolor[HTML]{FFFFFF}\textcolor{red}\xmark} &
  \multicolumn{1}{c|}{\cellcolor[HTML]{FFFFFF}\textcolor{red}\xmark} &
  \multicolumn{1}{c|}{\cellcolor[HTML]{FFFFFF}\textcolor{green}\cmark} &
  \multicolumn{1}{c|}{\cellcolor[HTML]{FFFFFF}\textcolor{red}\xmark} &
  \multicolumn{1}{c|}{\cellcolor[HTML]{FFFFFF}\textcolor{green}\cmark} &
  \textcolor{red}\xmark \\
\rowcolor[HTML]{EFEFEF} 
DataTales \cite{sultanum2023datatales} &
  \multicolumn{1}{c|}{\cellcolor[HTML]{EFEFEF}\textcolor{red}\xmark} &
  \multicolumn{1}{c|}{\cellcolor[HTML]{EFEFEF}\textcolor{red}\xmark} &
  \multicolumn{1}{c|}{\cellcolor[HTML]{EFEFEF}\textcolor{green}\cmark} &
  \multicolumn{1}{c|}{\cellcolor[HTML]{EFEFEF}\textcolor{green}\cmark} &
  \multicolumn{1}{c|}{\cellcolor[HTML]{EFEFEF}\textcolor{green}\cmark} &
  \multicolumn{1}{c|}{\cellcolor[HTML]{EFEFEF}\textcolor{red}\xmark} &
  \multicolumn{1}{c|}{\cellcolor[HTML]{EFEFEF}\textcolor{red}\xmark} &
  \multicolumn{1}{c|}{\cellcolor[HTML]{EFEFEF}\textcolor{green}\cmark} &
  \multicolumn{1}{c|}{\cellcolor[HTML]{EFEFEF}\textcolor{red}\xmark} &
  \multicolumn{1}{c|}{\cellcolor[HTML]{EFEFEF}\textcolor{red}\xmark} &
  \multicolumn{1}{c|}{\cellcolor[HTML]{EFEFEF}\textcolor{red}\xmark} &
  \multicolumn{1}{c|}{\cellcolor[HTML]{EFEFEF}\textcolor{green}\cmark} &
  \multicolumn{1}{c|}{\cellcolor[HTML]{EFEFEF}\textcolor{green}\cmark} &
  \multicolumn{1}{c|}{\cellcolor[HTML]{EFEFEF}\textcolor{red}\xmark} &
  \multicolumn{1}{c|}{\cellcolor[HTML]{EFEFEF}\textcolor{red}\xmark} &
  \multicolumn{1}{c|}{\cellcolor[HTML]{EFEFEF}\textcolor{green}\cmark} &
  \multicolumn{1}{c|}{\cellcolor[HTML]{EFEFEF}\textcolor{green}\cmark} &
  \multicolumn{1}{c|}{\cellcolor[HTML]{EFEFEF}\textcolor{red}\xmark} &
  \multicolumn{1}{c|}{\cellcolor[HTML]{EFEFEF}\textcolor{green}\cmark} &
  \multicolumn{1}{c|}{\cellcolor[HTML]{EFEFEF}\textcolor{red}\xmark} \\
\rowcolor[HTML]{FFFFFF} 
DataNarrative \cite{islam2024datanarrative} &
  \multicolumn{1}{c|}{\cellcolor[HTML]{FFFFFF}\textcolor{red}\xmark} &
  \multicolumn{1}{c|}{\cellcolor[HTML]{FFFFFF}\textcolor{red}\xmark} &
  \multicolumn{1}{c|}{\cellcolor[HTML]{FFFFFF}\textcolor{green}\cmark} &
  \multicolumn{1}{c|}{\cellcolor[HTML]{FFFFFF}\textcolor{green}\cmark} &
  \multicolumn{1}{c|}{\cellcolor[HTML]{FFFFFF}\textcolor{red}\xmark} &
  \multicolumn{1}{c|}{\cellcolor[HTML]{FFFFFF}\textcolor{green}\cmark} &
  \multicolumn{1}{c|}{\cellcolor[HTML]{FFFFFF}\textcolor{red}\xmark} &
  \multicolumn{1}{c|}{\cellcolor[HTML]{FFFFFF}\textcolor{green}\cmark} &
  \multicolumn{1}{c|}{\cellcolor[HTML]{FFFFFF}\textcolor{green}\cmark} &
  \multicolumn{1}{c|}{\cellcolor[HTML]{FFFFFF}\textcolor{red}\xmark} &
  \multicolumn{1}{c|}{\cellcolor[HTML]{FFFFFF}\textcolor{red}\xmark} &
  \multicolumn{1}{c|}{\cellcolor[HTML]{FFFFFF}\textcolor{green}\cmark} &
  \multicolumn{1}{c|}{\cellcolor[HTML]{FFFFFF}\textcolor{green}\cmark} &
  \multicolumn{1}{c|}{\cellcolor[HTML]{FFFFFF}\textcolor{red}\xmark} &
  \multicolumn{1}{c|}{\cellcolor[HTML]{FFFFFF}\textcolor{red}\xmark} &
  \multicolumn{1}{c|}{\cellcolor[HTML]{FFFFFF}\textcolor{red}\xmark} &
  \multicolumn{1}{c|}{\cellcolor[HTML]{FFFFFF}\textcolor{green}\cmark} &
  \multicolumn{1}{c|}{\cellcolor[HTML]{FFFFFF}\textcolor{red}\xmark} &
  \multicolumn{1}{c|}{\cellcolor[HTML]{FFFFFF}\textcolor{green}\cmark} &
  \textcolor{red}\xmark \\
\hline
\end{tabular}%
}
\end{table*}

\begin{figure*}[t]
    \centering
    \includegraphics[width=.85\textwidth]{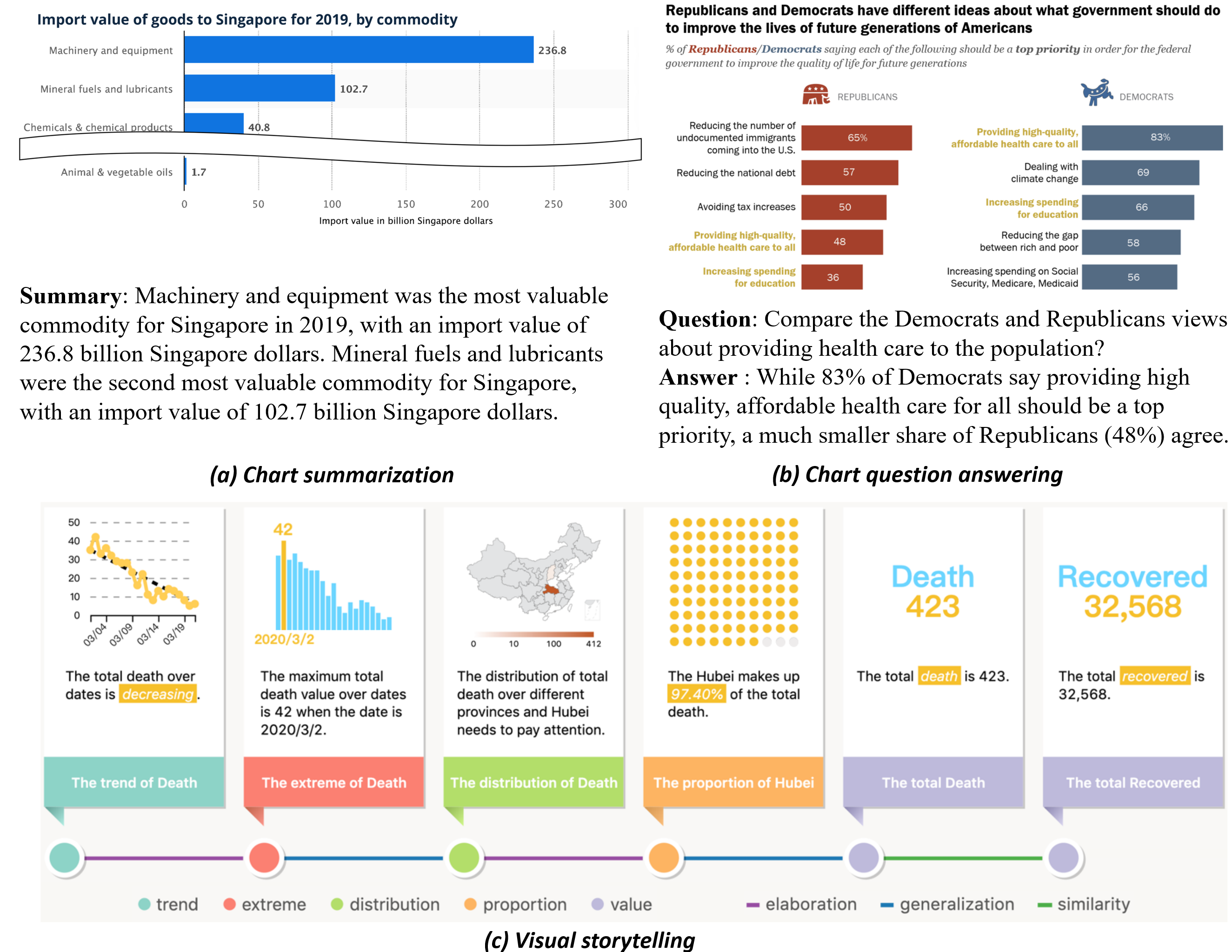}
    \caption{An overview of the types of NLG tasks. Here, sub-figure \textbf{(a)} depicts an example output of Chart Summarization \cite{kantharaj-etal-2022-chart}, sub-figure \textbf{(b)} depicts an example Chart Question-Answering from \cite{kantharaj-etal-2022-opencqa}, and sub-figure \textbf{(c)} denotes an example generation of visual story from Calliope \cite{shi2021calliope}.    
    }
    \label{fig:why-dimension}
\end{figure*}

\noindent \textbf{(a) Why?} 
The \textit{why} dimension of the problem space characterizes why the NLG is performed. In the context of visualization, several key tasks and corresponding applications necessitate the use of NLG techniques (see Figure \ref{fig:nlg-vis-overview}a). For example, \textbf{chart summarization} focuses on generating natural language descriptions that explain how to interpret a chart and/or what are some important patterns, trends, and outliers in the chart~\cite{kantharaj-etal-2022-chart}. In \textbf{chart question answering} (QA), the goal is to take charts and natural language questions as input and automatically generate the answer to facilitate visual data analysis~\cite{hoque2022chart}. Such questions often require textual explanations as answers~\cite{kantharaj-etal-2022-opencqa}. Finally, \textbf{data-driven storytelling} employs narrative techniques to guide the audience through a sequence of visualizations and texts~\cite{riche2018data,gershon2001storytelling, kwon2014visjockey, segel2010narrative}. Recently, there have been attempts to automatically generate stories from data~\cite{shi2021calliope, chen2020towards, cui2020text-to-viz, wang2020datashot}, but automatic data storytelling is still in its infancy. We limit our survey of ChartQA and Data storytelling domain to mainly those that generate natural language output.

\noindent \textbf{(b) What?} The \textit{what} dimension refers to the possible input and output dimensions for the NLG task (see Figure \ref{fig:nlg-vis-overview}b). Several NLG systems take visualizations as \textbf{input} and produce summaries, captions, or data facts. In some cases, the system may have access to the underlying data table~\cite{obeid2020charttotext}, while in others, the input visualization may be in bitmap image format without access to the raw data table (e.g., ~\cite{chen2019figure, speafico2020neural, kantharaj-etal-2022-chart}). Instead of visualizations, certain systems directly take data tables as input, generating charts and textual explanations as output~(e.g.,~\cite{shi2021calliope, wang2020datashot}). 
Moreover, some NLIs integrate multiple input modalities including mouse or touch interactions combined with text and visualizations~(e.g., \cite{hoque2018applying-pragmatics, srinivasan2018orko, srinivasan2020inchorus}). However, the NLG \textbf{outputs} in these cases tend to be minimal. For outputs, some systems ~\cite{hsu-etal-2021-scicap-generating, chen2019figure, kantharaj-etal-2022-chart, kantharaj-etal-2022-opencqa} exclusively produce texts, while others combine texts with visualizations (e.g., \cite{shi2021calliope}). \change{Overall, the usages of different input and output dimensions depend on the tasks and application (\textit{why} dimension) being addressed.}

\noindent \textbf{(c) How?} 
The \textit{how} aspect of the problem space focuses on how the NLG method processes the input and how it generates the output (see Figure \ref{fig:nlg-vis-overview}c). \change{The nature of the NLG methods often depends on the \textit{what} dimension.} For example, many NLG systems take \textit{visualizations} in the form of bitmap image as input and apply computer vision techniques to extract image features and/or apply optical character recognition techniques~\cite{hsu-etal-2021-scicap-generating, chen2019figure}. Others accept \textit{data tables} and as input and apply either rule-based approaches~\cite{srinivasan2019augmenting, wang2020datashot} as well as deep learning models to generate texts as well as visualizations (e.g., ~\cite{shi2021calliope}). 

\noindent \textbf{(d) Where?} 
The \textit{where} part of the problem focuses on the placement of texts (see Figure \ref{fig:nlg-vis-overview}d). Most NLG systems output text as a separate entity without any explicit links to the related visualization \cite{kantharaj-etal-2022-chart} while others automatically annotate salient portions in visualizations with textual labels~\cite{hullman2013contextifier, gao2014newsviews}. There are also attempts to automatically link between texts and visualizations so that clicking on text highlights the corresponding portions in the chart and vice versa \cite{latif2021kori, kim2023emphasischecker}. 

\noindent \textbf{(e) When?} The \textit{when} dimension pertains to the temporal organization of generated texts (see Figure \ref{fig:nlg-vis-overview}e). This aspect is closely related to document planning in NLG architecture~\cite{reiter2007architecture} and narrative structure in data-stories~\cite{riche2018data}, which specifically concerns how the sequence of events, comprising textual elements, are temporally ordered and the potential variations in the pacing of the narrative. The narrative structure can be categorized in various ways: such as linear vs. non-linear sequence ~\cite{riche2018data} as well as author-driven (narrative flow intended by the author) vs. reader-driven (interactive exploration by readers) ~\cite{segel2010narrative}. 

\change{For each of the Sections \ref{sec:why}-\ref{sec-when}, we discuss the existing work and research opportunities in the NLG for the visualizations domain. In particular, we analyzed the different NLG systems to identify the major classes of problems and challenges inherent to each Wh-question. We summarize the key challenges and avenues for future work in Section \ref{challenges}.}

\section{Why?}
\label{sec:why}
In this section, we discuss the primary NLG tasks that require NLG techniques as well as different applications that necessitate such tasks.

\subsection{Downstream Tasks}
\label{subsec-chartsumm}

\noindent \textbf{Chart Summarization:} This task focuses on generating natural language descriptions that explain how to interpret a chart and/or what are some important patterns, trends, and outliers in the chart~\cite{kantharaj-etal-2022-chart}. Figure~\ref{fig:why-dimension}a shows an example output from a chart summarization model~\cite{kantharaj-etal-2022-chart} that describes some statistical facts regarding a bar chart. The chart summarization task has been presented in different variations in the existing literature. Earlier works primarily focused on chart captioning by explaining the elements and the visual encodings in charts~\cite{mittal-etal-1998-describing, ferres2013evaluating}. Others have focused on generating template-based approaches to generate sentences that describe simple statistical facts such as maximum and minimum values, and comparisons ~\cite{srinivasan2019augmenting, cui2019datasite}. More recently, there have been some attempts to generate paragraphs that describe more complex insights such as perceptual trends and patterns using deep learning methods~\cite{chen2019figure, speafico2020neural, mahinpei2022linecap, hsu-etal-2021-scicap-generating, kantharaj-etal-2022-chart, tang2023vistext}. Some of these works focus on specific types of visualizations such as line charts~\cite{speafico2020neural, mahinpei2022linecap}.

A key challenge in addressing the chart summarization task is the lack of large-scale real-world datasets that are required for training and evaluation. Recently, there have been some efforts to build benchmark datasets that contain real-world charts from different domains along with captions/summaries regarding these charts (see Table \ref{tab:Chartsumm-bench}). While these benchmarks provide a starting point for training and evaluation, they are still limited in terms of chart types and texts. For example, in these benchmarks, only some common chart types such as lines and bars are covered, while in some benchmarks the reference text captions are based on templates that may lack variations compared to human-authored captions tailored to specific charts. Another important limitation is that captions that describe perceptual or cognitive features, which are valuable to readers, are either absent or comprise only a small portion of the dataset~\cite{tang2023vistext}. 

\begin{table}[t!]
\centering
\caption{The table presents an overview of the existing benchmark datasets in the Chart Summarization task covering various chart types.}
\label{tab:Chartsumm-bench}
\resizebox{\columnwidth}{!}{%
\begin{tabular}{|l|l|l|}
\hline
                                           &        &                 \\
                                           &        &                 \\
\multirow{-3}{*}{Dataset} &
  \multirow{-3}{*}{\begin{tabular}[c]{@{}c@{}}Total \# of\\  Chart-\\ Summary \\ pairs\end{tabular}} &
  \multirow{-3}{*}{Chart types} \\ \hline
\rowcolor[HTML]{EFEFEF} 
Spreafico et al. \cite{speafico2020neural} & 100    & Line            \\ \hline
Obeid et al. \cite{obeid2020charttotext}   & 8,305  & Bar, Line       \\ \hline
\rowcolor[HTML]{EFEFEF} 
LineCAP \cite{mahinpei2022linecap}         & 3,528  & Line            \\ \hline
Chart-to-Text \cite{kantharaj-etal-2022-chart} &
  44,096 &
  Bar, Line, Area, Scatter, Pie, Table \\ \hline
\rowcolor[HTML]{EFEFEF} 
ChartSumm \cite{rahman2023chartsumm}       & 84,363 & Bar, Line, Pie  \\ \hline
VisText \cite{tang2023vistext}             & 12,441 & Bar, Line, Area \\ \hline
\end{tabular}%
}
\end{table}

\noindent \textbf{ChartQA:} For this task, the goal is to take a chart and a natural language question as input and automatically generate the answer to facilitate visual data analysis~\cite{hoque2022chart}. Such questions may require explanatory responses. For example, given the question “How have the house prices in Toronto changed over time?” and a line chart that shows home prices in different cities, the generated text could describe the price trends~\cite{kantharaj-etal-2022-opencqa}. Figure~\ref{fig:why-dimension}b  shows an example output from a ChartQA model~\cite{kantharaj-etal-2022-chart} that provides an explanatory answer. Others used natural language generation to explain how the model computes the answer to improve interoperability and transparency of the model~\cite{kim2020answering}. Overall, while the ChartQA task has predominantly concentrated on producing concise answers in the form of words or phrases, the exploration of generating explanatory answers has been limited.

Another line of work on natural language interfaces for visualizations \cite{setlur2016eviza, srinivasan2018orko, hoque2018applying-pragmatics} supports users in exploring data by answering user's queries in natural language through conversational responses. Song et al. \cite{song2023marrying} proposed a dialogue system designed for creating visualizations through a series of back-and-forth conversations between the user and the system. However, the outputs of these systems usually convey simple, template-based information. A more recent study explores how large language models can be used for complex multi-turn question-answering tasks involving scientific visualizations \cite{li2023scigraphqa}. 

\noindent \textbf{Visual Data Storytelling:} Data-driven storytelling, a popular method for conveying insights, employs narrative techniques to guide the audience through a sequence of visualizations and text \cite{riche2018data,gershon2001storytelling, kwon2014visjockey, segel2010narrative, hullman2013adeeper}. 
Often such stories are developed with cohesive narrative, integrating visual aids like highlighting and animations in charts accompanied by textual annotations. 
For example, a manually crafted data story might go through a sequence of line charts to explain the factors contributing to global warming, culminating in the conclusion that greenhouse gases are the primary cause~\cite{bloomberg-story}.  While there have been initial efforts to automatically generate stories from data \cite{shi2021calliope, chen2020towards, cui2020text-to-viz, wang2020datashot, sultanum2023datatales}, automatic data storytelling is still in its early stages. 
Calliope \cite{shi2021calliope}, generates visual data stories from a spreadsheet, combining a series of visualizations with textual outputs (see Figure \ref{fig:why-dimension}c). Additionally, it offers an online editor for modifications of these stories. Datashot \cite{wang2020datashot} focuses on producing informative factsheets from tables, integrating texts in generated visualizations as factsheets. Others focused on generating infographics that combine visualization with accompanying text~\cite{cui2020text-to-viz, chen2020towards}. Recent research such as Erato \cite{sun2023erato} and Socrates \cite{wu2023socrates}, have focused on incorporating user feedback in textual form at the story generation phase, rather than eliciting feedback after the completion of story generation. 

\subsection{Applications}
The above NLG tasks can offer many benefits and potential applications such as enhancing accessibility~\cite{Kim2021}, supporting visual data analysis~\cite{hoque2018applying-pragmatics}, and improving information retrieval algorithms~\cite{li2013towards}. For example, Chart summarization and ChartQA can enhance chart accessibility for people who are blind since they can use screen readers to understand what is being presented in visualizations~\cite{voxLens, srinivasan2023azimuth, kim2023exploring, alam2023seechart}. SeeChart is a Chrome extension that automatically analyzes SVG charts to generate captions and subsequently supports people who are blind to explore charts~\cite{alam2023seechart}. VoxLens \cite{voxLens} is another tool for developers that provides a descriptive summary of a chart and then allows users to interact with the visualization by using voice-activated commands. Azimuth \cite{srinivasan2023azimuth} creates dashboards for screen reader-based navigation along with text descriptions to support dashboard comprehension and interaction.  
NLG outputs can assist the authoring of data-driven articles and expedite the writing process~\cite{sultanum2023datatales}. Finally, the generated captions can be used to index visualization to improve visualization recommendation~\cite{oppermann2020vizcommender} and retrieval applications~\cite{li2013towards, hoque2020searching}.

\subsection{Discussion}
While some significant progress has been made in defining several NLG tasks and creating benchmarks related to visualizations, several new tasks and applications can be explored in the future.

\noindent \textbf{Combining question answering with data-story generation:} For the ChartQA task~\cite{hoque2022chart}, most existing works focus on only the natural language understanding to produce answers which are just a number or word/phrase~\cite{hoque2022chart} while open-ended QA tasks are rare. The OpenCQA task does focus on open-ended question answering with charts, however, the output is only limited to text~\cite{kantharaj-etal-2022-opencqa}. Producing an answer that combines both textual explanation and visualization can be very useful in many scenarios. For instance, in addressing a question like \textit{``How have the house prices in Toronto changed over time?''}, the system could use a line chart to depict the variation in average house prices over time while providing a text summary that explains the trends in pricing. Answering open-ended questions can be even more effective by creating a visual story with a series of visualizations and texts. For example, given the question \textit{``What is really warming the world?''}\footnote{\href{https://www.bloomberg.com/graphics/2015-whats-warming-the-world/}{What is really warming the world?}}, a manually crafted story can show a sequence of line charts progressively to tell a compelling story about how greenhouse gas is the main contributing factor. Automatically creating similar stories using NLG methods to answer such questions is a challenging task that can be explored in the future.

\noindent \textbf{Summarizing and question answering with dashboards:} While existing research in Chart Summarization \cite{chen2019figure, speafico2020neural, hsu-etal-2021-scicap-generating, kantharaj-etal-2022-chart, rahman2023chartsumm} and open-ended ChartQA \cite{kantharaj-etal-2022-opencqa} utilizes only a single chart or the associated data table, generating NLG outputs for dashboard and multiple-coordinated views are rare. While some works on NLI\cite{hoque2018applying-pragmatics, setlur2020sneakpique} explored natural language interactions with multiple views, they were not focused on generating NLG output. In real-world scenarios, people usually interact with dashboards with multiple views and often ask natural language queries that require reasoning from multiple views to generate explanatory output to those queries. In this context, a starting point might be to collect a substantial number of real-world, human-annotated queries about dashboards that require explanatory responses.

\noindent \textbf{Verifying facts about charts:} As we will show in Section~\ref{sec-how}, the generated summaries about charts sometimes present misleading information, advocate for a biased viewpoint, or present a statement that is not factually correct. Visualizations created by humans can also spread misinformation \cite{lo2022misinform}. In this context, an important task would be to take a visualization and related descriptions as input and detect any incorrect facts or misinformation. Several works have begun exploring this problem. For example, Kim et al. \cite{kim2023emphasischecker} highlight text portions (e.g., a description of a trend) that cannot be supported by the chart, while Akhtar et al. automatically check the veracity of claims about simple bar charts~\cite{akhtar2023reading}. Huang et al. present a typology of different factual errors in NLG outputs of visualizations and present a task on correcting the factual errors in chart captions~\cite{huang2023lvlms}. In the future, more effort is needed to detect factual errors in NLG outputs as well as other issues such as biased output or disinformation which can be harmful to users.

\section{What?}
\label{sec-what}
In this section, we elaborate on the \textit{what} dimension of the problem space that characterizes the possible input and output dimensions for the NLG task.
\subsection{Input}
\label{subsec-input}
Typically, NLG systems for visualizations take data tables or visualizations as input, in addition, they may also take textual and multimodal inputs.
 
\subsubsection{Visualization} 
Most NLG systems take visualizations as input and produce summaries, captions, or data facts. In some cases, the system may have access to the underlying data table~\cite{obeid2020charttotext, rahman2023chartsumm, tang2023vistext, srinivasan2019augmenting, choi2022intentable, liu2023autotitle}, while in others, the input visualization may be in bitmap image format without access to the raw data (e.g., ~\cite{chen2019figure, speafico2020neural, hsu-etal-2021-scicap-generating, kantharaj-etal-2022-chart, mahinpei2022linecap, tan2022scientific, liu2023autotitle}). In general, generating captions from bitmap images is more challenging as the model needs to understand the chart content using computer vision techniques.

Regarding visualization types, most NLG systems are limited to taking some basic chart types (bar, line, etc.) as input, as shown in Table~\ref{tab:Chartsumm-bench}. One limitation of these systems is that less frequent chart types like radar charts, heatmaps, and box plots have not been used as input yet. Similarly, existing NLG systems rarely support multiple visualizations as input. One notable exception is ChartStory \cite{zhao2023chartstory}, a data storytelling system that takes an ensemble of visualizations from an exploratory analysis as inputs, identifies story pieces in a narrative form from these visualizations, and generates explanations to annotate the charts. 
 
\subsubsection{Data table} 

Instead of visualizations, certain systems directly take data tables as input, generating charts and textual explanations as output~(e.g.,~\cite{wang2020datashot, shi2021calliope,  sun2023erato, choi2022intentable, liu2023autotitle}). For example, DataShot takes a tabular dataset with multiple columns to extract various data facts and present these facts with infographics~\cite{wang2020datashot}. Calliope~\cite{shi2021calliope} and Erato~\cite{sun2023erato} also take spreadsheets but create visual stories with a series of texts and visualizations. When the model takes data tables as opposed to visualizations as input, it usually needs to automatically determine suitable visualizations that can be created from the data table before extracting chart summaries or data facts.

\subsubsection{Text} 

\begin{figure*}[t]
    \centering
    \includegraphics[width=\textwidth]{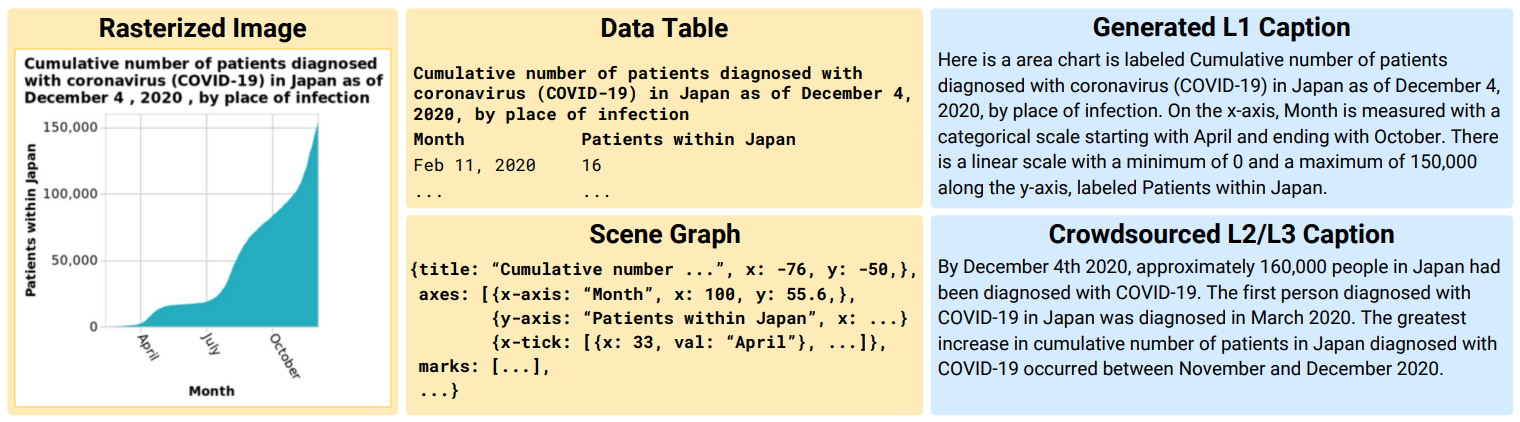}
    \caption{The figure denotes an example of different types of inputs, i.e., rasterized chart image, data table, scene graph (refers to a structured format similar to a web page's Document Object Model comprising the characteristics of a chart), and chart captions generated by the VisText system \cite{tang2023vistext}. The system produces Level 1, Level 2/Level 3 captions based on \cite{lundgard2021accessible}. The semantic levels are discussed  in Section \ref{subsubsec: output-text}.}
    \label{fig:vistext}
\end{figure*}

Some NLG systems take textual data as input in addition to visualizations and/or data tables(e.g., ~\cite{kantharaj-etal-2022-opencqa, tang2023vistext}). For example, Kantharaj et al. take open-ended questions and a chart as input, generating explanatory answers~(e.g.,~\cite{kantharaj-etal-2022-opencqa, masry2023unichart}). Besides questions about charts, others utilize textual inputs containing various metadata about charts (e.g., visual attributes, title, topics) ~\cite{mittal-etal-1998-describing, demir2012summarizing, ferres2013evaluating, speafico2020neural} as well as tokens extracted from the given chart  (e.g., \cite{kantharaj-etal-2022-opencqa}). Others attempt to capture the visual encoding information and the properties of visual elements in a chart (e.g., coordinates and colors of bar objects). For example, in VisText, in addition to the chart image and data table, a scene graph provides a hierarchical representation of a chart’s visual properties similar to a web page’s Document Object Model (DOM) to generate semantically rich captions (see Figure~\ref{fig:vistext}) \cite{tang2023vistext}. Such scene graphs can be accurately recovered from the charts that are available in SVG format. Moreover, scene graphs may provide a richer source of information than data tables as they encode visual properties of the chart (e.g., coordinates and colors) and are less noisy than tokens recovered via OCR.

\subsubsection{Multimodal} 

Research in the area of interactive visualizations has extensively explored various methods such as touch, pen  \cite{walny2012understanding}, body movements in front of large displays \cite{christopher2011infovis}, gestures \cite{badam2016supporting}, and even the synchronization of smartwatches with larger screens \cite{horak2018whendavid}. Despite these advancements, the use of natural language as an input has been largely overlooked previously. 
However, in recent years some natural language interfaces for visualizations have started to integrate natural language with other inputs like mouse and touch. For instance, Evizeon \cite{hoque2018applying-pragmatics} and Orko \cite{srinivasan2019augmenting} showcase this integration in map charts and network diagrams, while Eviza \cite{setlur2016eviza} highlights how natural language combined with mouse input can help clarify ambiguities. Additionally, the Sneak Pique \cite{setlur2020sneakpique} system enhances user interaction by offering auto-completion widgets that enable users to select natural language queries alongside textual input. Furthermore, research  into multi-modal interactions on tablet devices has shown that combining inputs like pen, touch, and speech is more effective and preferred by users, as it offers more flexibility in query expression~\cite{kassel2018valletto, srinivasan2020inchorus, saktheeswaran2020touchspeech}. In general, the above systems mostly highlight answers to the user's query through visualizations while the natural language output tends to be minimal and template-based(e.g., error messages and feedback to the user such as how many results are found). The full potential of combining natural language with other modalities in various contexts, from large-screen displays to mobile devices, remains an untapped area with significant opportunities for further exploration.

\subsection{Output}
\label{subsec-output}
Similar to various Input modalities of NLG systems for visualizations, Output can encompass different modes, i.e., Text, Visualizations, or Text and Visualizations combined. In this section, we discuss different modalities of Output.

\subsubsection{Text Only}
\label{subsubsec: output-text}
Many NLG systems for visualizations only output texts. However, the granularity levels and semantic contents of these texts can vary a lot.

\noindent \textbf{Granularity Levels:} Given a chart as input, some systems produce a sentence~\cite{hsu-etal-2021-scicap-generating, kim2020answering, mahinpei2022linecap, liu2023autotitle} or a list of sentences as data facts~\cite{srinivasan2019augmenting, shi2021calliope, choi2022intentable}, while many others produce a paragraph ~\cite{demir2012summarizing, speafico2020neural, obeid2020charttotext, kantharaj-etal-2022-opencqa} or both~\cite{chen2019figure}. More recently, there have been also attempts to generate multi-paragraph texts to support authors in creating data-driven articles~\cite{sultanum2023datatales}. In general, generating multi-paragraphs is more challenging to produce as the model needs to ensure that the generated texts are organized coherently.

\noindent \textbf{Semantic Levels:}  In terms of covering different semantic levels~\cite{lundgard2021accessible}, some systems only generate visual encoding descriptions~\cite{mittal-etal-1998-describing} and basic statistical information, while others cover sentences with high-level patterns and trends~\cite{kantharaj-etal-2022-chart}. We analyze the semantic contents of the NLG models using the framework developed by Lundgard and Satyanarayan~\cite{lundgard2021accessible} which identifies the following four levels:
\textit{Level \textbf{1}} describes aspects of the chart's construction (e.g., marks and visual encodings);  \textit{Level \textbf{2}} describes summary statistics and their interrelations (e.g., extrema and comparisons); \textit{Level \textbf{3}} synthesizes perceptual and cognitive phenomena (e.g., complex trends and patterns); and  \textit{Level \textbf{4}} provides domain-specific information (e.g., social and political context). 

Table \ref{tab:semantic-table} provides a summary of how existing research works are related to each of the semantic levels. Early works \cite{mittal-etal-1998-describing, demir2012summarizing, ferres2013evaluating, green2004autobrief} primarily concentrated on creating text-based descriptions for charts that contain low-level attribute information like colors, labels, etc., which can be associated with the semantic \textit{\textbf{Level 1}}. Subsequent studies \cite{chen2019figure, speafico2020neural, mahinpei2022linecap, hsu-etal-2021-scicap-generating, tan2022scientific, obeid2020charttotext} also generate \textit{\textbf{Level 2}} information. For example, 
Chen et. al. \cite{chen2019figure} focused on generating captions of various types of charts that describe the maxima or minima in the chart, while Hsu et. al. \cite{hsu-etal-2021-scicap-generating} proposed a system that generates captions of scientific figures which often describes the correlations between attributes present in the charts. Recent works have mostly focused on generating more higher level of summaries or captions, exemplified by the the works of \cite{tang2023vistext, kantharaj-etal-2022-chart, rahman2023chartsumm}. For instance, Tang et al. \cite{tang2023vistext} introduced an approach where each visualization is accompanied by three distinct captions (i.e., \textit{\textbf{Level 1},\textbf{ Level 2}, and \textbf{Level 3}}), each addressing a unique semantic level introduced by \cite{lundgard2021accessible}, while \cite{kantharaj-etal-2022-chart} included contextual and domain-specific information (\textit{\textbf{Level 4}}) as well in the generated summary. 

% \usepackage{multirow}
% \usepackage{graphicx}
\begin{table}[]
\centering
\caption{The table summarizes how existing NLG methods in the Chart Captioning and Chart Summarization domain are associated with the Four levels of semantics introduced by \cite{lundgard2021accessible}. Here, ``\textcolor{green}\cmark'' denotes a research work falls under any one of the levels, while ``\textcolor{red}\xmark'' denotes otherwise.}
\label{tab:semantic-table}
\resizebox{\columnwidth}{!}{%
\begin{tabular}{|c|cccc|}
\hline
 &
  \multicolumn{4}{c|}{Semantic Levels covered} \\ \cline{2-5} 
\multirow{-2}{*}{NLG Methods} &
  \multicolumn{1}{c|}{Level 1} &
  \multicolumn{1}{c|}{Level 2} &
  \multicolumn{1}{c|}{Level 3} &
  Level 4 \\ \hline
\rowcolor[HTML]{EFEFEF} 
Mittal et al. \cite{mittal-etal-1998-describing} &
  \multicolumn{1}{c|}{\cellcolor[HTML]{EFEFEF}\textcolor{green}\cmark} &
  \multicolumn{1}{c|}{\cellcolor[HTML]{EFEFEF}\textcolor{red}\xmark} &
  \multicolumn{1}{c|}{\cellcolor[HTML]{EFEFEF}\textcolor{red}\xmark} &
  \textcolor{red}\xmark \\ \hline
AutoBrief \cite{green2004autobrief} &
  \multicolumn{1}{c|}{\textcolor{green}\cmark} &
  \multicolumn{1}{c|}{\textcolor{red}\xmark} &
  \multicolumn{1}{c|}{\textcolor{red}\xmark} &
  \textcolor{red}\xmark \\ \hline
\rowcolor[HTML]{EFEFEF}
Reiter et al. \cite{reiter2007architecture} &
  \multicolumn{1}{c|}{\cellcolor[HTML]{EFEFEF}\textcolor{green}\cmark} &
  \multicolumn{1}{c|}{\cellcolor[HTML]{EFEFEF}\textcolor{red}\xmark} &
  \multicolumn{1}{c|}{\cellcolor[HTML]{EFEFEF}\textcolor{red}\xmark} &
  \textcolor{red}\xmark \\ \hline
Demir et al. \cite{demir2012summarizing} &
  \multicolumn{1}{c|}{\textcolor{green}\cmark} &
  \multicolumn{1}{c|}{\textcolor{red}\xmark} &
  \multicolumn{1}{c|}{\textcolor{red}\xmark} &
  \textcolor{red}\xmark \\ \hline
\rowcolor[HTML]{EFEFEF} 
Ferres et al. \cite{ferres2013evaluating} &
  \multicolumn{1}{c|}{\textcolor{green}\cmark} &
  \multicolumn{1}{c|}{\textcolor{red}\xmark} &
  \multicolumn{1}{c|}{\textcolor{red}\xmark} &
  \textcolor{red}\xmark \\ \hline
Chen et al. \cite{chen2019figure} &
  \multicolumn{1}{c|}{\textcolor{red}\xmark} &
  \multicolumn{1}{c|}{\textcolor{green}\cmark} &
  \multicolumn{1}{c|}{\textcolor{red}\xmark} &
  \textcolor{red}\xmark \\ \hline
\rowcolor[HTML]{EFEFEF} 
Spreafico et al. \cite{speafico2020neural} &
  \multicolumn{1}{c|}{\cellcolor[HTML]{EFEFEF}\textcolor{red}\xmark} &
  \multicolumn{1}{c|}{\cellcolor[HTML]{EFEFEF}\textcolor{green}\cmark} &
  \multicolumn{1}{c|}{\cellcolor[HTML]{EFEFEF}\textcolor{red}\xmark} &
  \textcolor{red}\xmark \\ \hline
Obeid et al. \cite{obeid2020charttotext} &
  \multicolumn{1}{c|}{\textcolor{red}\xmark} &
  \multicolumn{1}{c|}{\textcolor{green}\cmark} &
  \multicolumn{1}{c|}{\textcolor{red}\xmark} &
  \textcolor{red}\xmark \\ \hline
\rowcolor[HTML]{EFEFEF} 
SciCAP \cite{hsu-etal-2021-scicap-generating} &
  \multicolumn{1}{c|}{\cellcolor[HTML]{EFEFEF}\textcolor{red}\xmark} &
  \multicolumn{1}{c|}{\cellcolor[HTML]{EFEFEF}\textcolor{green}\cmark} &
  \multicolumn{1}{c|}{\cellcolor[HTML]{EFEFEF}\textcolor{red}\xmark} &
  \textcolor{red}\xmark \\ \hline
LineCAP \cite{mahinpei2022linecap} &
  \multicolumn{1}{c|}{\textcolor{red}\xmark} &
  \multicolumn{1}{c|}{\textcolor{green}\cmark} &
  \multicolumn{1}{c|}{\textcolor{green}\cmark} &
  \textcolor{red}\xmark \\ \hline
\rowcolor[HTML]{EFEFEF} 
Tan et al. \cite{tan2022scientific} &
  \multicolumn{1}{c|}{\cellcolor[HTML]{EFEFEF}\textcolor{red}\xmark} &
  \multicolumn{1}{c|}{\cellcolor[HTML]{EFEFEF}\textcolor{green}\cmark} &
  \multicolumn{1}{c|}{\cellcolor[HTML]{EFEFEF}\textcolor{red}\xmark} &
  \textcolor{red}\xmark \\ \hline
Chart-to-Text \cite{kantharaj-etal-2022-chart} &
  \multicolumn{1}{c|}{\textcolor{green}\cmark} &
  \multicolumn{1}{c|}{\textcolor{green}\cmark} &
  \multicolumn{1}{c|}{\textcolor{green}\cmark} &
  \textcolor{green}\cmark \\ \hline
\rowcolor[HTML]{EFEFEF} 
ChartSumm \cite{rahman2023chartsumm} &
  \multicolumn{1}{c|}{\textcolor{green}\cmark} &
  \multicolumn{1}{c|}{\cellcolor[HTML]{EFEFEF}\textcolor{green}\cmark} &
  \multicolumn{1}{c|}{\cellcolor[HTML]{EFEFEF}\textcolor{green}\cmark} &
  \textcolor{green}\cmark \\ \hline
VisText \cite{tang2023vistext} &
  \multicolumn{1}{c|}{\textcolor{green}\cmark} &
  \multicolumn{1}{c|}{\textcolor{green}\cmark} &
  \multicolumn{1}{c|}{\textcolor{green}\cmark} &
  \textcolor{red}\xmark \\ \hline
\end{tabular}
}
\end{table}

\subsubsection{Visualization} A few systems integrate NLG outputs directly within visualizations by annotating different portions of the chart with text or by adding titles and labels. Contextifier ~\cite{hullman2013contextifier}, for instance, augments line charts with textual annotations in various salient points in the chart.  
The EmphasisChecker system takes a time-series chart as input from the user and identifies prominent visual features from the chart and annotates the features on top of the charts with different colors \cite{kim2023emphasischecker}. Furthermore, when the user inputs a caption for the corresponding figure, the system automatically highlights and links relevant pieces of text into the salient points of the input visualization. Natural language interfaces such as \cite{narechania2021nl4dv, mitra2022facilitating} produce a visualization as an outcome based on some specific user query, however, they do not focus on text generation.  
ChartStory \cite{zhao2023chartstory}, another recent work, takes an ensemble of visualizations, i.e., charts as inputs and identify story pieces in a narrative from the visualizations, and finally integrates similar charts that follow the same narrative and add textual captions to the charts/visualizations to generate a data story. 

\change{In addition to augmenting existing visualizations with texts, others have focused on generating visualizations either as static images \cite{
schetinger2023doom} or as code output~\cite{wu2024automated}. However, this body of work do not particularly focus on summarizing key insights from charts, rather they have minimal or no text descriptions.
}

\subsubsection{Multimodal} 
Multimodal outputs of an NLG system may include text, visualizations, as well as audio and haptic feedback. Some recent studies have aimed to automatically generate visual stories that include both text and visualizations~\cite{chen2020towards, cui2020text-to-viz, shi2021calliope, wang2020datashot}. These studies generate a visualization that contains annotated text which depicts various factual information in sequence. In a more recent study, Erato\cite{sun2023erato} and Socrates \cite{wu2023socrates}, researchers combine multiple systems altogether to generate a coherent story. Based on the initial input, these systems extract facts and generate a data story consisting of multiple visualizations that include annotations. Additionally, in some studies, such as \cite{liu2021advisor, ren2017chartaccent}, authors often employ methods to generate visualizations with annotation to better represent the stories. They focus on creating narrative-driven or annotated visual stories, indicating a trend toward more integrated and informative visualization techniques. For accessible visualizations, other forms of output such as audio feedback and data sonification have also been explored to support people with visual disabilities~\cite{alam2023seechart,voxLens}.

\subsection{Discussion}
There are several avenues for enhancing the input and output modalities of the NLG task.

\noindent \textbf{Benchmark datasets with varying charts and semantic contents:} As we have discussed, current datasets lack varieties in terms of chart types (Table ~\ref{tab:Chartsumm-bench}) and the types of semantic contents in the captions (Table~\ref{tab:semantic-table}).  To address these limitations, there is a pressing need to develop more benchmark datasets sourced from real-world data, offering a broader spectrum of chart types and visual styles. Similarly, certain semantic levels (e.g., levels 3 and 4) are rarely present in current datasets \cite{kantharaj-etal-2022-chart}. This indicates a gap in research, underscoring the need for human-written summaries covering a diverse range of semantic contents. Finally, there is a resource limitation of comprehensive benchmark datasets consisting of coherent narratives and corresponding visualizations that are required in Visual Storytelling, which is crucial for advancing automatic visual story generation.

\noindent \textbf{Interactive NLG for visualizations:} Most existing NLG methods generate captions or data stories without any input from users during the generation stages. As such, when the output does not match with user's desired information needs, users are left with no choice in revising the output. 
In the future, a human-in-the-loop approach that enables users to provide feedback through text and direct manipulation could fulfill their information needs more effectively. For example, a user could select a segment of a line chart and the model could dynamically re-generate the caption focusing on that segment only. Another example is a text-based interface where the user can have interactive dialogues with the NLG system to explore and summarize data insights \cite{song2023marrying}.

\noindent \textbf{Introduction of new input/output modalities:} While existing works explored the possibility of introducing multimodal inputs and outputs there are several possible ways to further explore this avenue. A recent study \cite{hall2022chironomia} demonstrated remote presentations by overlaying interactive visualizations on a presenter's webcam feed so that presenters can interact with the data using hand gestures, enriching the storytelling and making presentations more engaging. In the future, such interactions can be integrated with NLG systems to interactively generate explanations based on the hand gestures and instructions made by the user. Moreover, NLG systems can be adopted for mobile and wearable devices where the natural language can be combined with other input and output modalities such as touch, audio, and haptic feedback. Understanding the nature of user queries about visualizations on wearable devices~\cite{rey2023towards} and how to adapt the NLG output to the limited screen space of wearable and mobile devices~\cite{lee2021mobile} are interesting directions for exploration.

\section{How?}
\label{sec-how}

In this Section, we discuss how various NLG methods process the input (e.g., visualizations,  data tables, natural language queries) and how they generate the textual output using various methods ranging from rule-based methods to deep learning models to the latest large language models.

\subsection{How the \textit{Input} is Processed?}

\noindent\textbf{Processing visualizations:} If the input visualization is given in bitmap image format, the model needs to apply some computer vision techniques first to understand the content of the chart. A straightforward approach is to apply Optical Character Recognition (OCR) to extract the texts ~\cite{kantharaj-etal-2022-chart}. However, such OCR output does not capture various important visual features from the chart. Therefore, others extracted visual features using various deep-learning architectures. For example, Kantharaj et al. ~\cite{kantharaj-etal-2022-opencqa} extracted the visual features of different marks in the chart image (e.g., bars, lines) using Mask R-CNN  \cite{he2017maskrcnn} with Resnet-101 as its backbone and then feed both OCR-text and image features to VL-T5 \cite{cho2021vlt5} which is trained on vision-language data. Several others have utilized the ResNet framework   \cite{chen2019figure, tan2022scientific, liu2020autocaption}  or the DenseNet framework \cite{mahinpei2022linecap}. Chen et al.,  \cite{chen2020towards} focus on parsing structural information (both local and global) from bitmap timeline infographics. The proposed system used a ResNeXt network \cite{xie2017resnext} to extract global Information that includes the representation scale, layout, and orientation of the timeline, and a FasterRCNN network \cite{ren2017fasterrcnn} to extract local information that involves the location, category, and pixel data of each visual element on the timeline. VisText ~\cite{tang2023vistext} applied a bottom-up feature extractor based on FasterRCNN~\cite{anderson2018bottom} which was originally used for image captioning and fed the features to the VL-T5 model. 

CNN models primarily handle image data, but vision transformers such as VL-T5~\cite{cho2021vlt5}  are capable of simultaneously processing both images and textual information. This dual-input approach enables them to gather more comprehensive contextual data. However,  these models typically necessitate training on extensive vision-language datasets to ensure high-quality performance.

\noindent\textbf{Processing other inputs:} Processing tables and textual data is relatively less challenging. Early work takes the underlying data table of the chart as input and follows some heuristics to decide what should be the content of the caption \cite{mittal-etal-1998-describing, ferres2013evaluating}, primarily by following Reiter's NLG pipeline for data-to-text generation~\cite{reiter2007architecture}. 
DataShot \cite{wang2020datashot} transforms raw tabular data into data facts by constructing data subspaces, enumerating fact types, and calculating fact scores based on the original tabular data. This process involves slicing and dicing the data table with different dimensions to enumerate the subspaces using the BUC algorithm \cite{han2006concepts}. Calliope \cite{shi2021calliope} and Erato \cite{sun2023erato} also use data spreadsheets as inputs, utilizing a Monte-Carlo Tree-based search algorithm to extract data facts, to generate a coherent fact-based data story. A few systems also apply heuristics and grammar-based parsing to process multimodal inputs that combine text and visualizations with direct manipulation methods although the NL output of these systems is minimal\cite{hoque2018applying-pragmatics, srinivasan2018orko}. Recent Deep learning based NLG systems simply flatten the tables and other data (e.g., textual queries, visualization specification) as a linear list of tokens and feed them to the models such as T5 or VL-T5 (e.g., ~\cite{kantharaj-etal-2022-chart, kantharaj-etal-2022-opencqa, tang2023vistext}).

While heuristics-based text processing relies on predefined rules to analyze data efficiently, deep learning-based text processing systematically learns from data to optimize performance and accuracy. However, pre-training or fine-tuning deep learning models usually require a significant amount of training data (e.g., charts, data tables, and human written captions), posing a challenge in many task scenarios.

\begin{figure*}[t]
    \centering
    \includegraphics[width=\textwidth]{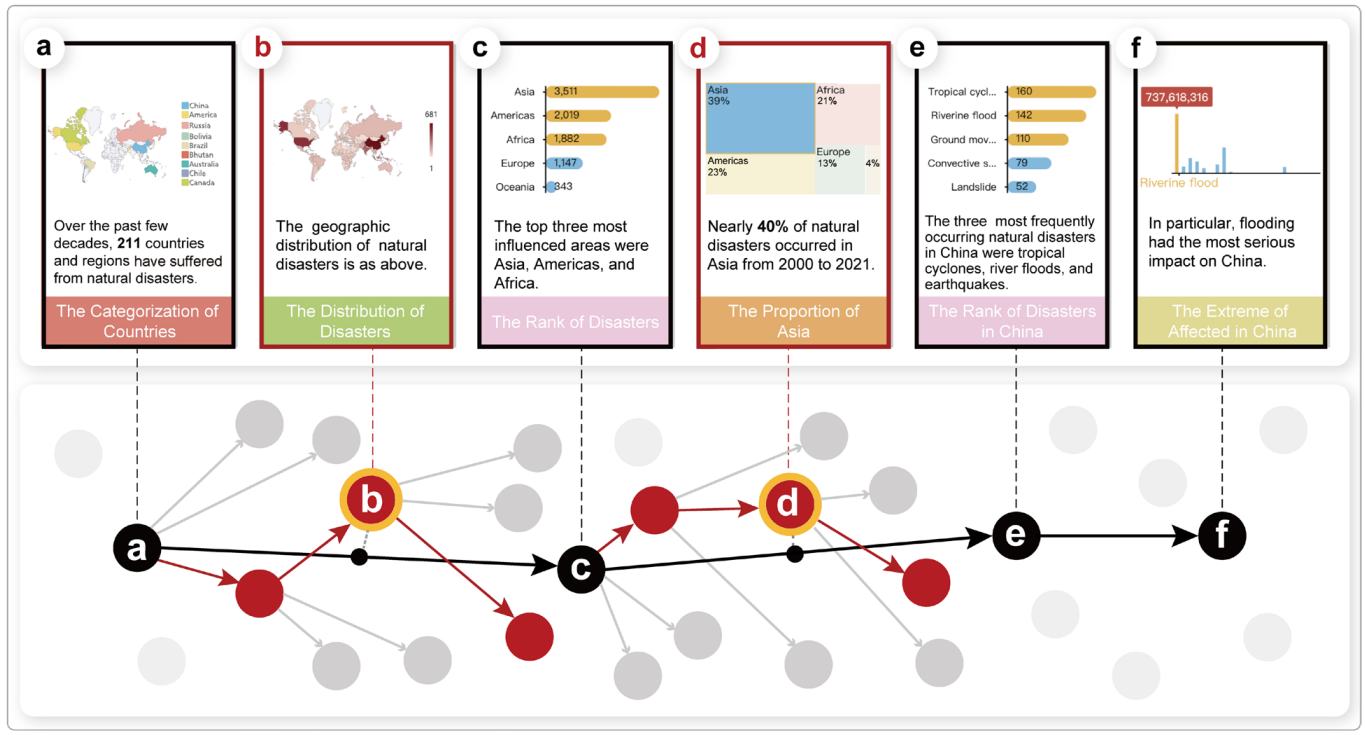}
    \caption{An example of Visual Storytelling system from \cite{sun2023erato}. The example presents a compelling storyline about global natural disasters. It features data points labeled in black (a, c, e, f), which were carefully chosen by a professional data analyst to serve as pivotal moments in the narrative. Additionally, points labeled in red (b, d) were generated through an interpolation algorithm.}
    \label{fig:visstory}
\end{figure*}

\subsection{How the \textit{output} is Generated?}
\subsubsection{Rule-based Method}
Early work on generating chart captions follows some planning-based approaches which can be captured by Reiter's architecture for data-to-text generation~\cite{reiter2007architecture}.
For example, Mittal et al.~\cite{mittal-etal-1998-describing} determine the structure of the caption based on some heuristics that rely on the marks and visual encodings of the chart and choose a complexity metric to select details of the caption. It then uses a text planner which uses predefined templates to generate the description of the chart. The iGRAPH-Lite system~\cite{ferres2013evaluating} also uses templates to provide a short description of what the chart looks like. Others perform statistical analysis (e.g.,  extrema, outliers) to present data facts as a list of sentences \cite{cui2019datasite, srinivasan2019augmenting}. Demir et al. \cite{demir2012summarizing} also compute statistics to generate bar chart summaries in a bottom–up manner to simultaneously construct the discourse and sentence structures. A key limitation of the above body of work is that sentences are generated using predefined templates, which may lack variations in terms of reported insights, grammatical styles, and lexical choices compared to data-driven models.

Several works also produce multimodal outputs combining both texts and visualizations as visual stories using pipeline-based approaches. DataShot \cite{wang2020datashot} transforms raw tabular data into data facts and computes the importance of different facts based on some properties of the fact (e.g., statistical properties). The selected facts are then organized with suitable layouts and visualization styles. Other works such as Calliope \cite{shi2021calliope} and Erato \cite{sun2023erato} introduce a story generation engine that utilizes a logic-oriented Monte Carlo tree search algorithm, which incrementally constructs the story by evaluating the importance of each data fact by exploring the data space which is initially retrieved from a spreadsheet. An example of the Erato system is presented in Figure \ref{fig:visstory}. In contrast, Zhao et al., \cite{zhao2023chartstory} introduce an interface that organizes and orders the charts within each story piece from the input ensemble of charts by constructing a weighted graph of the charts and computing a minimum spanning tree. In subsequent steps, the system generates appropriate captions and appropriate annotations, which contribute to the formation of the final visual story, providing a detailed and enriched narrative experience to the end user.

While rule-based methods in visual storytelling are efficient at identifying specific facts and arranging them sequentially, they tend to report simple statistical facts rather than complex insights that involve complex trends and patterns. There is also a lack of content planning that would construct a narrative structure with coherent facts (i.e., building relationships between facts) in the generation process. 

\subsubsection{Deep Learning Method}
Early deep learning NLG methods for visualizations rely on Convolutional Neural Network (CNN) and LSTM while more recent methods explored transformer-based models. 
Several works utilized a combination of CNN and LSTM architecture to extract features and then used template-based approaches to produce the chart captions\cite{chen2019figure, tan2022scientific,  liu2020autocaption}.  Andrea and Carenini \cite{speafico2020neural} employed an encoder-decoder architecture using two LSTM networks as the input and output both are in natural language. Chen et al. \cite{chen2020towards} focus on parsing structural information (both local and global) from bitmap timeline infographics. They utilize a ResNeXt network \cite{xie2017resnext} to extract global Information that includes the representation scale, layout, and orientation of the timeline and a FasterRCNN network \cite{ren2017fasterrcnn} to extract local information that involves the location, category, and pixel data of each visual element on the timeline. The final infographics visualization with textual annotation is constructed by augmenting new information to the bitmap timeline. 

In recent years, Transformer-based models (e.g., T5, BERT, and GPT) have outperformed their counterparts like LSTM architecture on various tasks which motivated researchers to fine-tune Transformer models on specific NLG tasks for visualizations \cite{devlin2019bert}. Tan et al. \cite{tan2022scientific} utilized GPT-2 \cite{radford2019language}, an encoder transformer architecture for caption generation tasks from an input scientific figure. In the same vein, some recent studies \cite{obeid2020charttotext, sun2023erato, kantharaj-etal-2022-opencqa, kantharaj-etal-2022-chart} fine-tuned transformers models like BERT \cite{devlin2019bert}, T5 \cite{raffel2020t5} to generate a summary or explanatory text given a visualization. 

\begin{figure*}[t]
    \centering
    \includegraphics[width=.85\textwidth]{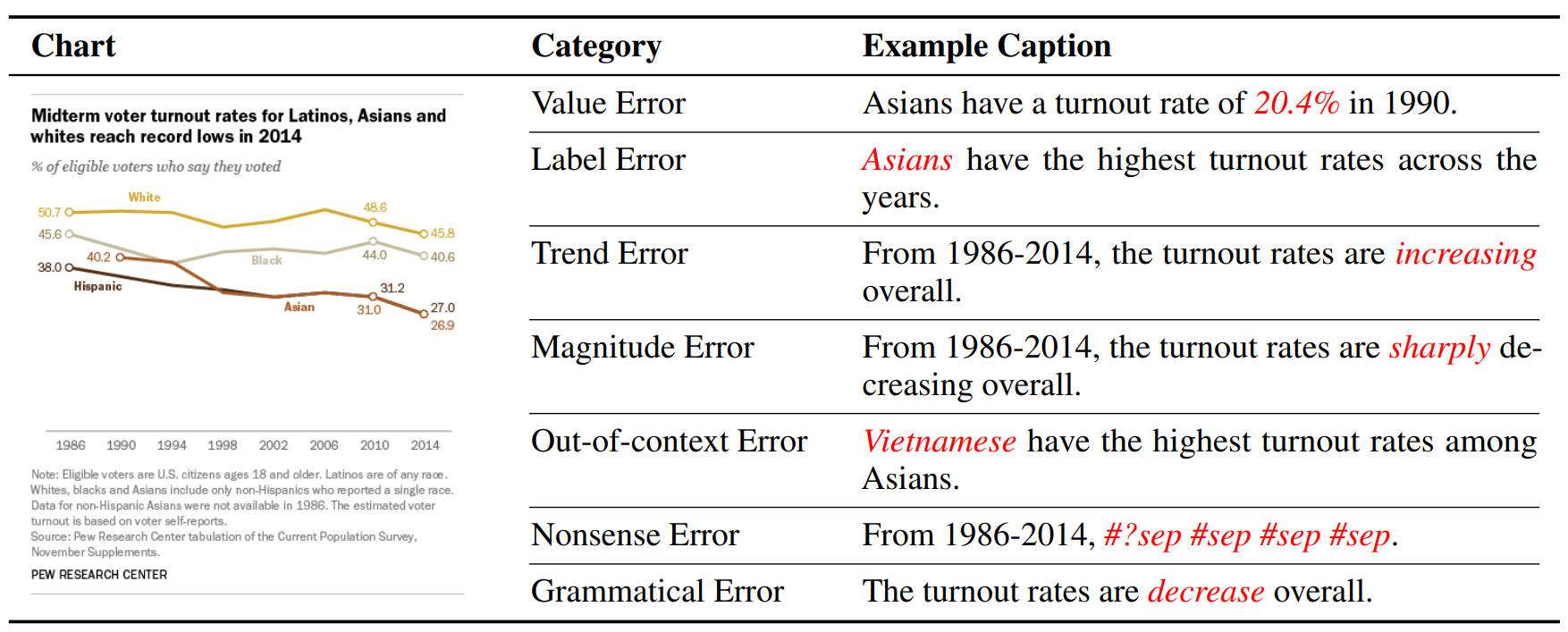}
    \caption{The figure presents a typology of errors proposed by \cite{huang2023lvlms} in the generated chart caption, namely, ``Value Error'' (incorrect data value), ``Label Error'' (incorrect label or category), ``Trend Error'' (incorrect presentation of a trend), ``Magnitude Error'' (incorrect presentation of the extent or degree of a trend's change), ``Out-of-context Error'' (value mentioned in the caption that does not exist in the chart), ``Nonsense Error'' (illogical inclusion of words), and ``Grammatical Error'' (grammatical mistakes). 
    }
    \label{fig:factual}
\end{figure*}

\begin{figure}[h]
    \centering
    \includegraphics[width=\columnwidth]{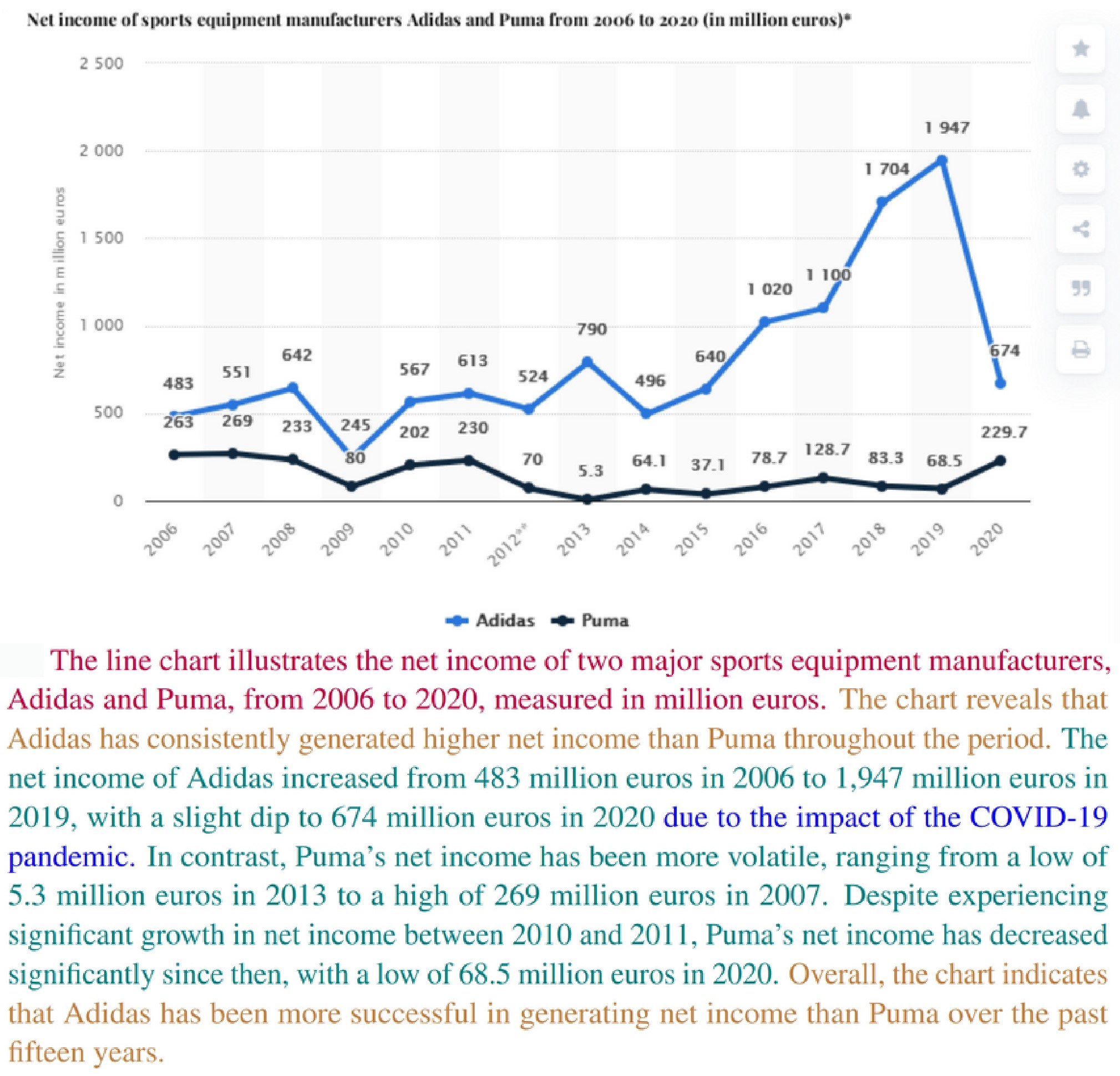}
    \caption{
    An example of potential model bias marked in \change{Blue} produced by ChatGPT (v3.5)\cite{Chatgpt} when prompted to generate summaries based on the given data table related to the chart. Here, text in \textcolor{fluorescentpink}{Purple} refers to Level 1, \textcolor{lightseagreen}{Green} refers to Level 2, \textcolor{gold}{Gold} refers to Level 3
    corresponding to the semantic levels of summary introduced by \cite{lundgard2021accessible}.}    
    \label{fig:bias}
\end{figure}

The rise of LLMs also prompted several researchers to study their effectiveness in chart summarization and story generation. An important question here is how to effectively construct prompts that help unlock the potential benefits of LLMs. DataTales \cite{sultanum2023datatales} leverages the linguistic generation capabilities of LLMs to generate coherent and relevant narratives given a chart as input. Long et al. found that prompt construction that involves few-shot demonstrations to LLM like ChatGPT 3.5  is almost as effective as fine-tuning models like T5~\cite{do2023llms}. Another recent study \cite{feng2023promptmag} combines natural language prompts with an image generation model \cite{rombach2022stablediff}, facilitating the creation of diverse visualizations that accurately reflect the specified prompts. Additionally, generative text-to-image models like DALL-E have been employed to explore the generation of visualizations, although the resulting output primarily manifests as an image rather than text\cite{schetinger2023doom}.

\subsection{Discussion}

While the recent surge of LLM has significantly moved forward the domain of NLG for visualizations the generated outputs suffer from various errors. We characterize these errors below and provide an overview of key areas of improvement.

\noindent \textbf{Data extraction challenges:} Charts often appear in bitmap image format in various sources (e.g., textbooks, scientific documents, and online articles), however, many NLG systems assume that the underlying data table is available~\cite{obeid2020charttotext, rahman2023chartsumm} or the chart is in SVG format from which it is easier to recover the properties of visualizations~\cite{tang2023vistext}. Some research attempted to recover the data table of the chart using chart data extraction models such as ChartOCR \cite{luo2021ChartOCR} as a pre-processing step. However, these OCR-based methods often fail to extract accurate and crucial information from charts, particularly from the ones that have many data points. Fine-tuning end-to-end vision-language models~\cite{lee2023pix2struct} that can better understand the structure and layout of different visualization types would be a promising approach to improve the chart data extraction challenge.

\noindent \textbf{Addressing factual errors:} NLG models sometimes output various errors that result in factually incorrect statements in the context of the given chart. Figure \ref{fig:factual} presents several examples of factual errors. For example, sometimes it reports an incorrect data value or labels while at other times it describes a trend incorrectly. We also notice that sometimes the model hallucinates where it outputs a value or label that does not exist in the chart and is not related to the chart (e.g., \textit{`Vietnamese'} did not appear in the given chart.) Given the pressing need to address these errors, there have been some recent attempts to establish the new task of chart caption factual error correction~\cite{huang2023lvlms} and hallucination detection \cite{sadat2023delucionqa, obaid-ul-islam-etal-2023-tackling}. \change{Future studies can build on these datasets of factual errors and develop new models to tackle these issues. More specifically, we need more  datasets with NLG outputs annotated with error types. Subsequently, a promising direction could be to fine-tune open-source LLMs with such annotated data to detect and rectify the errors.}

\noindent \textbf{Addressing bias in model outputs:} One notable concern for large language models is the model bias. In an exploratory study involving several line charts from the Statista dataset~\cite{kantharaj-etal-2022-chart}, we observed instances where trends around the peak period of Covid-19 were erroneously attributed to the pandemic, which might be a spurious correlation. Figure \ref{fig:bias} represents such an example of a potential bias of the ChatGPT \cite{openai2023chatgpt} model when asked to generate a summary of the given chart for which the data table is flattened as a linear list of tokens in the input prompt. From the output summary, we can see that, ChatGPT introduces the \textit{COVID pandemic} as the major reason for a slight dip in \textit{Addidas net income}. Based on the increase in \textit{Puma net income} in the similar period, the suggestion of COVID-19 pandemic as the cause of the fluctuation for `Adidas' might not necessarily be accurate. Further analysis is needed to better understand the nature of model bias in the context of visualization-related tasks.

\noindent \textbf{Utilization of advanced large vision-language models:} 
Recent NLG methods rely on large language models which are known for producing fluent texts, however, these LLMs may not be optimal for chart-specific tasks because they are trained on large text corpus and/or image-text pairs without any specific focus on chart comprehension. UniChart takes an initial step in addressing the problem by building a large corpus of diverse charts (611k) and designing several low-level pertaining objectives that focus on reasoning and text generation for charts~\cite{masry2023unichart}.  
 
\begin{figure}[t]
    \centering
    \includegraphics[width=\columnwidth]{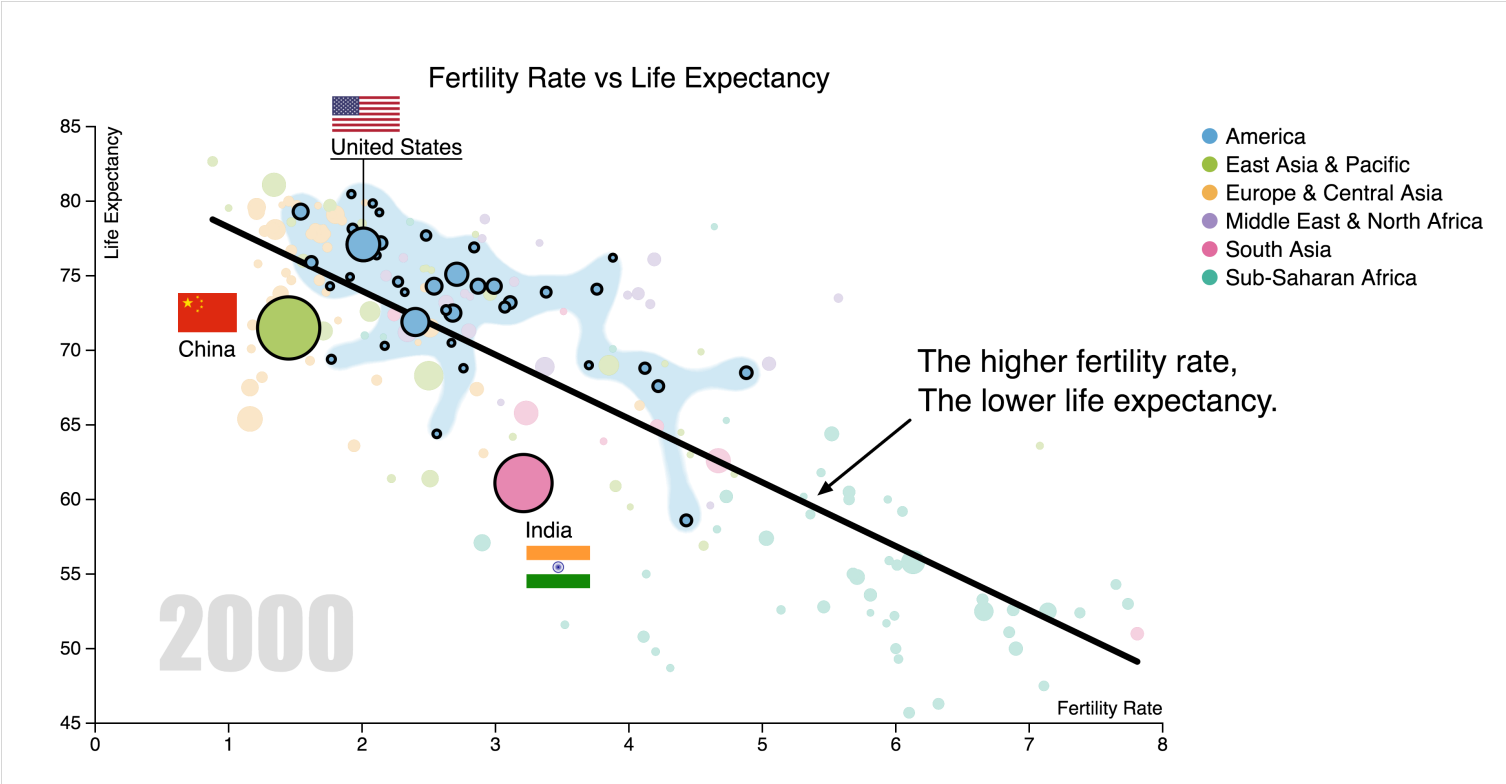}
    \caption{An example of text annotation integrated inside the visualization. Figure collected from \cite{ren2017chartaccent}. The figure explores the connection between the fertility rate and life expectancy in the United States, China, and India with text and image annotations. Additionally, for a broader perspective, countries from North and South America are marked in blue, offering a comparative regional insight.}
    \label{fig:chartaccent}
\end{figure}

\begin{figure}[t]
    \centering
    \includegraphics[width=\columnwidth]{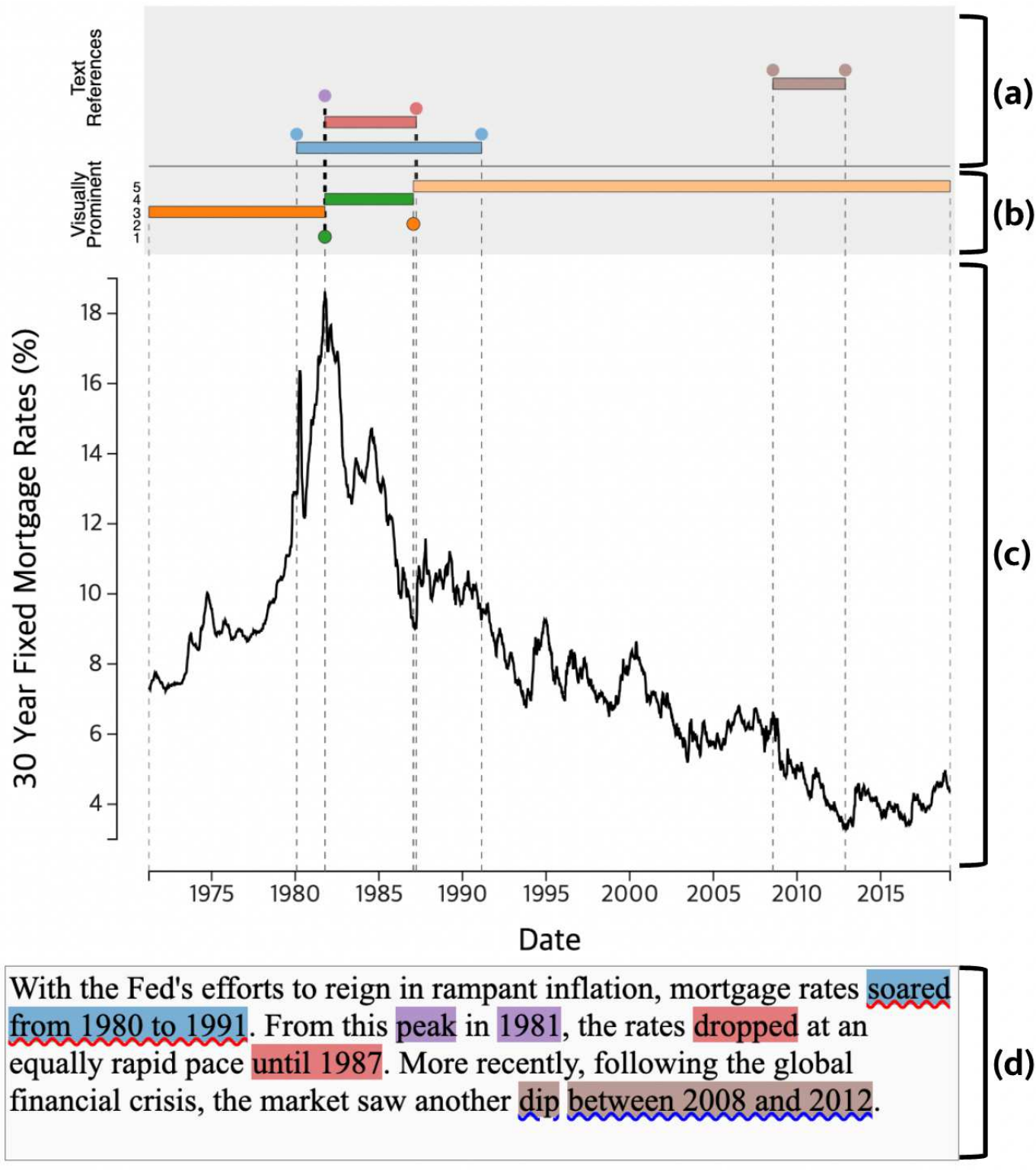}
    \caption{\textsc{EmphasisChecker}  links text and visualization. Initially, the author writes a caption in the textbox marked in \textbf{(d)} based on the chart \textbf{(c)}. The system also marks visually important features of the chart (see the portion marked by \textbf{(b)}) and the text references (see the portion marked by \textbf{(a)}).
    Here, orange \textcolor{orange}{\rule{2mm}{2mm}} color represents unmatched features and green \textcolor{green}{\rule{2mm}{2mm}} color represents matched features (at the top).  
    Additionally, the system provides a connection between the chart and the text in section \textbf{(a)} by displaying (i.e., blue \textcolor{blue}{\rule{2mm}{2mm}}, red \textcolor{red}{\rule{2mm}{2mm}}, purple \textcolor{purple}{\rule{2mm}{2mm}}, and brown \textcolor{brown}{\rule{2mm}{2mm}} marks above the chart). The system adds a \textcolor{red}{red} squiggly underline 
    % (
    % \begin{tikzpicture} \draw[red, decorate, decoration={snake, amplitude=0.5mm, segment length=1.5mm}] (0,0) -- (0.4,0); \end{tikzpicture}
    % ) 
    under the phrases that are not matched with any data in the chart and a \textcolor{blue}{blue} squiggly underline 
    % (
    % \begin{tikzpicture} \draw[blue, decorate, decoration={snake, amplitude=0.5mm, segment length=1.5mm}] (0,0) -- (0.4,0); \end{tikzpicture}
    % ) 
    under the phrases that do not match any of the important chart features.
    }
    \label{fig:emphasis-checker}
\end{figure}
\section{Where?}

\label{sec-where}
The \textit{where} aspect of the problem focuses on the placement of text, which can be achieved through various methods, including generating separate text output, linking portions of text with visualizations, embedding visualizations within paragraphs, or adding textual annotations to visualizations.

\subsection{Separate Text Output}
Most NLG systems output text as a separate entity without any explicit links to the related visualization \cite{kantharaj-etal-2022-chart, tang2023vistext, kantharaj-etal-2022-opencqa} (see Figure (a) in \ref{fig:why-dimension}). \change{Similarly, some NLG systems can generate textual captions for charts based on author's intent \cite{choi2022intentable}, and can automatically generate titles \cite{liu2023autotitle} based on facts and user feedback that are separate from visualizations.} \change{Some data storytelling systems also produce texts without linking them to relevant portions in the visualization \cite{zhao2023chartstory, islam2024datanarrative}}. \change{Similarly, LIDA \cite{dibia2023lida}, a visualization authoring tool, incorporates a multi-step process that generates intermediate textual content by analyzing the underlying data tables and the desired goals for generating the visualizations.}  While generating separate output simplifies the problem, the onus is on the reader then to coordinate between the texts and visualizations, which can be mentally taxing~\cite{kim2018facilitating}.

\subsection{Linked Text with Visualization} 
A potential solution to address the challenge of separate text output is to apply computational techniques to automatically link relevant portions of the chart with texts~\cite{kong2014extracting, kim2018facilitating, latif2021kori, shi2021calliope, kim2023emphasischecker, sultanum2023datatales}. For example, Kong and Agrawala~\cite{kong2014extracting} developed a crowd-sourcing-based approach for automatically creating references between text phrases that refer to different bars in a bar chart so that clicking on the text highlights relevant bars in the chart. Kori is a mixed-initiative interface enabling users to construct interactive references between text and charts~\cite{latif2021kori}. VisJockey is a storytelling tool that helps readers make connections between text and visualization in a data story by manually creating references~\cite{kwon2014visjockey}. However, the above methods are not fully automatic. Voder generates data facts and then supports interactive visual linking between individual facts and the corresponding elements of the visualization~\cite{srinivasan2019augmenting}. TimelineCurator automatically extracts temporal references in the text to generate a visual timeline~\cite{brehmer2019timeline}. More recently, Kim et al. presented EmphasisChecker~\cite{kim2023emphasischecker}, which automatically highlights textual phrases that refer to different portions of a line chart (See Figure \ref{fig:emphasis-checker}). Despite these efforts to automatically link texts and visualizations, concerns remain about the accuracy of these methods.

\subsection{Text Annotations within Visualization} 
In contrast to separate text outputs, some approaches automatically annotate salient portions in visualizations with textual labels~\cite{hullman2013contextifier, gao2014newsviews, liu2021advisor}. Contextifier takes a line chart about stocks and relevant articles as input and then automatically annotates the salient points in the line chart with relevant summaries extracted from the articles \cite{hullman2013contextifier}. Similarly, NewsViews \cite{gao2014newsviews} leverages a news corpus and a related data table corpus as inputs to the system. They followed the ``inverted pyramid'' approach in journalism which denotes the first several sentences to capture the most important information in the article and employ a query extractor to extract crucial facts that are later used as annotations to the output visualizations. ChartAccent \cite{ren2017chartaccent} provides an interactive platform for users to incorporate annotations directly into their visualizations (presented in Figure \ref{fig:chartaccent}). Similarly, the Calliope system~\cite{shi2021calliope} combines both captioning as a distinct element from the chart and annotated visualizations as part of the final output (see Figure \ref{fig:why-dimension}c). \change{In the same vain, ChartSpark \cite{xiao2023let}, a recent visualization tool designed for creating pictorial visualizations that infuse charts with semantic context using a text-to-image generative model, ensures key visual attributes like trends are maintained. The visualizations generated by the tool include rich textual annotations within the charts.} Overall, research on the automatic annotation of visualizations with text remains relatively limited.

\subsection{In-situ Visualization Within Text}
Another possible way to coordinate between texts and visualizations is to provide in-situ visualizations within texts. For example, Charagraph dynamically generates interactive charts and annotations within text documents through manual annotation \cite{masson2023Charagraph}. The reader can preview the selection by hovering over options in the menu, and clicking an option immediately creates a chart from the selected values and it gets annotated within the text paragraph thus creating a Charagraph. Charagraph supports common data exploration tasks through interactive features such as identifying and comparing values.
However, such approaches have not been applied in the context of NLG outputs yet.

\subsection{Discussion}
As previously mentioned, the majority of NLG systems output texts separately without any explicit links to the visualization which makes it difficult for the users to mentally integrate the generated texts with visualizations. While there are a few methods for automatically linking texts and visualizations, they usually rely on some heuristics that result in frequent errors in the extracted references. More advanced vision-language models may be employed to improve the automatic linking between texts and visualization.

A recent study revealed that users generally prefer visualizations with substantial text annotations over those with only text or fewer annotations in visualizations~\cite{stokes2022striking}. The study also found that users prefer text explaining the statistical or relational aspects of a chart, eliciting more insights related to statistics or relational comparisons than text focusing on basic or encoded elements of the chart. Therefore, additional efforts are required to generate multimodal outputs in a coordinated manner, ensuring that salient features of the chart are accurately detected and annotated with informative text. A promising direction could involve a human-in-the-loop approach that combines manual annotation~\cite{ren2017chartaccent} with automated suggestions for annotation.

\section{When?}
\label{sec-when}
The \textit{when} dimension pertains to the temporal organization of generated texts. 
This aspect is closely related to document planning in NLG architecture~\cite{reiter2007architecture} and narrative structure in data-stories~\cite{riche2018data}, which specifically concerns how the sequence of events, comprising textual elements, is temporally ordered and the potential variations in the pacing of the narrative. The narrative structure can be categorized in various ways.

\subsection{Linear vs. Non-linear sequence} The narrative structure that follows a linear sequence is mostly common with highly explanatory long-form text, that often has a beginning, and an end and follows a linear progression according to \cite{riche2018data}. Most of the existing works in visual storytelling \cite{shi2021calliope, sun2023erato} utilize a Monte Carlo Tree Search method \cite{browne2012asurvey, silver2016mastering} to initiate the logical linear ordering of facts which is later utilized to generate a narrative story that follows a linear sequence (see Figure ~\ref{fig:why-dimension}c and Figure ~\ref{fig:visstory}). In contrast,  Cui et al.~\cite{cui2020text-to-viz} and Chen et al.~\cite{chen2020towards} employed deep learning-based methods to generate linear sequential stories with infographics. Non-linear narratives, on the other hand, provide a more exploratory approach, allowing audiences to interact with the story and discover information in a non-sequential manner~\cite{riche2018data}. 
 This flexibility in storytelling can enhance engagement, allowing for personalized experiences where the reader determines the order and exploration depth of the narrative. This is mainly observable in studies such as \cite{ren2017chartaccent, wu2023socrates} that elicit user interaction in the story generation process via a user interface. In contrast, Zhao et al.~\cite{zhao2023chartstory} automatically generate a narrative story in a non-linear manner by following design patterns for data comics \cite{bach2018design}. 

\subsection{Author-driven vs. Reader-driven} The author-driven approach is characterized by a linear ordering of visualizations, heavy messaging, and no interactivity. In this structure, the author controls the narrative path and the messaging, guiding the viewer through the content in a predetermined manner without deviation. Most of the automatic visual story generation systems such as \cite{shi2021calliope, sun2023erato, cui2020text-to-viz, chen2020towards} follow this structure. In contrast, the reader-driven approach adopted by \cite{ren2017chartaccent, wu2023socrates, zhao2023chartstory} allows for no prescribed ordering, omits messaging, and offers free interactivity to users allowing them to choose the narrative structure of the story along with the facts that requires to be included in the story.  

\subsection{Discussion}
The \textit{when} aspect of the problem dimension is rarely explored by the existing NLG research for visualizations and most models opted for a simple linear narrative structure. One starting point in this direction would be to collect and annotate a large collection of visual data stories with a variety of narrative structure~\cite{segel2010narrative, hullman2013adeeper}. Another promising direction would be to leverage LLM such as LLama-2 \cite{touvron2023llama}, GPT-4 \cite{openai2023gpt4} and Gemini \cite{google2023gemini} to automatically generate different possible narrative structures in a particular scenario and involve users in the loop to refine the narrative structure for a story. 

\change{While the \textit{when} dimension primarily concerns visual storytelling applications, it is also relevant to static data report generation, where ordering the generated texts coherently is a key concern. Specifically, determining how to order key insights in a generated data-driven article is a topic that has been rarely addressed in the context of data report generation. Therefore, future research should address this concern.}

\change{
\section{Key Takeaways, Open Challenges, and Future Directions}
\label{challenges}
So far, we have analyzed the problem of NLG for visualizations through the lens of individual \change{five Wh-questions} dimensions. In this section, first we will highlight key takeaways for the research community. Then,  we will discuss the challenges and opportunities for future research in a holistic manner, examining the intersections of these dimensions. 
 \subsection{Key Takeaways}
The primary motivation for employing NLG in visualization is to facilitate understanding and interaction with visual data through natural language. This motivation enables the introduction of various tasks (\textit{\textbf{Why}} dimension) such as chart summarization, question answering, and data-driven storytelling, each requiring a nuanced understanding of visual data and its effective verbal articulation. On the other hand, the \textit{\textbf{What}} dimension highlights the diversity of input data (visualizations, data tables, multimodal inputs) and the variety of natural language outputs (summaries, captions, data facts, narratives) revealing the versatility of NLG systems in processing different forms of visual data and generating relevant textual content, which directly supports the objectives identified in the \textit{\textbf{Why}} dimension. The methods and technologies employed to transform visual inputs into textual outputs constitute the \textit{\textbf{How}} dimension, which includes the application of computer vision, optical character recognition, rule-based approaches, and deep learning models, which is crucial for achieving the goals set forth in the \textit{\textbf{Why}} dimension. Furthermore, the placement of generated text (the Where dimension) in relation to the visual data addresses the integration of textual and visual information, which is vital for ensuring that the textual outputs of NLG systems are effectively aligned with the corresponding visual elements. This dimension is directly related to a downstream task (\textit{\textbf{Why}}), the input/output content (\textit{\textbf{What}}), and input/output processing (\textit{\textbf{How}}). Finally, the \textit{\textbf{When}} dimension focuses on the temporal organization of textual narratives, crucial for data storytelling and interactive exploration. It encompasses the pacing and sequencing of textual elements, impacting how information is conveyed and understood over time, supporting the goals of the \textit{\textbf{Why}} dimension by ensuring that narratives are not only informative but also engaging and dynamically aligned with user interaction. The above five interrelated questions can provide a holistic lens for characterizing the NLG problem and the design space of proposed solutions for researchers. 

\subsection{Open Challenges and Future Directions}

Based on our analysis of the NLG problem space for visualizations, we now discuss the open challenges and directions for future research.

\noindent\textbf{Improve generalizability of NLG models:} 
In real-world applications, there are many different tasks, such as visualization education, explaining uncertainty, and fact-checking with visualizations. A common strategy has been to fine-tune various models originally trained on vision tasks and language. However, such task-specific models are not capable of solving a wide range of chart-related tasks, limiting their real-world applicability. Instruction tuning can be a promising direction to fine-tune the model on a large collection of diverse tasks \cite{masry2024chartinstruct, masry2024chartgemma}, supporting a wide array of real-world chart comprehension and reasoning scenarios. This approach can expand the scope and applicability of our models to new kinds of tasks. Nonetheless, more effort is needed in developing and adopting vision-language models \cite{liu2023visual} that can perform various NLG tasks with visualizations without further fine-tuning.

\noindent\textbf{Introduce new benchmarks:} Another challenge for improving the generalizability of models is the lack of benchmark datasets on the NLG task for visualization. Moreover, current datasets lack diversity in terms of topics, chart types, and visual styles \cite{masry2023unichart}. Most of them lack rich semantics that explains the domain and contextual aspects of various patterns, trends, and outliers in text descriptions \cite{lundgard2021accessible}. This limitation hinders robust benchmarking and model development. There is a need for creating more diverse and comprehensive datasets that cover a wide range of chart types and visual styles. These datasets should also include rich semantic annotations to provide context and domain-specific information, which will enable better training and evaluation of NLG models. Developing such benchmarks will facilitate the creation of more robust and generalizable models.

\noindent\textbf{Incorporate human-in-the-loop:} \change{Some previous works in data storytelling have integrated the storyteller's preferences into automated story generation workflows through conversational recommendations, as demonstrated by Wu et al. in Socrates \cite{wu2023socrates}. In the domain of data visualization generation, Song et al. \cite{song2023marrying} proposed a method that constructs visualizations through a series of interactions between the system and the visual data analyst. Their dialogue system dynamically generates responses based on context, which may include text, data, or data visualizations.} In a similar vein, incorporating human-in-the-loop approaches would empower users to iteratively refine the generated results based on their specific needs. This interaction allows \change{data analysts} to provide feedback and corrections, which can be used to improve the model's performance over time. By enabling users to have a more active role in the generation process, NLG systems can produce more accurate and contextually appropriate outputs, tailored to the user's requirements. This approach can enhance user satisfaction and trust in NLG systems.

\noindent \textbf{Addressing chart data Extraction:} Robust data extraction from image-based charts remains a challenging problem. Addressing this, alongside developing dedicated models for correcting factual errors, is key for real-world applications. Ensuring the accuracy of the extracted data is critical, as any errors can lead to misleading or incorrect interpretations of the visualized information. Developing advanced data extraction techniques and integrating them with fact-checking mechanisms will help to ensure the reliability and trustworthiness of NLG outputs. This is particularly important for applications in domains such as journalism, healthcare, and finance, where accuracy is paramount.

\noindent \textbf{Enhance chart accessibility for diverse user groups:} While NLG outputs have been integrated with several tools for chart accessibility~\cite{alam2023seechart, voxLens}, they primarily target limited user groups, such as those with visualization disabilities. Numerous other user groups, including individuals with low visualization literacy~\cite{echeverria2018driving}, those with intellectual and developmental disabilities~\cite{wu2021understanding}, and individuals living with dementia~\cite{ang2023advocating}, encounter challenges in comprehending both the chart content and the associated generated captions. Tailoring and adapting the generated captions to suit the unique needs of diverse user groups can contribute significantly to promoting accessible and inclusive visualizations to a broader range of populations.

\noindent\textbf{Addressing ethical concerns:} Addressing ethical concerns related to NLG outputs for visualizations has been underexplored in the research community. In Section \ref{sec-how} we already discussed how model bias in NLG outputs can lead to unfair or discriminatory outcomes, which can have serious ethical implications. Recent research reveals the bias with hallucinated outputs from foundation models like GPT4 for various tasks\cite{cui2023holistic, islam2024large}. However, a holistic analysis of bias and inference concerns arising from LLM-generated texts for visualizations is missing. A promising direction could be to employ techniques for bias detection and mitigation, such as using diverse training data, implementing fairness constraints, and conducting thorough evaluations of model outputs.  

It is also crucial to examine whether LLMs mislead with generated texts and to ensure ethical standards are maintained in all applications of explaining visualizations. Research suggests that when visualizations are designed poorly or maliciously, they can become misleading or even deceiving~\cite{lo2022misinformed}. One key research question is do LLMs get deceived by misleading charts and consequently produce misleading texts? There is a pressing need to address ethical issues and to create guidelines and best practices for the ethical use of NLG in visualizations. This will help to build trust and confidence in these systems among users and stakeholders.
}

\section{Conclusion}
\label{sec-concl}

\change{In this paper, we have presented a systematic review of the current state-of-the-art NLG systems for visualization by surveying related studies from the domains of human-computer interaction, information visualization, and natural language processing. Through this analysis, we proposed a taxonomy based on \change{five \textit{Wh-questions}} presented in Section \ref{sec: pso} to define the problem space, synthesized findings from existing literature relevant to these aspects, and outlined crucial research gaps.

Our review highlights several key challenges requiring further exploration, such as improving the generalizability of NLG models and developing diverse benchmark datasets with rich semantic annotations. These steps will enhance model robustness and versatility. Incorporating human-in-the-loop approaches can be promising in refining generated results, increase accuracy, and build user trust. Furthermore, addressing robust data extraction from image-based charts is critical for ensuring reliability, especially in precision-demanding domains like journalism, healthcare, and finance. NLG for visualization can also enhance chart accessibility for diverse user groups to promote inclusivity. Finally, addressing ethical concerns such as fairness and bias mitigation remains a significant challenge. We hope this survey will inspire other researchers to address these challenges and make important contributions to this emerging area of visualization research at the intersection of natural language generation and visualization.}

\section*{Acknowledgement}
This work was supported by the Natural Sciences and Engineering Research Council (NSERC), Canada, Canada Foundation for Innovation and the CIRC  grant on Inclusive and Accessible Data Visualizations and Analytics. We thank anonymous reviewers for their valuable comments and suggestions.

\newpage
\bibliographystyle{eg-alpha-doi}

% \bibliography{cqa}
\bibliography{newBiblio}
% biblatex with biber
% \printbibliography                

%-------------------------------------------------------------------------
% \newpage
\newpage

\noindent \textbf{Short Biographies of Authors}

\noindent \textbf{Enamul Hoque} 

\noindent Enamul Hoque is an Associate Professor 
at York University where he directs the Intelligent Visualization Lab. Previously, he was a postdoctoral fellow 
at Stanford University. He received a Ph.D. degree in Computer Science from the University of British Columbia. His research focuses on combining information visualization and human-computer interaction with natural language processing to address the challenges of the information overload problem.   
Along with his collaborators, he developed several natural language interfaces for supporting data analysis and question answering with visualizations~\cite{hoque2018applying-pragmatics, sneak-peak, miva-2021}. More recently, he worked on various natural language generation tasks for visualizations including automatic chart summarization~\cite{obeid2020charttotext, kantharaj-etal-2022-chart, masry2023unichart} and chart question answering~\cite{hoque2022chart, kim2020answering, kantharaj-etal-2022-opencqa, masry-etal-2022-chartqa}.  
He served as an Area Chair for the ACL Rolling Review (2021-2023) and as a program committee member (2018-2022) for the IEEE Vis. 
His research has been supported by Natural Sciences and Engineering Research Council of Canada, the National Research Council of Canada, and the Canada Foundation for Innovation among others. 

\noindent \textbf{Mohammed Saidul Islam}

\noindent Mohammed Saidul Islam is pursuing Master's in Computer Science at York University under the supervision of Enamul Hoque. He completed his BSc from Islamic University of Technology (IUT) in Bangladesh. His research interests include Natural Language Processing, Natural Language Generation with Charts, Fact verification, etc. 
\end{document}